\documentclass[10pt,twocolumn,letterpaper]{article}

\usepackage{iccv}
\usepackage{times}
\usepackage{epsfig}
\usepackage{graphicx}
\usepackage{amsmath}
\usepackage{amssymb}
\usepackage{caption}
\usepackage{algorithm} 
\usepackage{algpseudocode} 
\usepackage{booktabs}
\usepackage{multirow}
\usepackage{enumitem}
\usepackage{rotating}
\usepackage{array,multirow,graphicx}
\usepackage{authblk}
\usepackage[percent]{overpic}

\usepackage[pagebackref=true,breaklinks=true,letterpaper=true,colorlinks,bookmarks=false]{hyperref}

\def\iccvPaperID{3479}

\iccvfinalcopy

\ificcvfinal\fi
\newcommand\vsp{\vspace*{-0.2cm}}
\newcommand\yb{\mathbf{y}}

\newcommand\tmone{{t\text{--}1}}
\newcommand\Yb{\mathbf{Y}}

\newcommand\xb{\mathbf{x}}

\newcommand\Db{\mathbf{D}}

\newcommand\Jb{\mathbf{J}}
\newcommand\ub{\mathbf{u}}
\newcommand\pb{\mathbf{p}}

\newcommand\zb{\mathbf{z}}

\newcommand\Bb{\mathbf{B}}

\newcommand\rb{\mathbf{r}}
\newcommand\varepsilonb{{\boldsymbol \varepsilon}}

\usepackage{pifont}

\def\eg{\emph{e.g.}} 

\DeclareMathOperator*{\argmin}{arg\,min}

\def\comment#1{{}}

\begin{document}

\title{Lucas-Kanade Reloaded: End-to-End Super-Resolution from Raw Image Bursts}

\author{ \vspace{-0.9cm} Bruno Lecouat$^{1,2}$ Jean Ponce$^{1,3}$ Julien Mairal$^2$\\
    $^1$Inria and DIENS (ENS-PSL, CNRS, Inria), Paris, France\\
    $^2$Univ. Grenoble Alpes, Inria, CNRS, Grenoble INP, LJK, 38000 Grenoble, France \\
    $^3$Center for Data Science, New York University, New York, USA\\

    { \tt  firstname.lastname@inria.fr}
    }

\twocolumn[{%
\small

\newcommand\h{0.25}
        \renewcommand\twocolumn[1][]{#1}%
        \maketitle
        \renewcommand{\arraystretch}{0.2} 
        \setlength\tabcolsep{0.5pt}
        \vspace{-0.8cm}
        \begin{tabular}{cccc}

            \includegraphics[trim=0 0 0 0, clip, width=\h\textwidth]{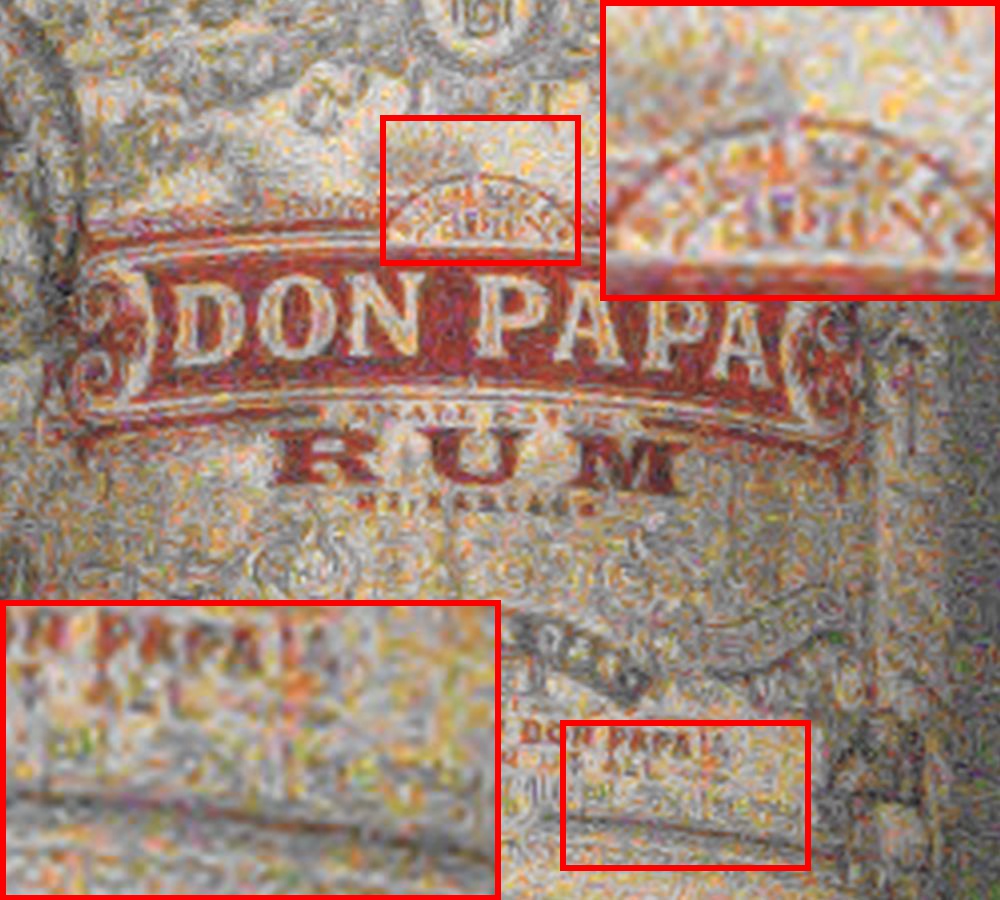} &  
            \includegraphics[trim=0 0 0 0, clip, width=\h\textwidth]{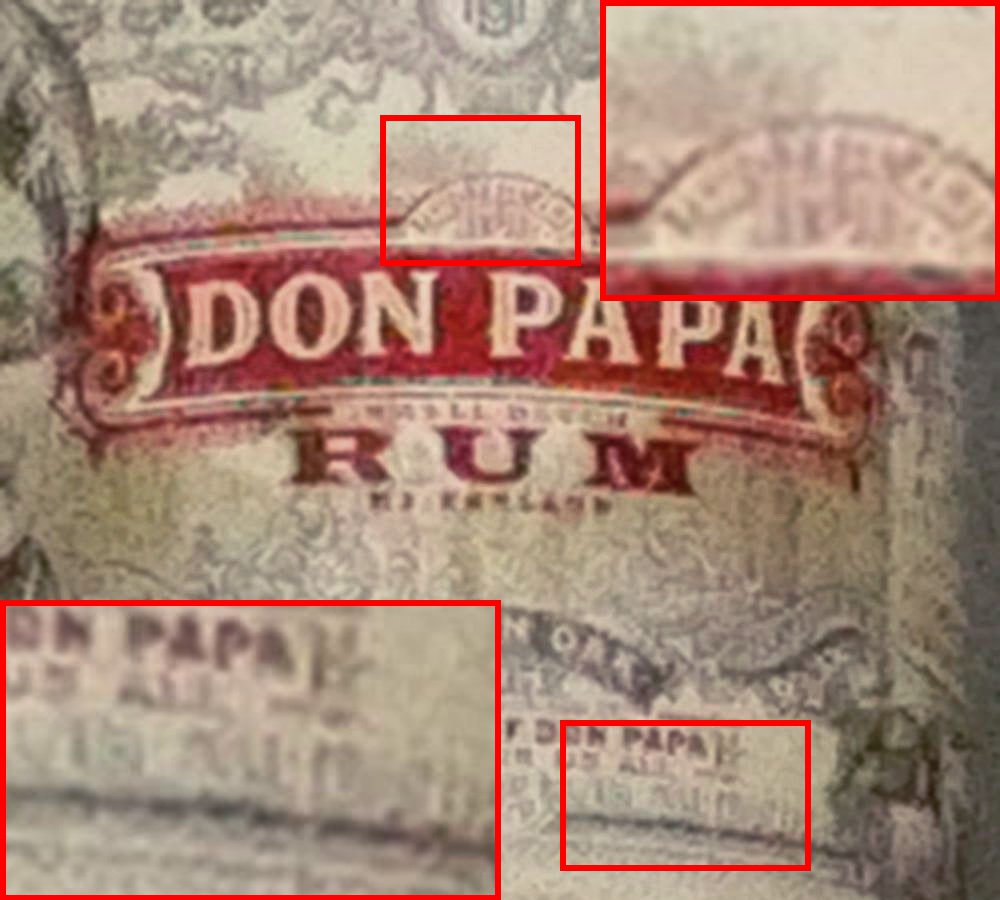} &  
            \includegraphics[trim=0 0 0 0, clip, width=\h\textwidth]{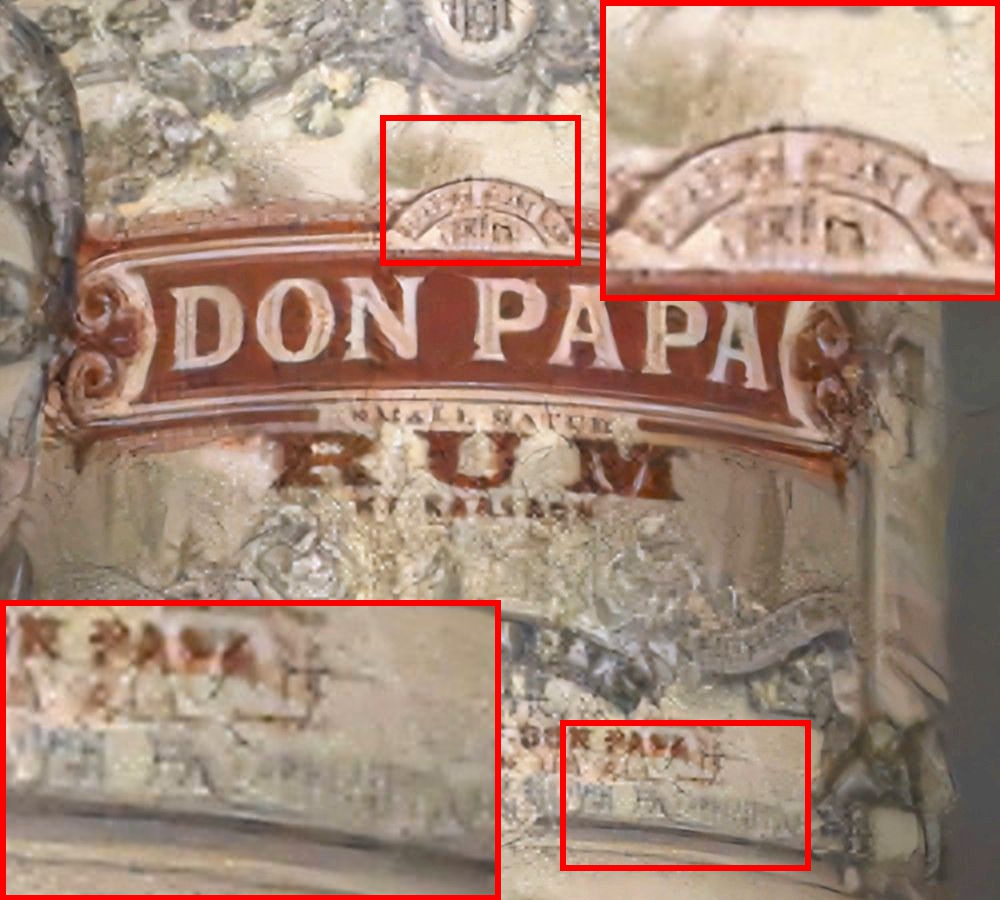} &  
            \includegraphics[trim=0 0 0 0, clip, width=\h\textwidth]{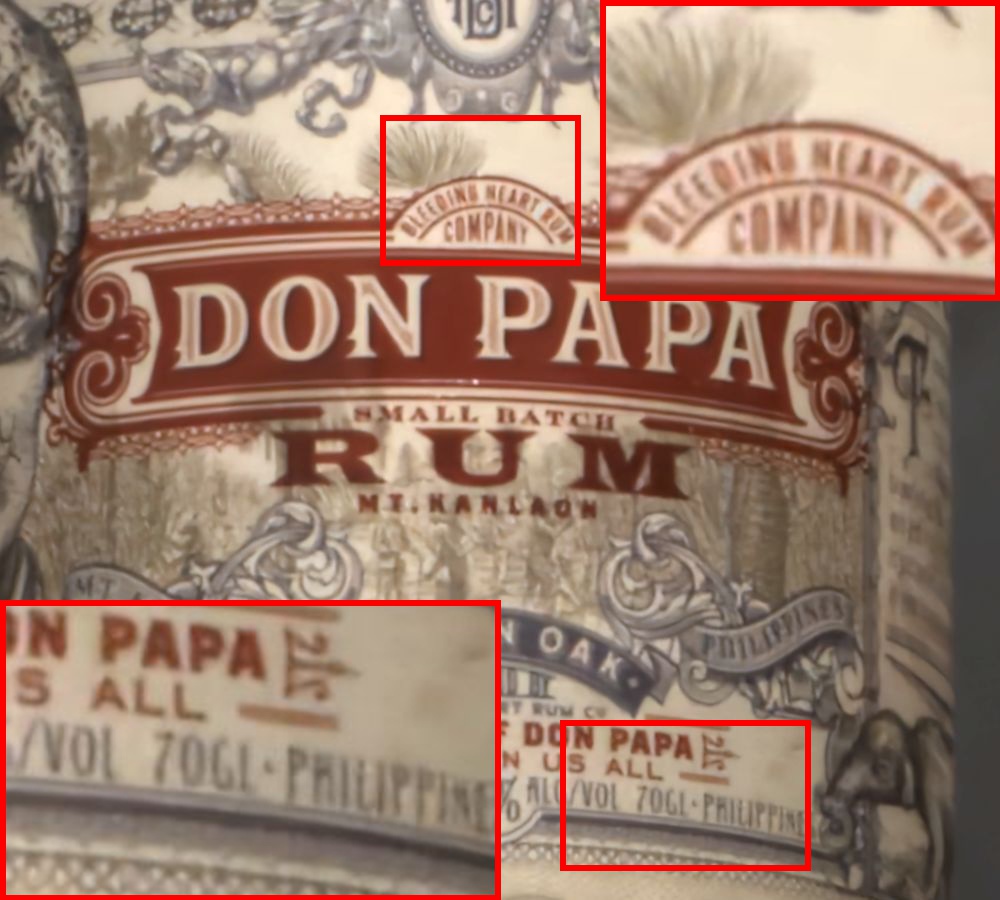} \\ 

        \includegraphics[width=\h\textwidth]{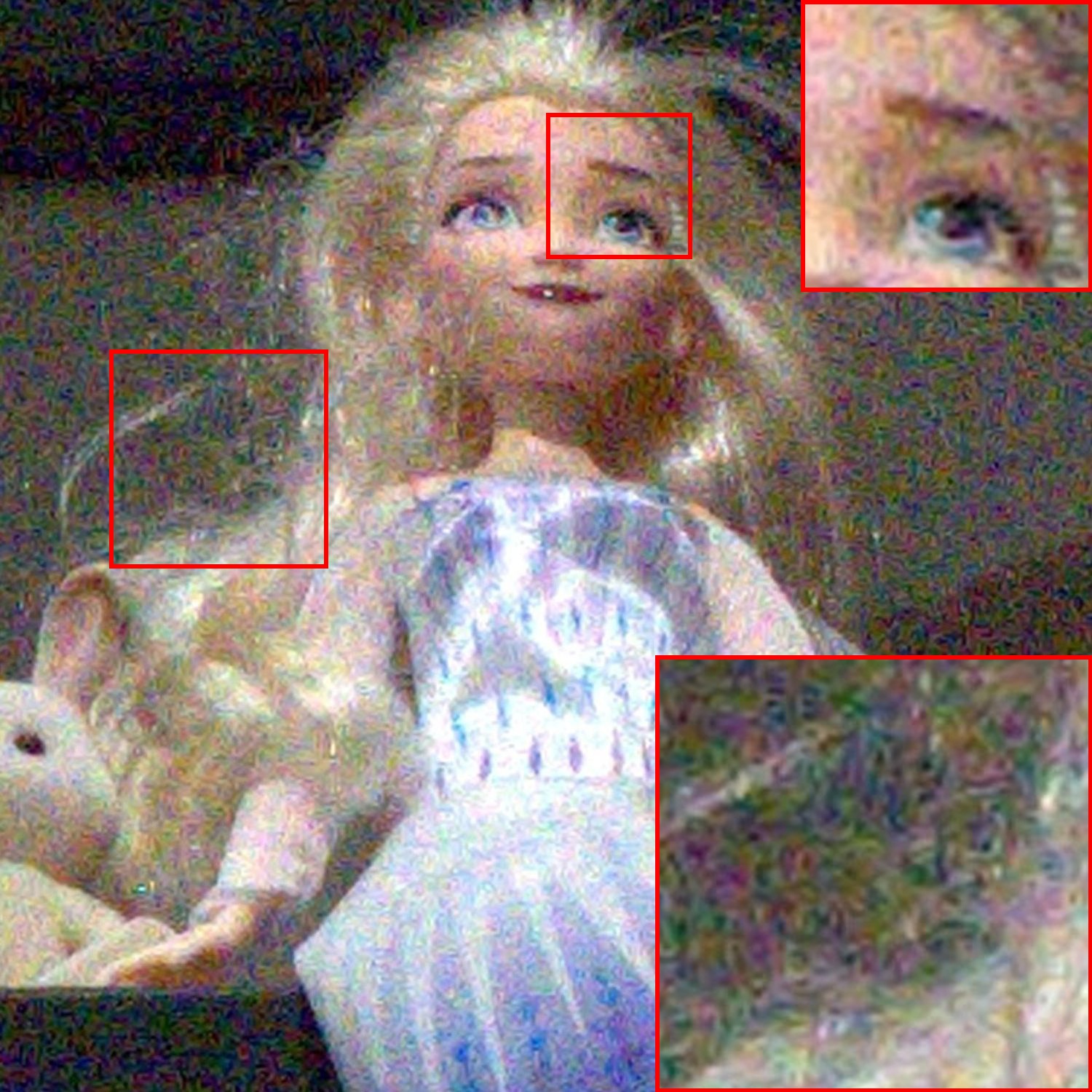} &  
        \includegraphics[width=\h\textwidth]{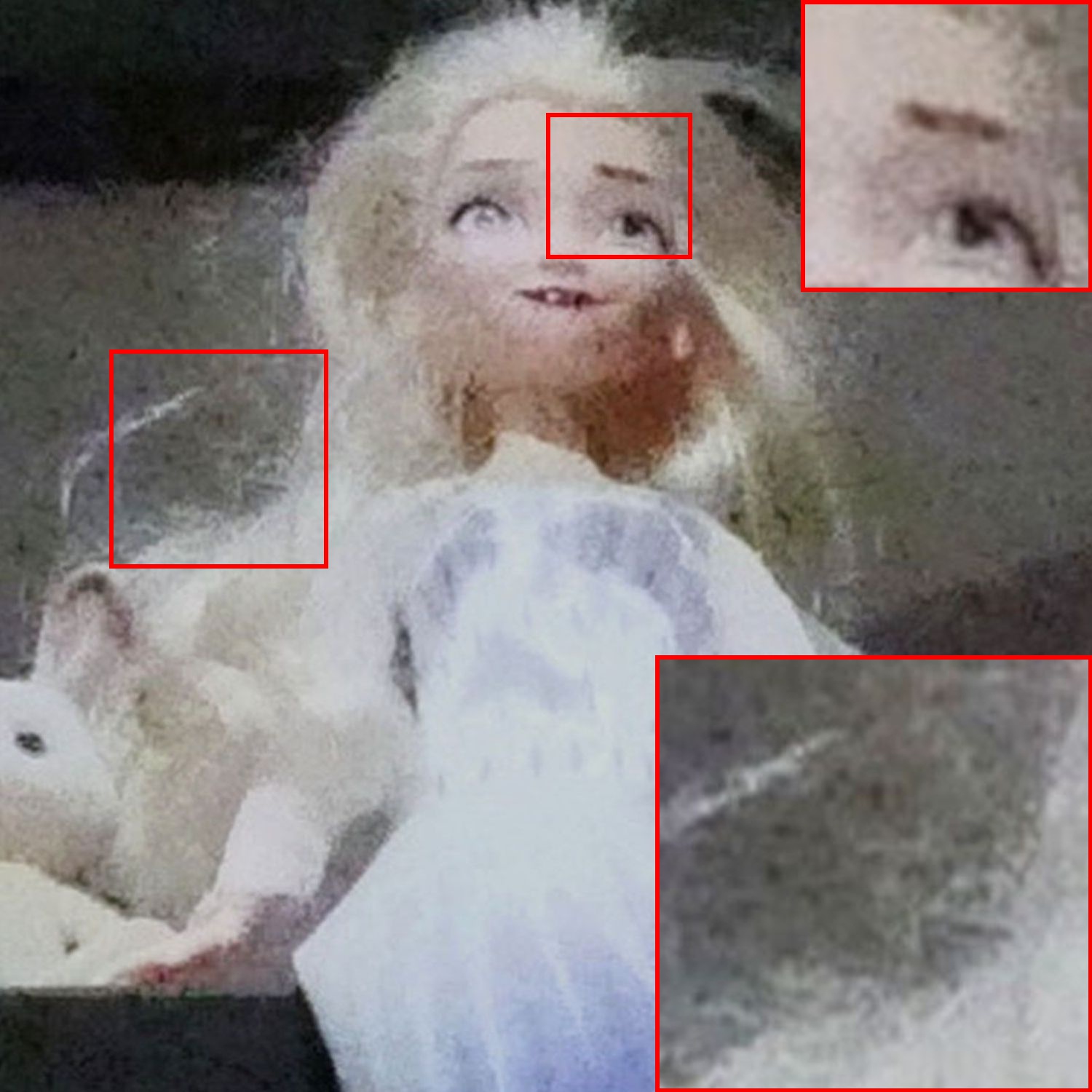} &  
        \includegraphics[width=\h\textwidth]{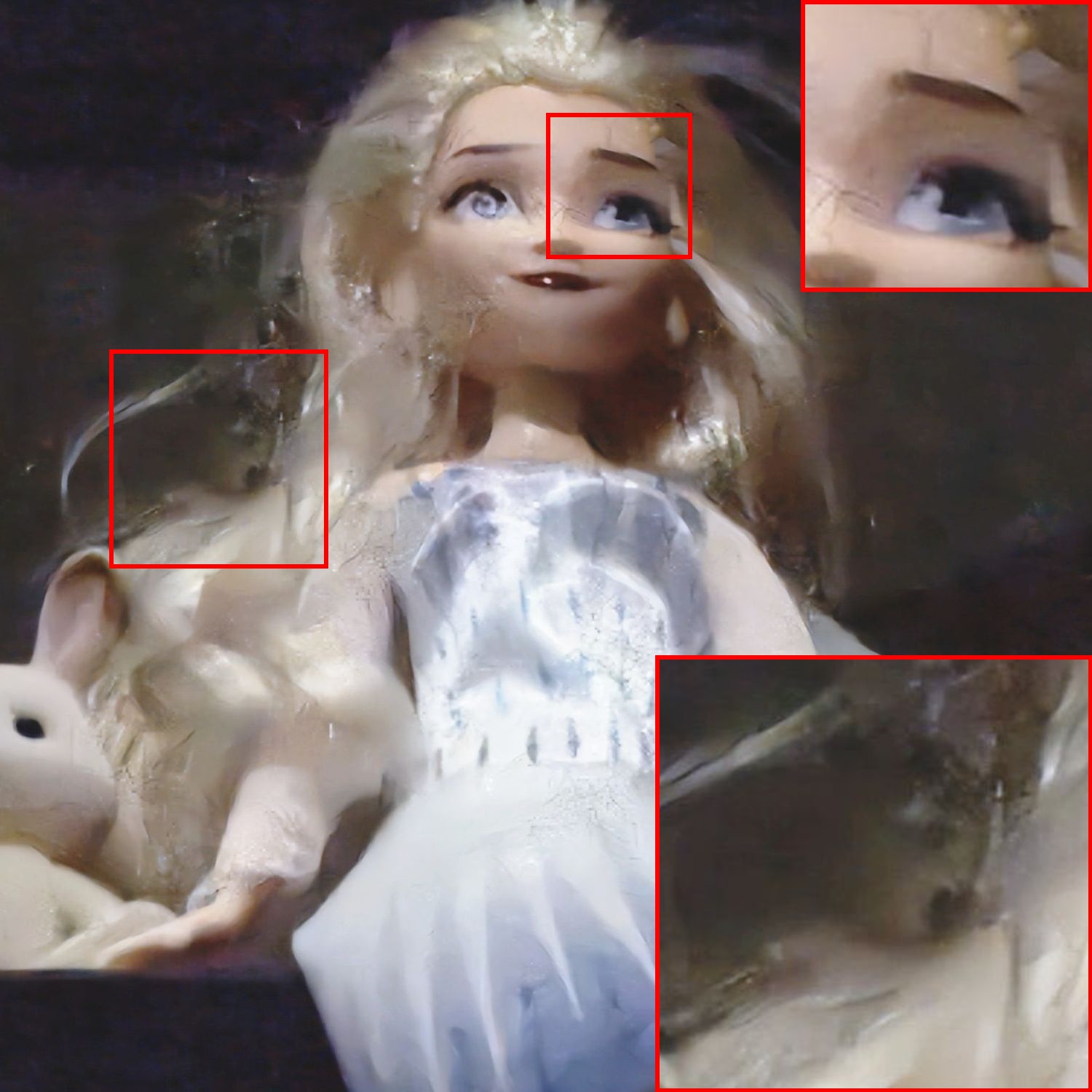} &  
        \includegraphics[width=\h\textwidth]{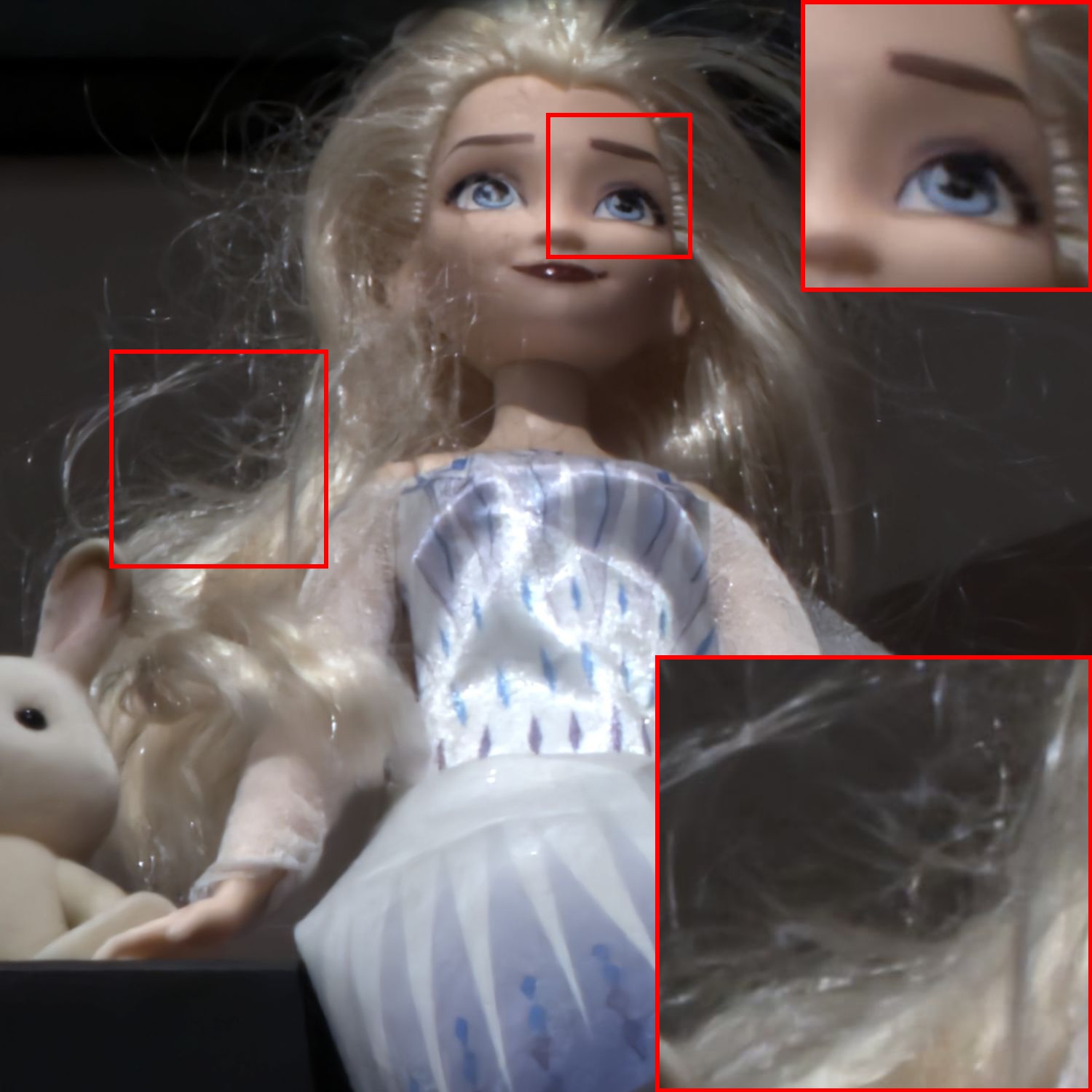} \\
 
        Output of Dcraw  & Camera's ISP (hi-quality jpeg) & Joint demosaick+single-image SR & Proposed method

        \end{tabular}
        \vsp
        \captionof{figure}{$\times 4$ super-resolution results obtained from a burst of 30 raw images acquired with a handheld Panasonic Lumix GX9 camera at 12800 ISO for the top image and 25600 for the bottom image. Dcraw performs basic demosaicking.} \vspace*{0.4cm}\label{fig:teaser}
    }]

   \begin{abstract}
   \vspace*{-0.4cm}

This presentation addresses the problem of reconstructing a
high-resolution image from multiple lower-resolution snapshots
captured from slightly different viewpoints in space and time.  Key
challenges for solving this
{\em super-resolution} problem include (i) aligning the input pictures
with sub-pixel accuracy, (ii) handling raw (noisy) images for maximal
faithfulness to native camera data, and (iii) designing/learning an
image prior (regularizer) well suited to the task.  We address these
three challenges with a hybrid algorithm building on the insight
from~\cite{wronski2019handheld} that aliasing is an ally  
in this setting, with parameters that can be learned end to
end, while retaining the interpretability of classical approaches to
inverse problems.  The effectiveness of our approach is
demonstrated on synthetic and real image bursts, setting a new state
of the art on several benchmarks and delivering excellent qualitative
results on real raw bursts captured by smartphones and prosumer
cameras. Our code is available at \url{https://github.com/bruno-31/lkburst.git}.

   \vspace*{-0.4cm}
\end{abstract}

\section{Introduction}
The problem of reconstructing high-resolution (HR) images from lower-resolution
(LR) ones comes in multiple flavors, that may significantly differ from each
other in both technical detail and overall objectives. When a single LR image
is available, the corresponding inverse problem is severely ill-posed, requiring very strong
priors about the type of picture under consideration~\cite{hou1978cubic,Yang11}. 
For natural images, data-driven methods based on
convolutional neural networks (CNNs) have proven to be very
effective~\cite{ledig2017photo,wang2018esrgan}. 
Generative adversarial
networks (GANs) have also been used to synthesize  impressive
HR images that may, however, contain ``hallucinated'' high-frequency details~\cite{dong2015image,lim2017enhanced}. 

%
%

In the true {\em super-resolution} setting~\cite{Milanfar11,Tsai84,Yang11},\footnote{``Single-image super-resolution''
  has become a popular nickname for single-image upsampling under strong priors; here, we use the classical definition of super-resolution from multiple LR snapshots~\cite{Tsai84,Yang11}.}
where multiple LR frames are available, HR details {\em are} present in the data, but they are spread among multiple misaligned images, with technical challenges such as recovering sub-pixel registration, but also the promise of recovering veridical information in applications
ranging from amateur photography to astronomy, biological and medical imaging, microscopy imaging, and remote sensing.

Videos are of course a rich source of multiple, closely-related pictures of the same scene, with several recent approaches to super-resolution in this domain, often combining data-driven priors from CNNs with self-similarities between frames~\cite{jo2018deep, li2020mucan, wang2019edvr}.  However, most digital videos are produced by a complex pipeline mapping raw sensor data to possibly compressed, lower-resolution frames, resulting in a loss of high-frequency details and spatially-correlated noise that may be very difficult to invert~\cite{farsiu2004advances}.  With the ability of modern smartphone and prosumer cameras to record raw image bursts, on the other hand, there is a new opportunity to restore the corresponding frames {\em before} the image signal processor (ISP) of the camera produces irremediable damage~\cite{bhat2021deep,wronski2019handheld}.  This is the problem addressed in this presentation, and it is challenging for several reasons: (i) images typically contain unknown motions due to hand tremor,\footnote{Image bursts acquired on a tripod may also present subpixel misalignments in practice due to floor vibrations, as observed in our experiments.}  making subpixel alignment difficult; (ii) converting noisy raw sensor data to full-color images is in itself a difficult problem known as {\em demosaicking}~\cite{kimmel1999demosaicing,lecouat2020fully}; and (iii) effective image priors are often data driven, thus requiring a differentiable estimation procedure for end-to-end learning.

In this paper, we jointly address these issues and propose a new approach that
retains the interpretability of classical inverse problem formulations while
allowing end-to-end learning of models parameters. This may be seen as
a bridge between the ``old world'' of signal processing and the ``brave new one'' of
data-driven black boxes, without sacrificing interpretability: On the one hand,
we address an inverse problem with a model-based
optimization procedure alternating motion  and HR image estimation 
steps, directly building on classical work from the 1980s~\cite{baker2004lucas,lucas1981iterative} and 1990s~\cite{hardie1997joint}.  On the other hand, we also fully exploit modern technology in the form of a {\em plug-and-play} prior~\cite{chan2016plug,venkatakrishnan2013plug} that gracefully mixes deep neural networks with variational approaches.  In turn, unrolling the optimization procedure~\cite{chen2018theoretical,lecouat2020fully,zhang2020deep} allows us to learn the model parameters end to end by using training data with synthetic motions~\cite{bhat2021deep}.

Since aliasing produces low-frequency artefacts associated with undersampled high-frequency components of the original signal, it is typically considered a nuisance, motivating camera manufacturers to add anti-aliasing (optical) filters in front of the sensor.\footnote {There is, however,
  a trend today toward removing these filters, as in the prosumer camera used in
  some of our experiments with real images.} Yet, aliased images carry high-frequency information, which may be recovered from multiple shifted measurements. Perhaps surprisingly, aliasing is thus an ally in the context of super-resolution, a fact already noted in earlier references, see~\cite{vandewalle2006aliasing}. 
As shown in the rest of this presentation, our approach to raw burst super-resolution
also exploits this insight, and it achieves a new state of the art on several standard benchmarks that use synthetic motion for ground truth. It also gives excellent qualitative results on real data obtained with smartphone and prosumer cameras. Interestingly, as illustrated by Figure~\ref{fig:teaser}, our method has turned out to be surprisingly robust to noise given the particularly challenging setting of
raw image super-resolution, which involves simultaneous blind denoising, demosaicking,  registration, and upsampling.


\vsp
\paragraph{Summary of  contributions.}
\begin{itemize}[leftmargin=*,itemsep=0pt,parsep=0pt,topsep=0pt]
   \item To the best of our knowledge, we propose the first model-based architecture learnable end to end for joint image alignment and super-resolution from raw image bursts. 
   \item We introduce a new differentiable image registration mo\-dule that can be applied to images of different resolutions, is readily integrable in neural architectures, and may find other uses beyond super-resolution.
   \item We show that our approach gives excellent results on both real image bursts
      (with up to $\times 4$ upsampling for raw images) and synthetic ones (up to $\times 16$ for RGB images).
   \end{itemize}

\section{Related Work}

\paragraph{Classical multiframe super-resolution.}
Tsai and Huang wrote the seminal paper in this setting~\cite{Tsai84}, with a restoration model in the frequency domain assuming known translations between frames. Most latter approaches have focused on the spatial domain, and they generally fall into two main categories~\cite{kohler2018multi}: In interpolation-based methods, LR snapshots aligned with sub-pixel precision are jointly interpolated into an HR image~\cite{hardie2007fast,takeda2007kernel}.  Impressive results have recently been obtained for hand-held cameras using the variant of this method proposed by Wronski {\em et al.}~\cite{wronski2019handheld}, whose insight of exploiting aliasing effects has been one of the inspirations of our work. However, due to the sequential nature of their algorithm, errors may propagate from one stage to the next, leading to sub-optimal reconstructions~\cite{park2003super}.  In contrast, iterative spatial domain techniques iteratively refine an estimate for the super-resolved image so as to best explain the observed LR frames under some image formation model. Variants of this approach include the early iterated backprojection algorithm of Irani {\em et al.}~\cite{irani1991improving}, the maximum likelihood technique of Elad and Feuer~\cite{elad1997restoration}, and the model regularized by bilateral total variation of Farsiu {\em et al.}~\cite{farsiu2004fast}.  The image formation parameters are either be assumed to be known a priori through calibration, or estimated jointly with the HR image.  In general, inter-frame motion can either be estimated separately, or be treated as an integral part of the super-resolution problem~\cite{bercea2016confidence,hardie1997joint}, thus avoiding motion estimation between LR frames, whose accuracy may be affected by undersampling~\cite{vandewalle2006super}. The method proposed in the rest of this paper combines the best of both worlds since it performs joint estimation while aligning the LR frames with the reconstructed HR image.

\vsp
\paragraph{Learning-based approaches.}
In this context, the multiframe case has received less attention than its single-image counterpart, for which several loss functions and architectures have been proposed~\cite{dong2015image,lim2017enhanced,zhang2020deep}.  Most multi-frame algorithms focus on video super-resolution. Model-based techniques learn non-uniform interpolation or motion compensation using convolutional neural networks~\cite{shi2016real} but the most successful approaches so far are model free, leveraging instead diversity with 3D convolutions or attention mechanisms~\cite{jo2018deep, wang2019edvr}. Learning-based methods have also been used in remote sensing applications, using 3D convolutions~\cite{molini2019deepsum} or joint registration/fusion architectures~\cite{deudon2020highres} for example. Finally, and closer to our work, Bhat et al.~\cite{bhat2021deep} have recently proposed a network architecture for raw burst super-resolution, together with a very interesting dataset featuring both synthetic and real images for training and testing.  It is important to note that learning-based approaches to super-resolution are typically trained on synthetically generated LR images~\cite{lugmayr2020ntire}, a strategy that may not generalize well to real photographs unless great care is taken in modeling the image corruption process~\cite{brooks2019unprocessing}.  Learning super-resolution models from real LR/HR image pairs is quite challenging since it requires in general using separate cameras with different lenses and spatial resolution, with inevitable spatial and spectral misalignments. As shown by our experiments, our method, although trained from synthetic LR images, gives excellent results with real bursts taken from different smartphones and cameras. Leveraging real images at training time is, for now, left for future work.

\begin{figure}
    \newcommand\x{0.24}
    \hspace{-0.2cm}
    \centering
    \setlength\tabcolsep{0.5pt}
    \renewcommand{\arraystretch}{1}
    \begin{tabular}{cc}
        \includegraphics[width=\x\textwidth]{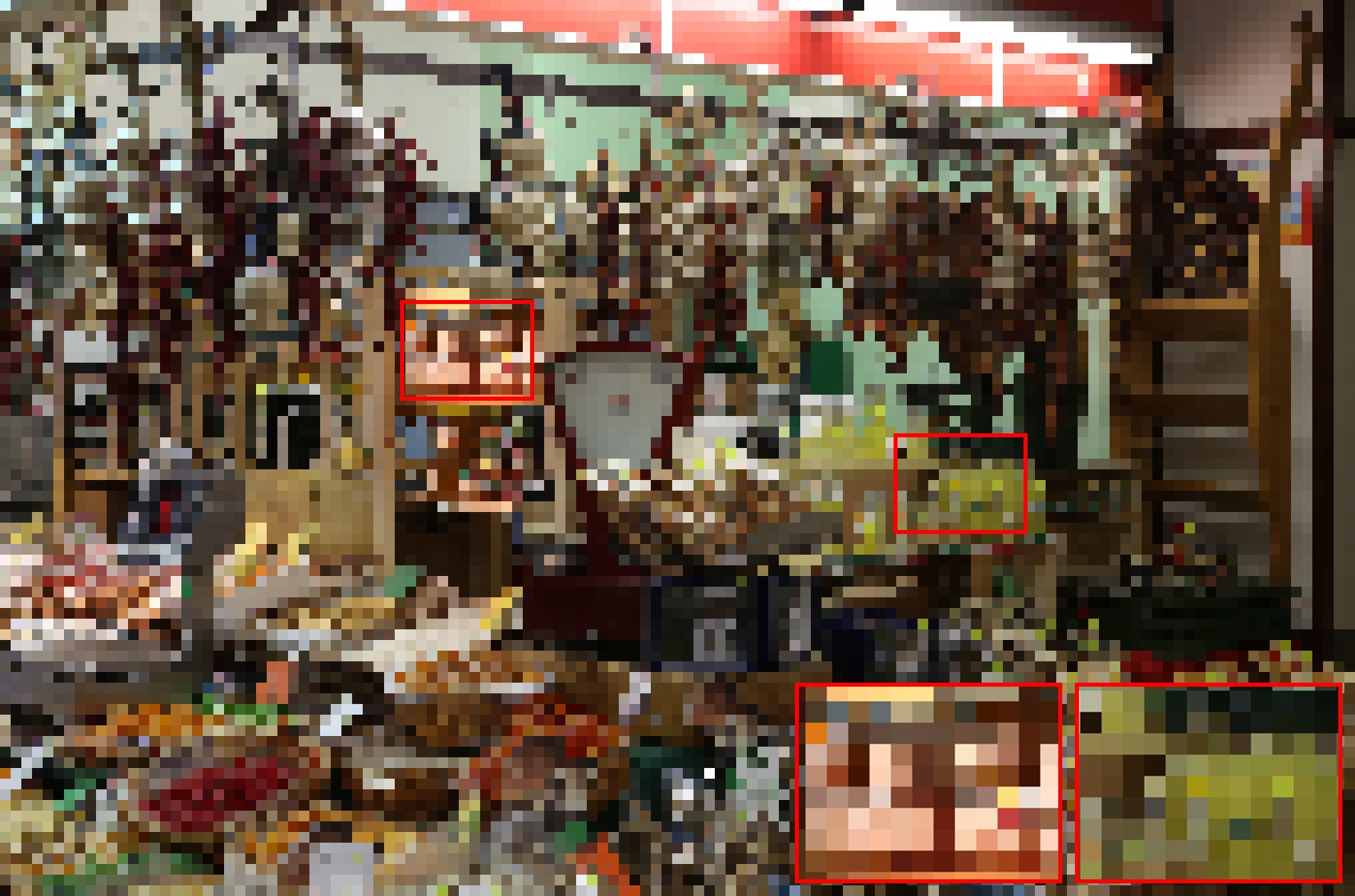}     & 
        \includegraphics[width=\x\textwidth]{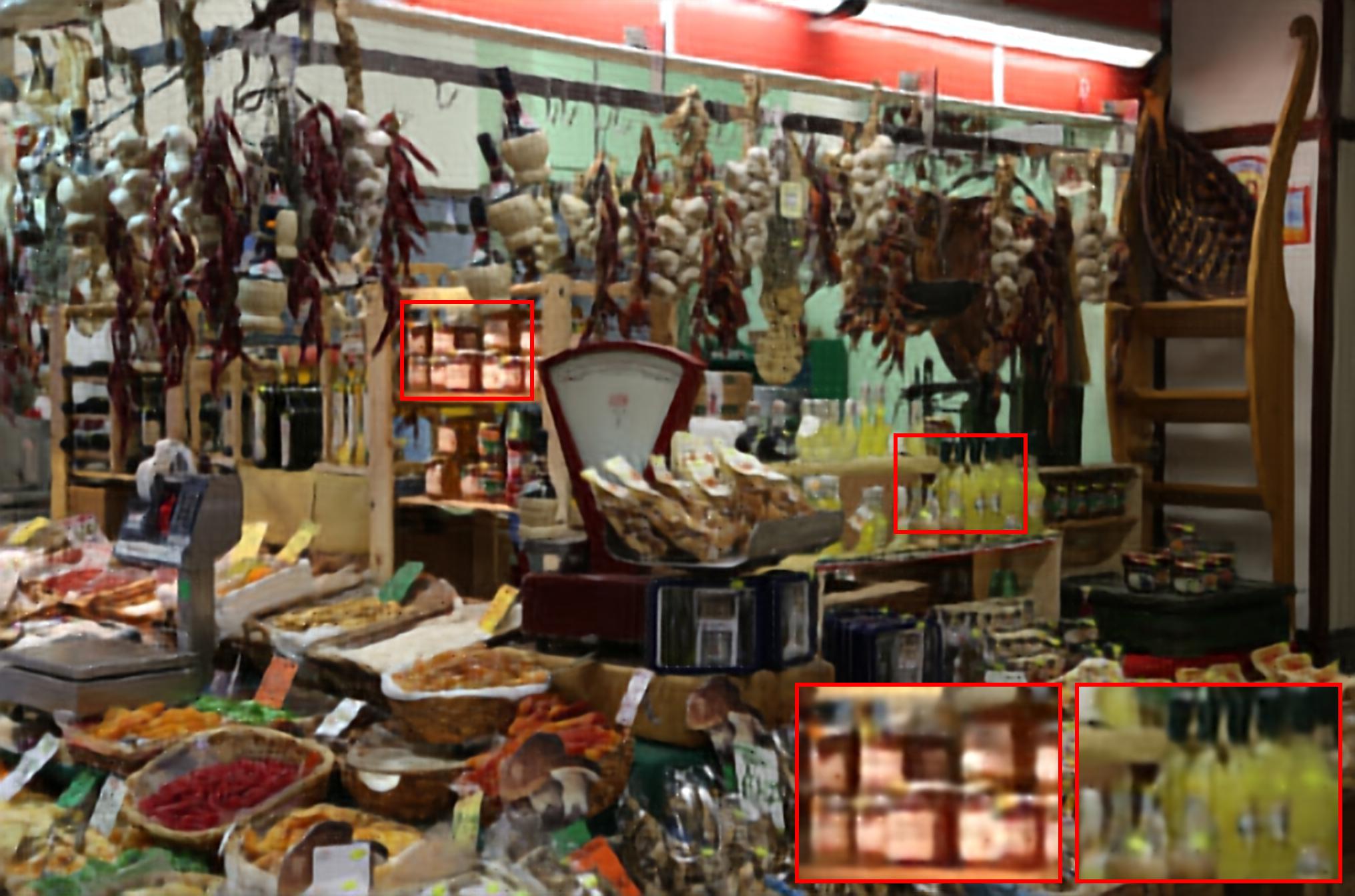}    \\ 
    \end{tabular}
    \caption{{Proof of concept for extreme $\times 16$ upsampling}. The right image is obtained by processing a burst of 20 LR images presented on the left obtained with synthetic random affine movements and bilinear downsampling.}\vspace*{-0.2cm}
    \label{fig:bilinear} 
\end{figure}

\section{Proposed Approach}

\def\texte#1{#1}
\def\RR{\mathbb{R}}
\def\CC{\mathbb{C}}
\def\KK{\mathbb{K}}
\def\NN{\mathbb{N}}
\def\PP{\mathbb{P}}
\def\AA{\mathbb{A}}
\def\LL{\mathbb{L}}
\def\SS{\mathbb{S}}
\def\barr{\bar{\mathbb{R}}}
\def\mat#1{{\mathcal{#1}}}
\def\vect#1{\mbox{\boldmath $#1$}}
\def\PPi{\mbox{\boldmath$\Pi$}}
\def\squig{\rightsquigarrow}
\def\eqdef{\buildrel \rm def \over =}

\def\comment#1{{}}
\def\qmatrix#1{\left[\begin{matrix}#1\end{matrix}\right]}
\def\st#1{{\tt #1}}


\comment{We now introduce our approach. First, we discuss the physics of the data
acquisition process and formulate the corresponding inverse problem.
Second, we propose an optimization procedure, which we turn into
a trainable algorithm with unrolled optimization principles. 
This results in a feedforward architecture, whose steps can be intepreted as
alternatively estimating alignments between images of the burst, and estimating
the high-resolution image. Model parameters can then be 
trained by using backpropagation and stochastic gradient descent,
as a traditional deep neural network.
Finally, we discuss implementation details.}

This section presents the three main components of our approach: its image
formation model, an optimization procedure for solving the
corresponding inverse problem, and its unrolled implementation in a
feedforward architecture whose parameters can be learned end to end.

\comment{First, we formulate the inverse problem corresponding to the
  physics of the data acquisition process.  Second, we propose an
  optimization procedure, which we turn into a trainable algorithm
  with unrolled optimization principles.  This results in a
  feedforward architecture which can be trained end-to-end as a
  traditional deep neural network.}

\subsection{Image formation model}
Image acquisition in a digital camera starts from
an instantaneous irradiance function
$f_{\gamma,t}:[0,1]^2\rightarrow\RR^+$ defined on a continuous retinal domain with
nonnegative values, such that $f_{\gamma,t}(\ub)$ is the
spectral irradiance value at point~$\ub$, time $t$, and wavelength $\gamma$,
accounting for blur
due to optics, atmospheric effects, etc.  The camera sensor 
 integrates~$f_{\gamma,t}$ in the spatial, time, and spectral domains to construct a {\em
  raw} digital image $\yb:[1,\ldots,n]^2\rightarrow\RR^+$, where each
pixel's spectral response is typically dictated by the $2\times 2$
RGGB Bayer pattern, with twice as many measurements for the green
channel than for the red and blue ones~\cite{kimmel1999demosaicing}. Modern cameras turn the
raw image $\yb$ into a full blown, three-channel {\em RGB} image $\xb$
with the same spatial resolution through an interpolation process
called {\em demosaicking}.

\begin{figure}[tb]
  \includegraphics[width=\columnwidth]{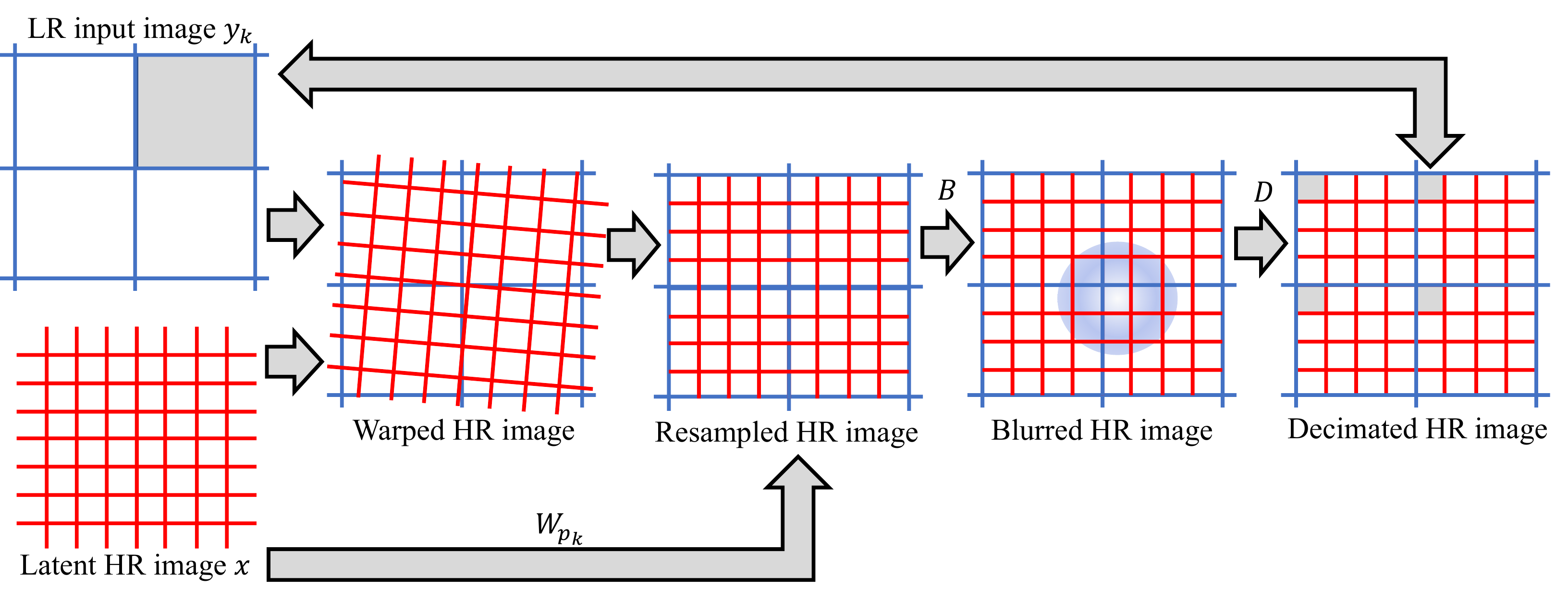}
  \caption{\small Image formation: The HR image $\xb$ is warped then resampled to align it
    with the LR image $\yb$ using the operateur $W_{\pb_k}$. It is then
    blurred by the operator $B$ to account for integration over LR
    pixels and finally downsampled in the spatial and spectral
    domains by the operator~$D$ (the spectral downsampling
    from RGB to R, G or B is not illustrated here for simplicity).}
\label{fig:imfor}
\end{figure}

In practice, we do not have access to $f_{\gamma,t}$ to use as ground
truth for learning an image restoration process, even when an accurate
model of the $f_{\gamma,t} \mapsto \xb$ map is available. Thus, we model instead the process $\xb\mapsto \yb_k$,
where $\xb$ is a latent high-resolution (HR) image we wish to recover,
and the low-resolution (LR) images $\yb_k$ ($k=1,\ldots,K$) have been
observed in a burst of length~$K$. We assume that $\xb$ is sharp,
without any blur, and noiseless. The burst images are obtained through
the following forward model (Figure~\ref{fig:imfor}):
\begin{equation}
  \yb_k=DBW_{\pb_k}\, \xb+\varepsilonb_k \,\,\text{for}\,\,k=1,\ldots,K, 
  \label{eq:imfor}
\end{equation}
where $\varepsilonb_k$ is some additive noise. Here, both the HR
image~$\xb$ and the frames $\yb_k$ of the burst are flattened into
vector form. The operator $W_{\pb_k}$ parameterized by
$\pb_k$ warps $\xb$ to compensate for misalignments between $\xb$ and
$\yb_k$ caused by camera or scene motion between frames, assumed here
to be a 6-parameter affine transformation of the image plane, then
resamples the warped image to align its pixel grid with that of
$\yb_k$. Finally, the corresponding HR image is blurred to account for
integration over both space (the LR pixel area, using either simple averaging or,
as in the figure, a Gaussian filter) and time (accounting for camera
and/or scene motion during exposure), and it is finally downsampled in
both the spatial and spectral domains by the operator $D$, with an (a
priori) arbitrary choice of {\em where} to pick the sample from (pixel
corner or center for example), the spectral part correponding to
selecting one of the three RGB values to assemble the raw image.  It
will prove convenient in the sequel to rewrite~(\ref{eq:imfor}) as
$\yb=U_\pb\xb+\varepsilonb$, where
\begin{equation}
  U_{\pb}\!=\!\qmatrix{DBW_{\pb_1}\\ \vdots\\ DBW_{\pb_K}}\!\!,
  \yb\!=\!\qmatrix{y_1\\ \vdots\\ y_K}\!\!,\pb=\qmatrix{p_1\\ \vdots
    \\p_K}\!\!,
  \varepsilonb\!=\!\qmatrix{\varepsilon_1\\ \vdots\\ \varepsilon_K}\!.
  \label{eq:imforc}
\end{equation}

Before closing this section, let us note that simpler motion models
with two (translation) or three (rigid motion) parameters, or (much)
more complex piecewise-affine or elastic models could 
be considered depending on the application. We focus here on the
scenario where a user wishes to zoom in on a relatively small crop
(say, between $100\times 100$ to $800\times 800$ pixels) of a
multi-megapixel image, and the affine model has proven effective
 with real handheld cameras in this setting. This 
implicitly corresponds to a globally piecewise-affine motion
model. 

\comment{\vsp
\paragraph{Motion models.}
In this paper, we consider two types of motions between images.
Following~\cite{bhat2021deep}, motion may be Euclidean---that is,
given pixel coordinates $a$, new coordinates $b$ are obtained
by rotation $R_\theta$ with angle $\theta$ and translation~$\tau=[\tau_1,\tau_2]$ such that $b = R_{\theta}a + \tau$,
leading to three parameters $\pb = [\theta, \tau_1, \tau_2]$ to estimate.
In order to be more realistic, we also consider more general affine
transformations $a = A b + \tau$, leading this time to six parameters.
Of course, such a parametrization may be two simplistic for aligning large images,
whose motion may be spatially non-uniform. Nevertheless, as shown in our experiments, it turned out to be a
good local model, allowing to align image crops (\eg, $100 \times 100$ up to $800 \times 800$)
obtained from real raw data acquired from handheld digital cameras.}

\subsection{Inverse problem and optimization}
Given the image formation model of
Eqs.~(\ref{eq:imfor})--(\ref{eq:imforc}), 
recovering the HR image $\xb$ from the $K$ LR frames $\yb_k$ in the burst
can be formulated as finding the values of $\xb$ and $\pb$ that minimize
\begin{equation}
 \frac{1}{2}\|\yb-U_\pb\, \xb\|^2+\lambda\phi_\theta(\xb),
  \label{eq:optprob}
\end{equation}
where $\phi_\theta$ is a parameterized regularizer, to be
detailed later, and $\lambda$ is a parameter  balancing the
data-fidelity and regularization terms.  Many methods are of course
available for minimizing this function. Like others (e.g.,
~\cite{krishnan2009fast}), and mainly for simplicity, we choose here a
quadratic penalty method~\cite[Sec. 17.1]{nocedal} often called
half-quadratic splitting (or HQS)~\cite{geman1995nonlinear}: the
original objective is replaced by
\begin{equation}
  E_\mu(\xb,\zb,\pb)= \frac{1}{2}\|\yb-U_\pb\,
  \zb\|^2+\frac{\mu}{2}\|\zb-\xb\|^2
  +\lambda \phi_\theta(\xb),
  \label{eq:HQS}
\end{equation}
where $\zb$ is an auxiliary variable, and $\mu$ is a parameter
increasing at each iteration, such that, as $\mu\rightarrow+\infty$,
the minimization of (\ref{eq:HQS}) with respect to $\xb$, $\zb$ and
$\pb$ becomes equivalent to that of (\ref{eq:optprob}) with respect to
$\xb$ and $\pb$ alone. Each iteration of HQS can be viewed as one step
of a block-coordinate descent procedure for minimizing $E$, changing
one of variables $\zb$, $\xb$ and $\pb$ at a time while keeping the
others fixed, with the value of $\mu$ increasing after each
iteration.  Convergence guarantees for quadratic penalty methods
require an approximate minimization of Eq.~(\ref{eq:HQS}) with
increasing precision over time~\cite{nocedal}. Following common
practice in computer vision (e.g.~\cite{krishnan2009fast}), we use HQS
without formally checking that its precision indeed increases with
iterations.
This very
simple procedure turns out to work well in practice. Its steps
are detailed in the next three paragraphs, the exponent $t$ being used
to designate the value of the variables at iteration $t$. The sequence
of weights $(\mu^{t})_{t \geq 0}$ is 
learned end-to-end as explained
in Section~\ref{subsec:unroll}.

\vsp
\paragraph{Updating $\zb$.}
Several strategies are possible for minimizing Eq.~(\ref{eq:HQS}) with
respect to $\zb$. Given the dimension of the problem, one may choose
for instance a fast iterative minimization procedure such as conjugate
gradient descent. Since an approximate minimization is sufficient for
our needs, we have chosen to use instead a single step of plain
gradient descent, which converges more slowly in theory, but is also
simpler and more easily amenable to the unrolled optimization strategy
for end-to-end learning that will be presented next. The update at
iteration $t$ is given by
\begin{equation}
  \zb^t\leftarrow
  \zb^{\tmone}-\eta_t\big[U_{\pb^{\tmone}}^\top(U_{\pb^{\tmone}}\zb^{\tmone}-\yb)+\mu(\zb^{\tmone}-\xb^{\tmone})\big],
\end{equation}
where $\eta_t > 0$ is some step size, also learned end to end.

\vsp
\paragraph{Updating the motion parameters $\pb$.}

Let $\pb_k$ denote the part of the parameter vector $\pb$ responsible for
the alignment of $\zb^t$ and $\yb_k$ in (\ref{eq:HQS}). The
corresponding
optimization problem can be rewritten as
\begin{equation}
   \min_{\pb_k}\frac{1}{2} \|\yb_k- D B W_{\pb_k} \zb^t\|^2.\label{eq:nonlin}
\end{equation}
This is a non linear least-squares problem, which can once again be
solved using many different techniques. Here, we pick a Gauss-Newton
approach, which corresponds to a variant of the Lucas-Kanade
algorithm~\cite{baker2004lucas,lucas1981iterative}, showing again that
a 40-year old technique can still be relevant today.  Specifically, we
perform one Gauss-Newton step at each iteration~$t$ for each $\pb_k$
in parallel:
\begin{equation}
  \pb_k^t \leftarrow  \pb_k^{\tmone} -  \left( \Jb_k^{t\top} \Jb_k^t \right)^{-1} \Jb_k^{t\top} \rb_k^t,  \label{eq:LK}
\end{equation}
where $\rb_k^t = U_{\pb_k^\tmone} \zb^t - \yb_k$ is the residual of
the non-linear least-squares problem~(\ref{eq:nonlin}), and
$\Jb_k^t = ({\partial U_{\pb_k^\tmone}}/{ \partial \pb_k})\zb^t$ is
the Jacobian of the $DBW_{\pb_k}$ operator.
The only difference with a Lucas-Kanade iteration is the presence of a
high-resolution frame~$\zb^t$ and the downsampling operator $DB$.
This is similar to~\cite{hardie1997joint}, or more
recently~\cite{bercea2016confidence,he2007nonlinear}, which align
high-resolution images with low-resolution ones.

\vsp
\paragraph{Estimating the HR image $\xb$.}
The $\xb$ update is obtained as
\begin{displaymath}
   \xb^t \leftarrow \argmin_{\xb} \frac{\mu_{\tmone}}{2}\|\zb^t-\xb\|^2+\lambda\phi_\theta(\xb),
\end{displaymath}
which amounts to computing the proximal operator of the prior
$\phi_\theta$.  In practice, we follow a ``plug-and-play''
approach~\cite{chan2016plug,ryu2019plug,venkatakrishnan2013plug}, and
replace the proximal operator by a parametric function
$f_\theta(\zb_t)$ (here, a CNN, see implementation details).  Using
such an implicit prior has proven very effective in our setting.
More traditional image priors such as total variation could of course
have been used as well.

\subsection{Unrolled optimization and backpropagation}\label{subsec:unroll}
The optimization procedure described so far requires choosing
hyper-parameters such as the sequence $(\mu_t)_{t\geq 0}$, and its
implicit prior also involves model parameters~$\theta$.  By using a
training set of $n$ LR burst/HR image pairs, we propose to learn all
these parameters in a supervised fashion.  We denote the training set
by $\left( \Yb_i , \xb_i\right)_{i=1}^n$, where
$\Yb_i =\{ \yb_j^i\}_{j=1}^K$ is the $i$-th burst of LR images
associated to the HR image~$\xb_i$.  We then unroll the optimization
procedure for $T$ steps and, denoting by $\hat{\xb}_T(\Yb_i)$ the
HR image estimated from burst $\Yb_i$, we
consider the objective function
\begin{equation}
   \frac{1}{n} \sum_{i=1}^n L(\hat{\xb}_T(\Yb_i) , \xb_i ), \label{eq:loss}
\end{equation}
where $L$ is the $\ell_2$ or $\ell_1$ loss (in
practive we have observed that the  $\ell_1$ loss performs slightly better).
Because every step of our estimation procedure is differentiable, we minimize~(\ref{eq:loss}) by stochastic gradient descent.

\vsp
\paragraph{Learned data prior.}
Good image priors are essential for solving ill-posed inverse
problems.  As noted earlier, instead of using a classical one, such
as total variation (TV) or bilateral total variation
(BTV)~\cite{farsiu2004fast}, we learn an implicit prior parameterized
by a convolutional neural network $f_\theta$ in a data-driven
manner.  We use the ResUNet architecture introduced in~\cite{zhang2020deep} in
practice. It involves four scales, each of which has an identity skip
connection between downscaling and upscaling operations.

\subsection{Implementation details and variants}\label{subsec:details}
\paragraph{Downsampling and blurring operators $\Db,\Bb$.}
We have tried different variants of downsampling/blurring strategies
such as Gaussian smoothing.  In practice, we have observed that
simple averaging, which is differentiable and
parameter-free, gives good results in all our experiments.
{As a consequence, we do not assume any knowledge about the blur used to
generate data, corresponding to an operator $B$ that only captures
blur due to photon integration on the sensor without
addressing optical blur. We argue that this
limited model is relevant because modern cameras and
smartphone are aliased \cite{wronski2019handheld}, which may explain the generalization to real images, as soon as
the scene is static.}
 

\vsp
\paragraph{Initialization by coarse alignement.}
To initialize the motion parameters~$\pb$, we cannot
minimize~(\ref{eq:nonlin}) as in the previous section, because no good estimate of the HR image is available. 
Therefore, we align each LR frame to an arbitrary one from the burst (\eg, the first one)
by using the Lucas-Kanade forward additive algorithm \cite{baker2004lucas,sanchez2016inverse} which is known to be robust to noise.
Note that another difficulty lies in the raw format of images. To
overcome this issue, we simply convert raw images into grayscale images by
using bilinear interpolation.  This is of course sub-optimal, but sufficient
for obtaining coarse motion parameters. 

\vsp
\paragraph{Initialization via coarse-to-fine strategy.}
For extreme upsampling factors ($\times 16$), we found a coarse-to-fine
initialization strategy to be useful: We
initialize the motion parameters $\pb_j^0$ and high-resolution image $\zb^0$ 
by using the output of the algorithm trained at a lower upsampling factor.
For instance, $\times 16$ can be obtained by applying twice a $\times 4$ algorithm, or
four times $\times 2$ algorithm.

\section{Experiments}\label{sec:exp}

Experiments were conducted on synthetic and real raw image bursts. We also provide experiments on RGB bursts in the appendix, allowing easier comparison with earlier approaches that cannot handle raw data.
 
\vsp
\paragraph{Training procedure and data.}
For synthetizing realistic \textit{raw bursts} from groundtruth RGB images, we follow
the approach described in \cite{bhat2021deep}, using the author's publicly available  
code\footnote{\url{https://github.com/goutamgmb/NTIRE21_BURSTSR}.} on the
training split of the Zurich raw to RGB dataset~\cite{ignatov2020replacing}. The
approach consists of applying the inverse RGB to raw pipeline introduced in
\cite{brooks2019unprocessing}. Displacements are randomly generated with
Euclidean motions and frames are downscaled with bilinear interpolation in
order to simulate LR frames containing aliasing. Synthetic, yet
realistic, noise is added to the frames, and color values are
discarded according to the Bayer pattern. 
Then, we train our models for minimizing the loss~(\ref{eq:loss}). 
We perform $100\,000$ iterations of the ADAM optimizer with a batch size of $10$, a burst size of $14$ and
with a learning rate of $3\times 10^{-5}$ decaying by a factor $2$ after
$50\,000$ iterations. Our approach is implemented in Pytorch and takes approximately $1.5$ days to train on an Nvidia Titan RTX GPU.  
We evaluate our models in all our experiments with a burst size of $14$ unless specified.

\vspace*{-0.2cm}
\paragraph{Extreme $\times 16$ upsampling on RGB images.}
As a proof-of-concept, we also perform experiments for an unusual $\times 16$
super-resolution task, using the coarse-to-fine strategy of Sec.~\ref{subsec:details}. A result is presented in Fig.~\ref{fig:bilinear}, showing impressive reconstruction and additional ones can be found in the appendix.
Even though not realistic, we believe the experiment
to be of interest, as it demonstrates the effectiveness of our approach in an
idealistic, yet extreme, setting.

\newcommand\w{0.16}
\begin{figure}[ht]
    \setlength\tabcolsep{0.5pt}
    \renewcommand{\arraystretch}{0.5}
    \centering
    \begin{tabular}{ccc}
        \includegraphics[width=\w\textwidth]{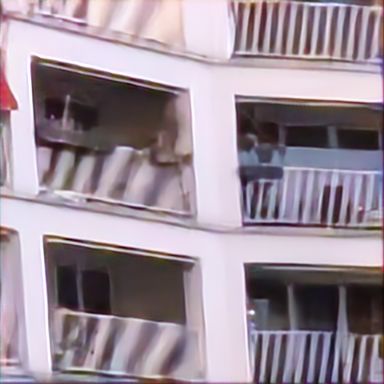}        & 
        \includegraphics[width=\w\textwidth]{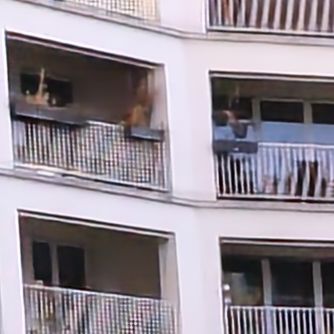}         & 
        \includegraphics[width=\w\textwidth]{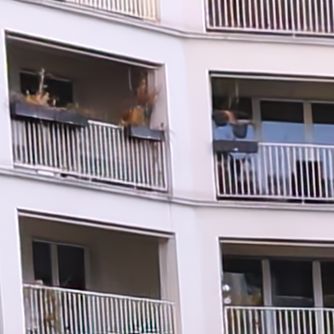}                                                                                                                                \\

        \includegraphics[width=\w\textwidth]{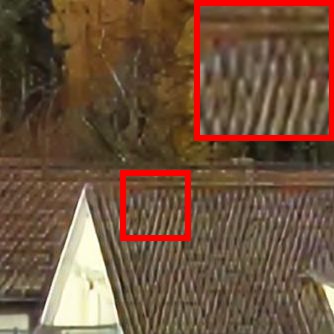}         & 
        \includegraphics[width=\w\textwidth]{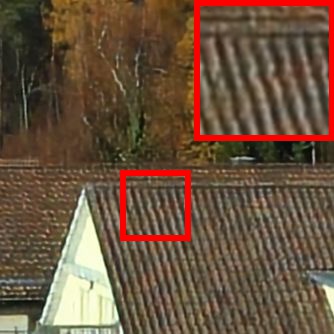}          & 
        \includegraphics[width=\w\textwidth]{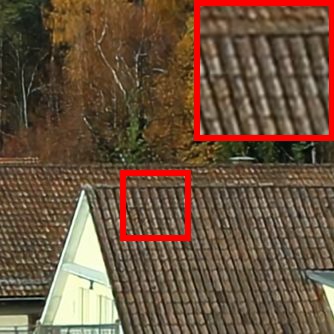}                                                                                                                                 \\
       \begin{tabular}{c} Demosaic+SISR \end{tabular} & ETH \cite{bhat2021deep} & Ours 
    \end{tabular}
    \caption{Visual comparison on \textbf{synthetic raw image bursts} used in~\cite{bhat2021deep}. Demosaic+SISR is our single-image baseline based on the ResUNet architecture~\cite{zhang2020deep} (see main text). The two right columns are produced by methods dedicated to raw burst processing, respectively~\cite{bhat2021deep} and ours.
    }\label{fig:raw}
\end{figure}

\paragraph{Evaluation on synthetic RAW images.}
The evaluation protocol of~\cite{bhat2021deep} allows us to perform quantitative
comparison with their state-of-the-art method for processing raw image bursts.  An additional
comparison with~\cite{wronski2019handheld} would have been interesting but
this method is part of a commercial product that could not be shared with us.

\begin{table}[h]
    \small
    \centering
    \begin{tabular}{llcc}
        \toprule
        Method                  & PSNR (db)      & Geom (pix) & SSIM  \\
        \midrule
        \multicolumn{4}{l}{\textit{Scores on public validation set}}  \\
        ETH \cite{bhat2021deep} & 39.09          & -          & -     \\
        Ours (refine)           & \textbf{41.45} & -          & 0.95  \\
        \midrule
        \midrule
        \multicolumn{4}{l}{\textit{Scores on our own validation set to conduct the ablation study}} \\
        Bicubic Single Image &  33.45 & - & - \\
        Multiframe L2 only & 34.21  & - &-\\
        Multiframe L2 + TV prior & 34.48 & -&- \\ 
        Single Image            & 36.80          & -          & -     \\
        Ours (no refinements)   & 40.38          & 0.55       & 0.958 \\
        Ours (refinements)      & \textbf{41.30} & 0.32       & 0.963 \\        
        \midrule
        Ours  (known motion)    & 42.41          & 0.00       & 0.971 \\        
        \bottomrule
    \end{tabular}
    \caption{\textbf{Results with synthetic raw image bursts} of 14 images generated from the Zurich raw to RGB dataset \cite{ignatov2020replacing} with synthetic affine motions. Reconstruction error in average PSNR and geometrical registration error in pixels for our models. ``known $\pb$'' is the oracle performance our model could achieve, if motion estimation was perfect.}
    \label{tab:my_label}\label{tab:raw}
\end{table}

We provide a quantitative comparison in Table \ref{tab:raw} with the model
introduced in \cite{bhat2021deep}, as well as a single-image upsampling
baseline based on the ResUNet architecture~\cite{zhang2020deep}, which we use
as a plug-and-play prior in our model.

To that effect, we first use the validation set of~\cite{bhat2021deep}
available online (with no overlap with the training set), for which motions are
unknown, allowing us to compare with their method, which we outperform by more
than 2dBs.
In order to perform further comparison and conduct the ablation study, 
we also build an additional validation set by randomly extracting 266 images
from the Zurich raw to RGB dataset, allowing us to generate validation data with 
known motion. 
We evaluate variations of our model in the same table, notably comparing the registration accuracy achieved by these variants by using the geometrical error presented in~\cite{sanchez2016inverse}.
More precisely, we perform a small ablation study by introducing a simpler baseline that does not perform joint alignement and only exploits the coarse registration module (no refine baseline). 
Performing motion refinement significantly improves the registration accuracy and subsequently the image reconstruction quality.
Last, we also report the oracle performance of our model with known motions. 
 
We provide a visual comparison in Figure  \ref{fig:raw}  with
single-image SR baselines and the state-of-the-art
method~\cite{bhat2021deep} for processing raw image bursts.
Only the two approaches processing bursts are able to recover
high-frequency details, demonstrating their ability to leverage and remove
aliasing artefacts, which are very present in the top image.
Significantly better quality results are obtained with our approach.

\begin{figure}[h!]
    \advance\leftskip-0.3cm
    \includegraphics[width=0.50\textwidth]{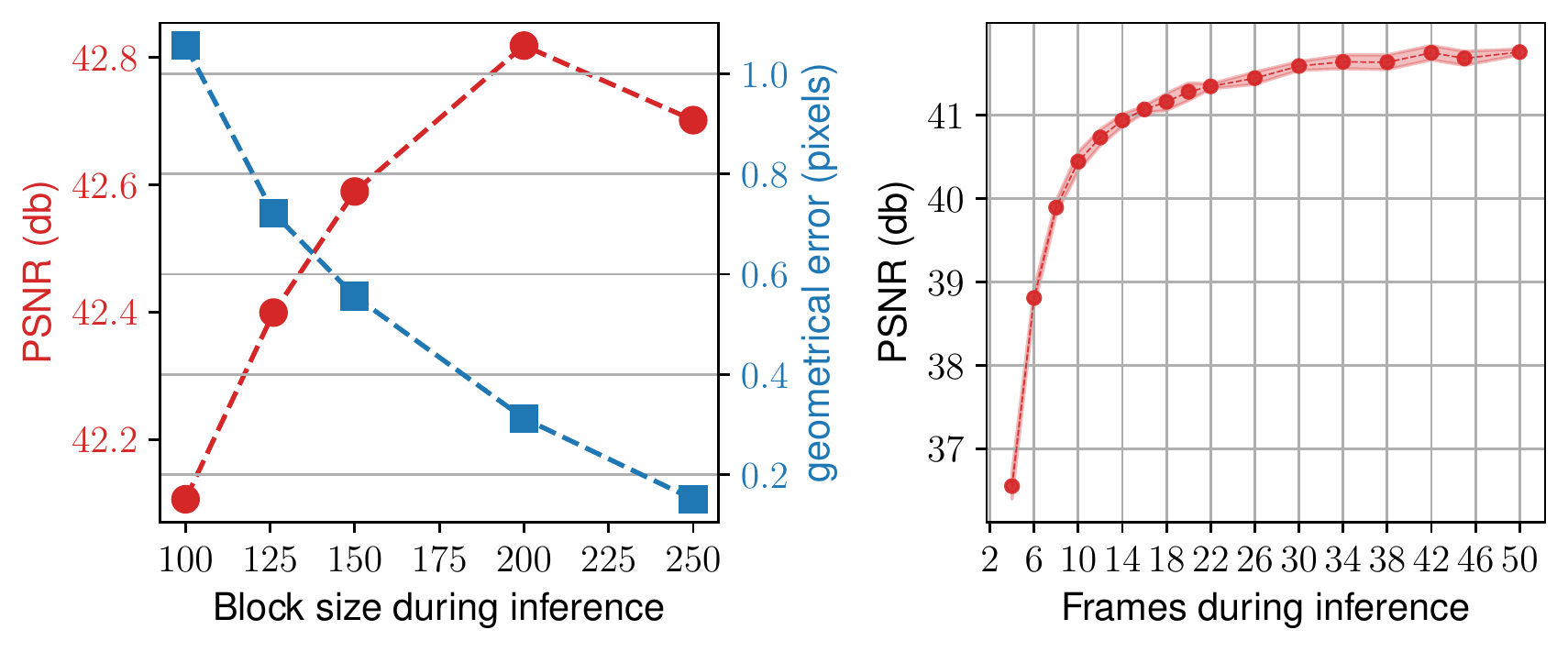}
    \caption{Left: Impact of the crop size on the registration and reconstruction performance. Right: Impact of the burst length, see main text for details.}
    \label{fig:crop_size}
    \vspace{-0.5cm}
\end{figure}

\vspace*{-0.2cm}
\paragraph{Impact of burst length and cropping size.}
The dataset Zurich rgb-to-raw~\cite{ignatov2020replacing} was very useful for training our
models, but it unfortunately features relatively small image crops of size $96
\times 96$ without giving access to the original megapixel images. By
experimenting with real raw data, it became apparent to us to our method was
performing better with larger crops (\eg, more than $200 \times 200$ pixels),
achieving better registration and visually better results.
To study the impact of the crop size and burst length, we have thus
synthesized additional raw bursts from the DIV2K dataset,
and report our experimental results in Figure~\ref{fig:crop_size},
confirming our findings.
Note that this does not appear to be a strong limitation of our approach, since in real-life
scenarios, we can always assume that the original megapixel image is available.
As expected, the performance of our approach is also increasing with the burst size,
even though our models were trained with bursts of size $14$.

\begin{figure}[h]
\small
\centering
\renewcommand{\arraystretch}{0.5}
\setlength\tabcolsep{0.5pt}
\begin{tabular}{ccc}
   \begin{overpic}[width=0.16\textwidth,tics=10,trim={6cm 6cm 0 0},clip]{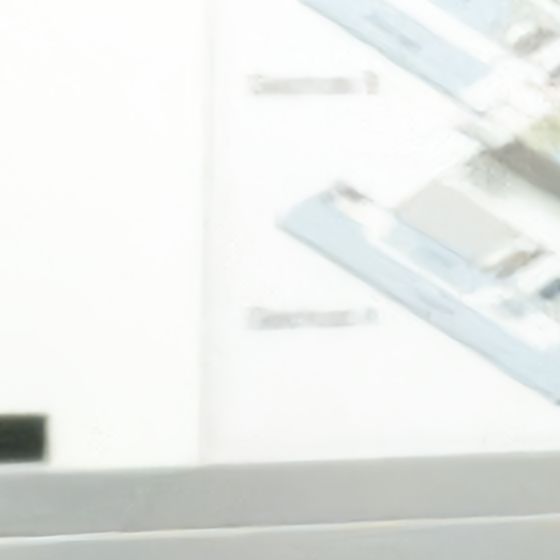}
        \put (0,90) {\small \textcolor{black}{47.49db}}
        \end{overpic}  &
\begin{overpic}[width=0.16\textwidth,tics=10,trim={6cm 6cm 0 0},clip]{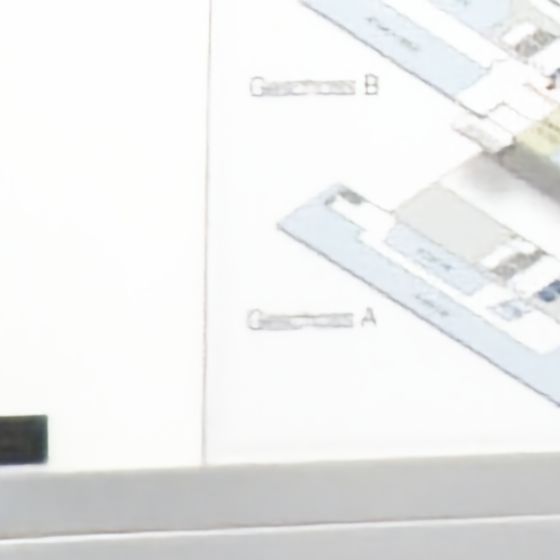}
\put (0,90) {\small \textcolor{black}{51.04db}}
\end{overpic}  &
\begin{overpic}[width=0.16\textwidth,tics=10,trim={6cm 6cm 0 0},clip]{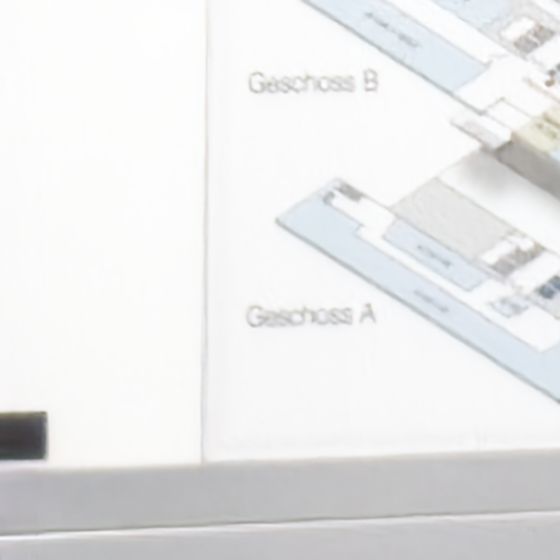}
\put (0,90) {\small \textcolor{black}{49.51db}}
\end{overpic}   \\
   \begin{overpic}[width=0.16\textwidth,tics=10,trim={0 5cm 5cm 0},clip]{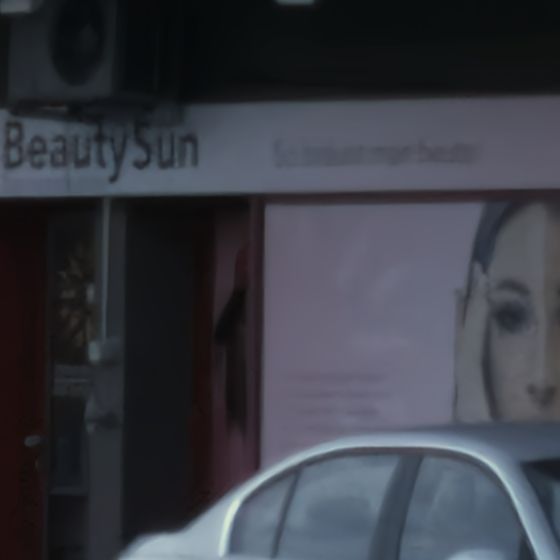}
    \put (0,90) {\small \textcolor{white}{45.13db}}
    \end{overpic}   &
\begin{overpic}[width=0.16\textwidth,tics=10,trim={0 5cm 5cm 0},clip]{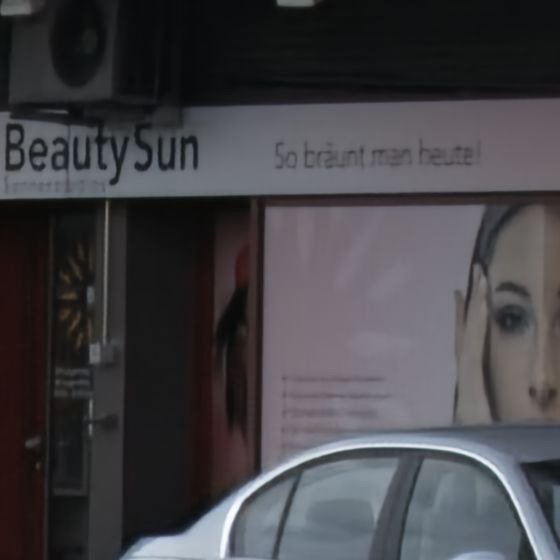}
    \put (0,90) {\small \textcolor{white}{49.66db}}
    \end{overpic}   &
\begin{overpic}[width=0.16\textwidth,tics=10,trim={0 5cm 5cm 0},clip]{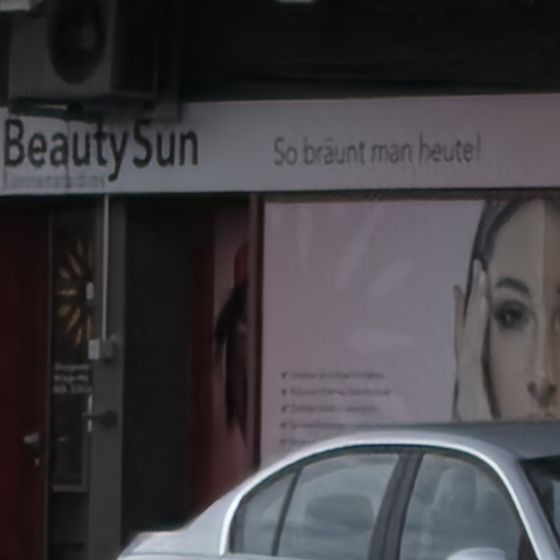}
\put (0,90) {\small \textcolor{white}{48.29db}}
\end{overpic}  \\
Single-image SR & ETH\cite{bhat2021deep} & Ours  
\end{tabular}
    \caption{\textbf{Results from real raw bursts} from dataset of \cite{bhat2021deep} including Aligned PSNR score (see main text).}\label{fig:real_eth}
\end{figure}

\vspace*{-0.2cm}
\paragraph{Results on real raw image bursts, dataset of~\cite{bhat2021deep}.}
In Figure~\ref{fig:real_eth}, we show a comparison with~\cite{bhat2021deep}
using their dataset featuring small crops of size $96 \times 96$. As discussed
previously, this setup is suboptimal for our approach, but still produces
visually pleasant results. Choosing which method performs best here is however
very subjective and we found conclusions hard to draw on this dataset. Whereas
the images produced by~\cite{bhat2021deep} may sometimes look slightly sharper,
one may argue that our approach seems to recover more reliable details, \eg,
the text is perhaps easier to read. 
Note that our models were trained on synthetic data only and we leave fine-tuning with real data
on this dataset for future work.
%
There is an attempt in \cite{bhat2021deep} to address the open problem of
quantitative evaluation with real data using a custom metric,
but, like any other attempt so far, it is flawed since (i) it is
based on the alignment method of \cite{bhat2021deep}, with an unavoidable
slight bias in its favor, and (ii) it assumes ground truth from a
particular Canon camera. 
Interestingly, this
score improvement does not always correlate with visual quality,
as shown by Figure 6. This is by no means a criticism of
[3]: we believe instead that quantitative evaluation on real
images is an extremely challenging problem, far from being
solved.
Since the submission of our paper, the results of the NTIRE 2021 burst
super-resolution challenge have been published \cite{bhat2021ntire}.  Our method ranked
third quantitatively in the "synthetic data" part of the challenge that
we entered.

\vspace*{-0.2cm}
\paragraph{Results on real raw image bursts from various devices.}
Finally, we demonstrate the effectiveness of our approach on 
real raw bursts acquired by different devices. We consider a Panasonic Lumix GX9 camera,
which is interesting for SR as it does not feature an optical anti-aliasing
filter, a Canon Powershot G7X camera, a Samsung S7 and a
Pixel 4a smartphones.
Results obtained in high noise regimes have already been presented in Figure~\ref{fig:teaser}, showing that our approach is surprisingly robust to noise. 
We believe that the result is of interest since it may allow photographers to use
high ISO settings in low-light conditions, without sacrificing image quality. 
Other results are presented in Figure~\ref{fig:real_eth} on low-noise outdoor conditions with bursts
of 20 to 30 raw images. In all cases, the method succeeds at recovering high-frequency details.
{Many more examples and comparisons with other multiframe methods are provided in the supplementary material. We also present failure cases, corresponding in large parts to scene motion.}
Last, we remark that our method is relatively fast at inference time.
Processing a burst of 20 raw $300 \times 300$ images takes for instance about
$1s$ on an Nvidia Titan RTX GPU, producing an upsampled image of size $1200 \times 1200$.







\begin{figure*}[h!]
    \vspace{-0.5cm}
    \renewcommand\w{0.28}
    \renewcommand{\arraystretch}{1}
    \centering
    \setlength\tabcolsep{0.2pt}
    \begin{tabular}{ccccccc}
 
        \parbox[t]{4mm}{\rotatebox[origin=l]{90}{Panasonic}} & 
        \includegraphics[width=0.12\textwidth]{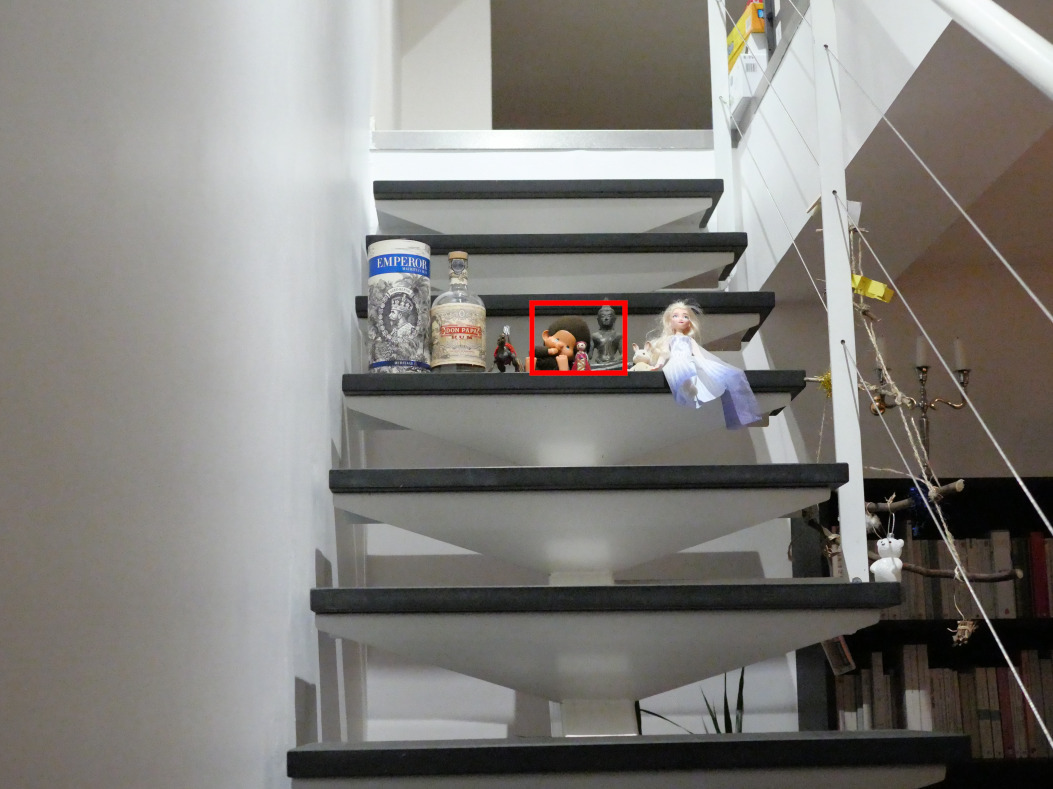}     & 
        \includegraphics[width=\w\textwidth]{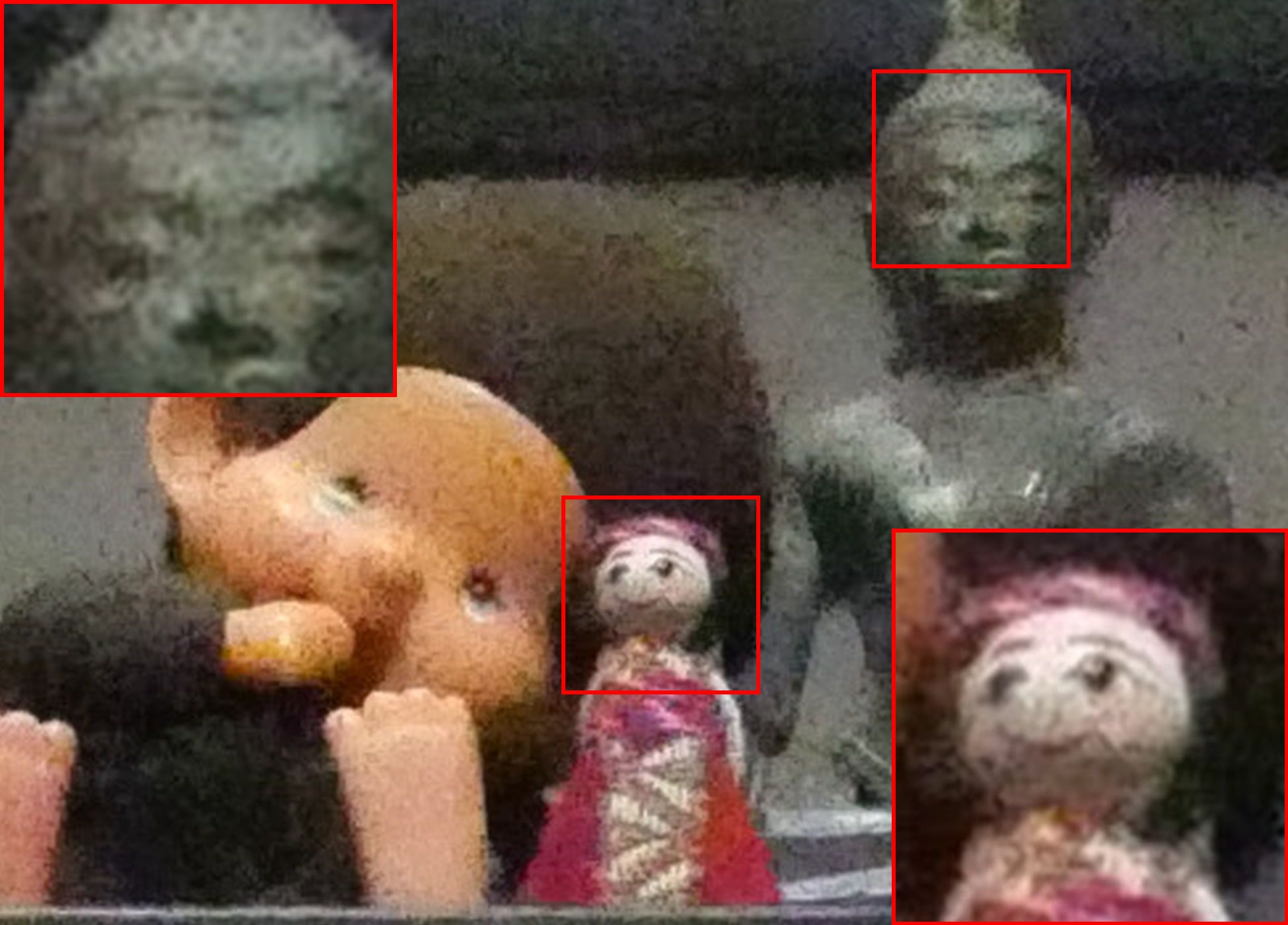}     & 
        \includegraphics[width=\w\textwidth]{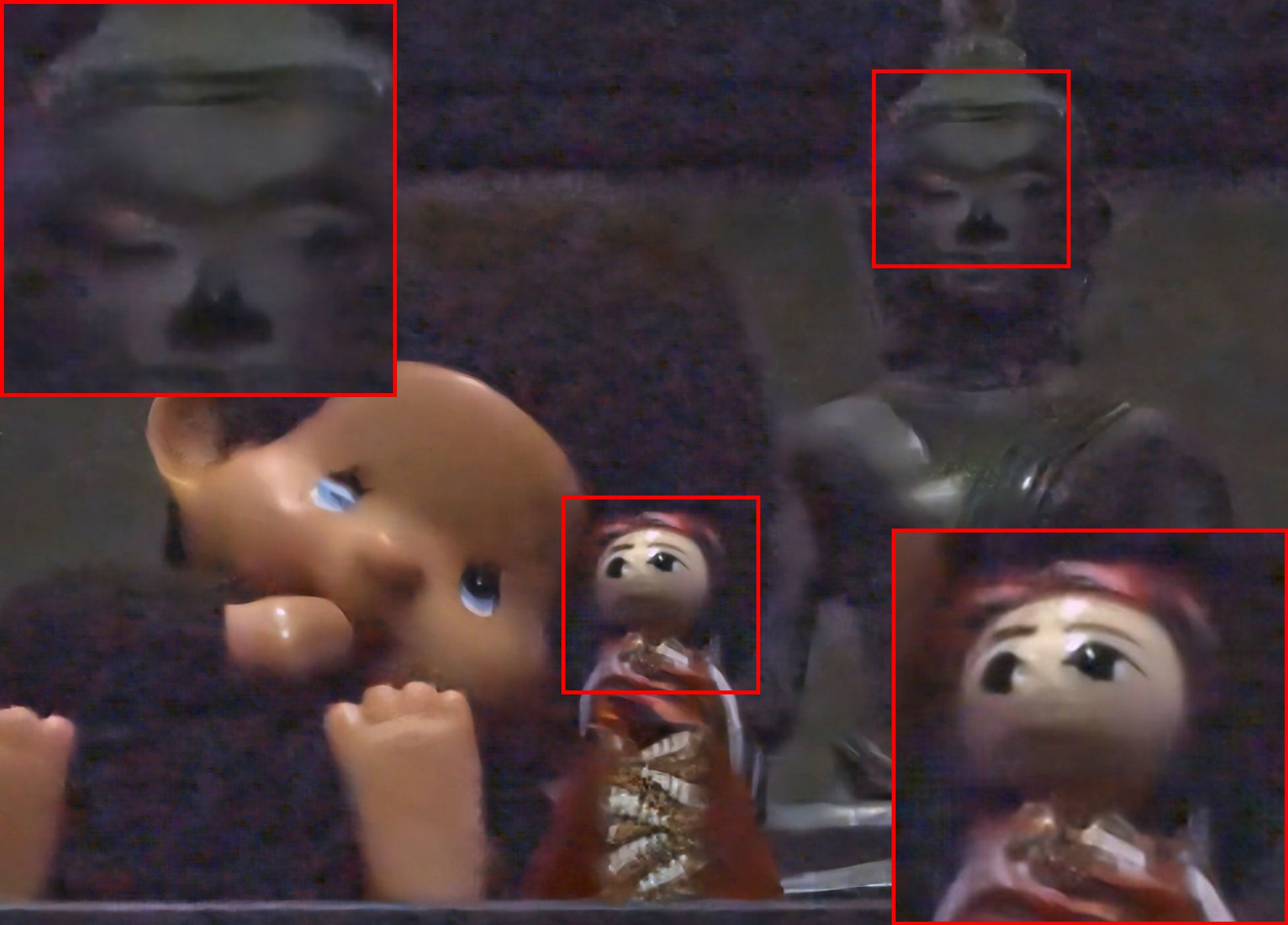}     & 
        \includegraphics[width=\w\textwidth]{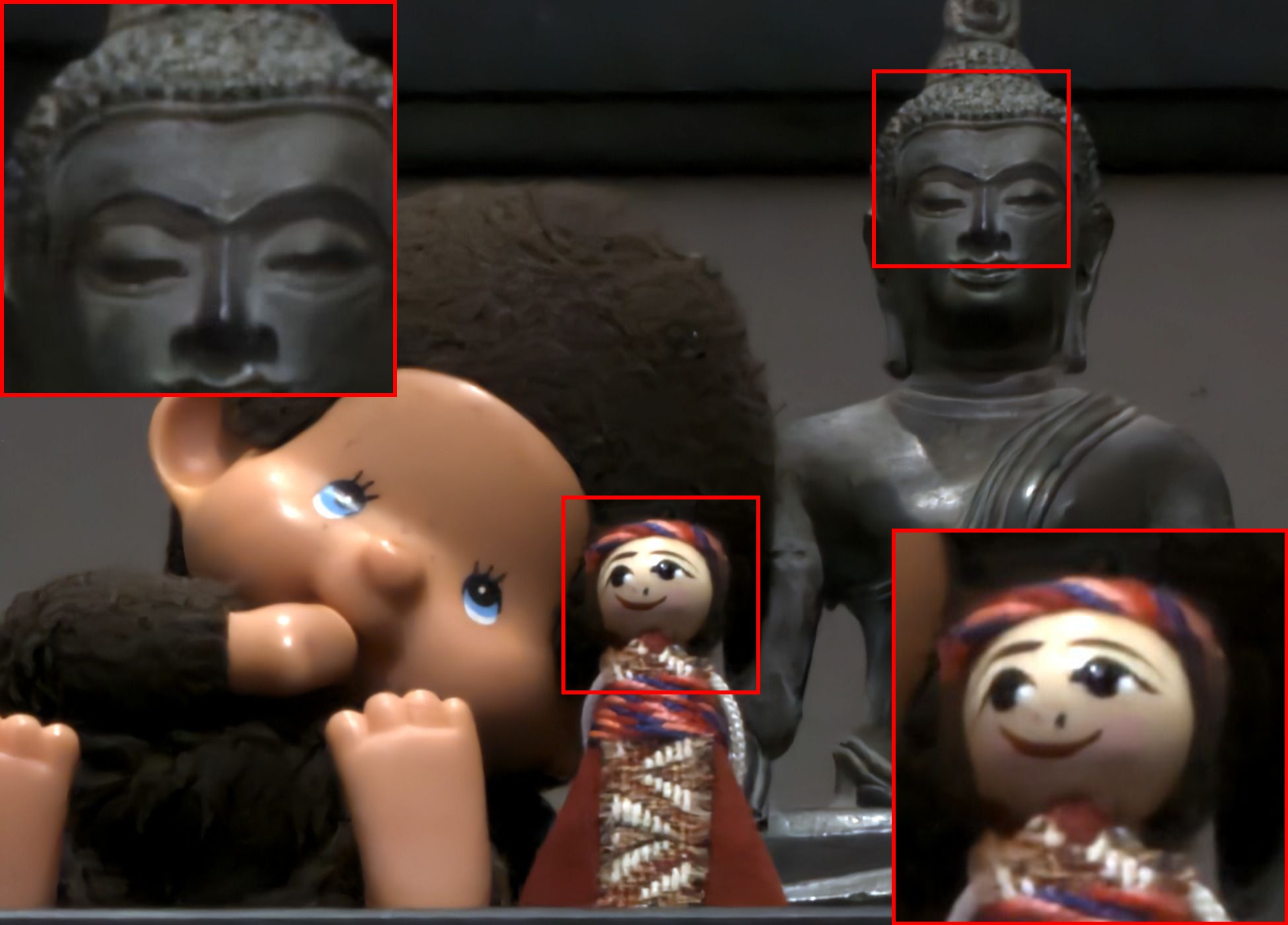}     \\ 
  
        \parbox[t]{4mm}{\rotatebox[origin=l]{90}{Panasonic}} & 
        \includegraphics[width=0.12\textwidth]{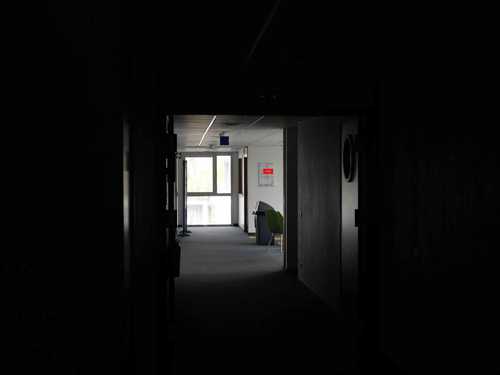}     & 
        \includegraphics[width=\w\textwidth]{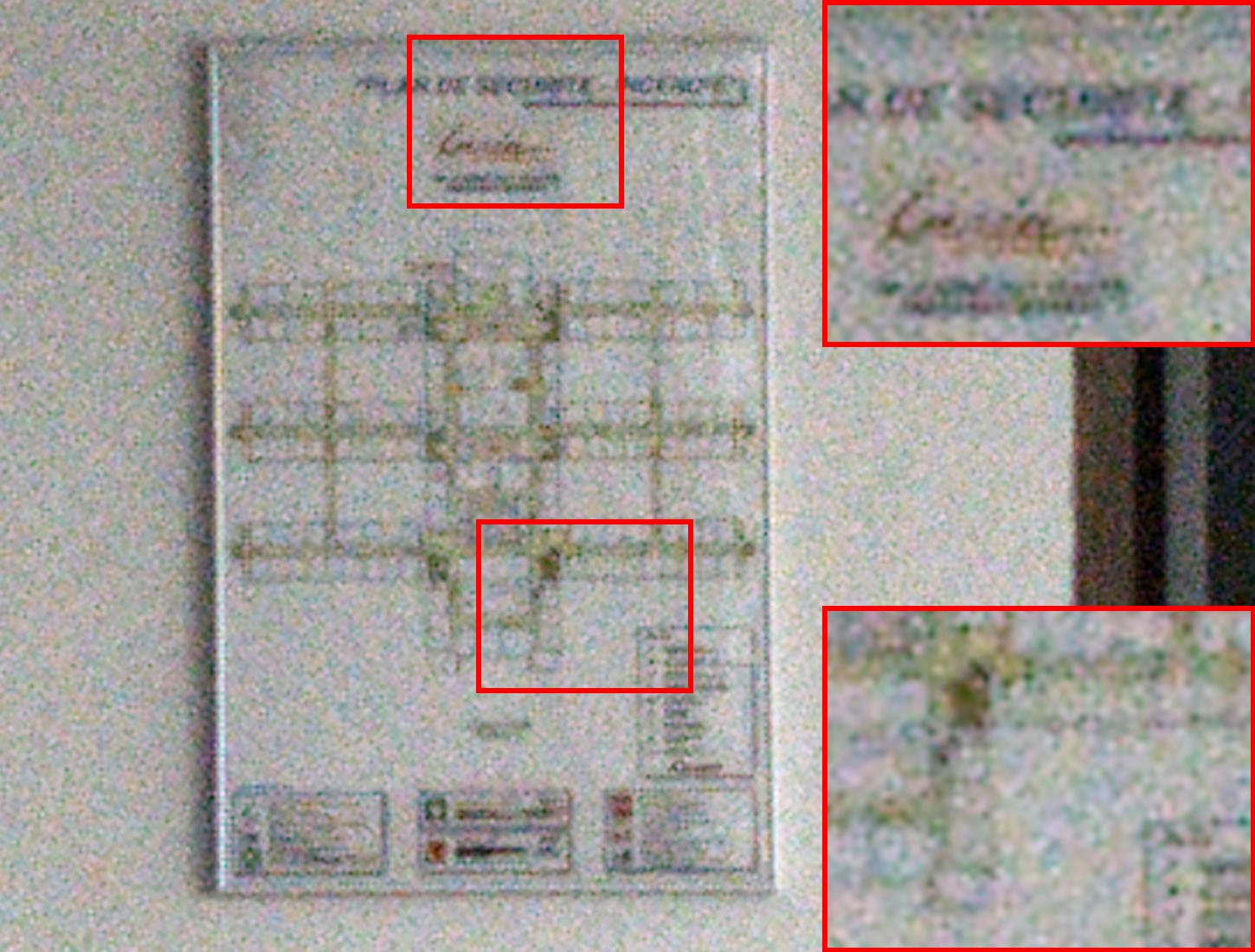}     & 
        \includegraphics[width=\w\textwidth]{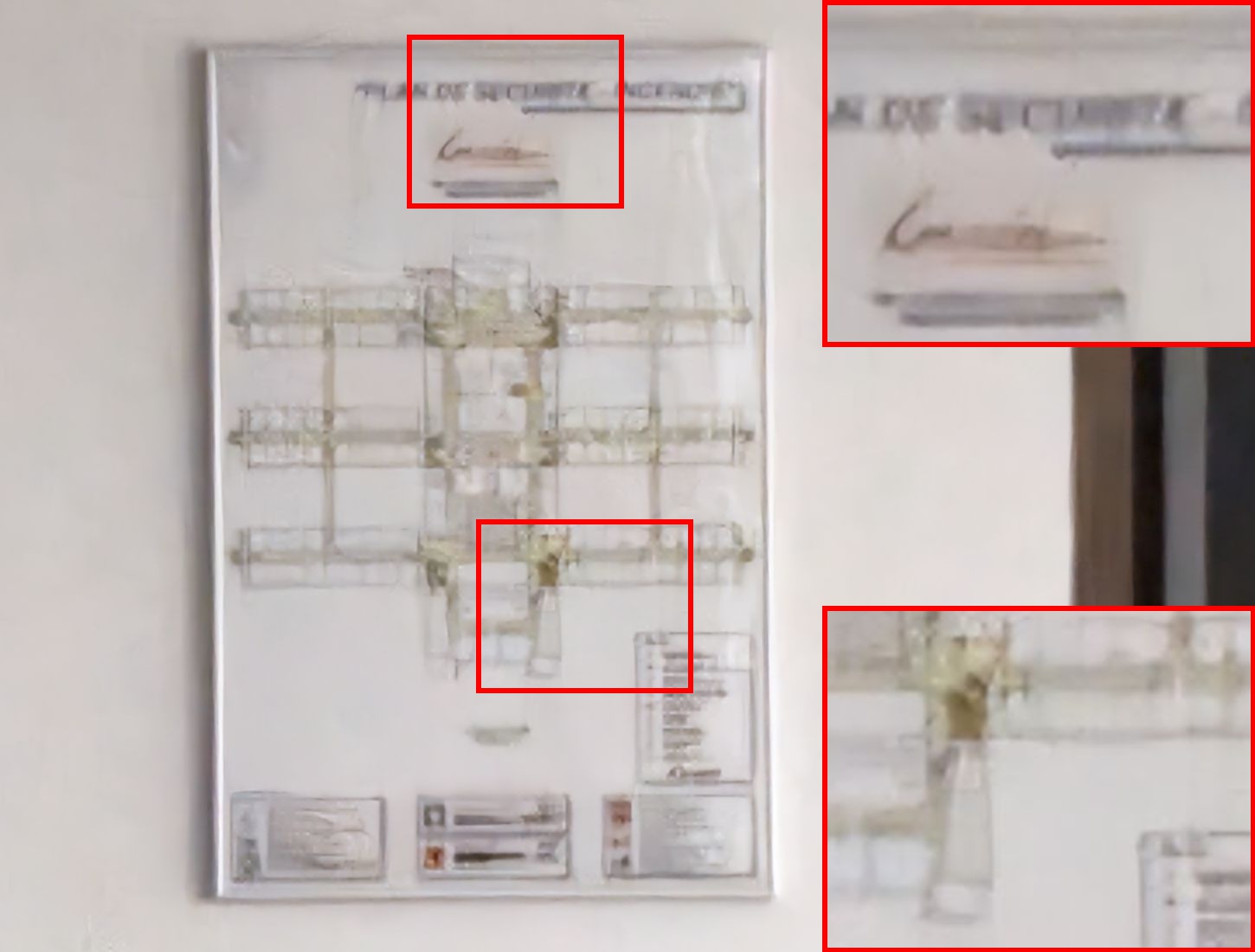}     & 
        \includegraphics[width=\w\textwidth]{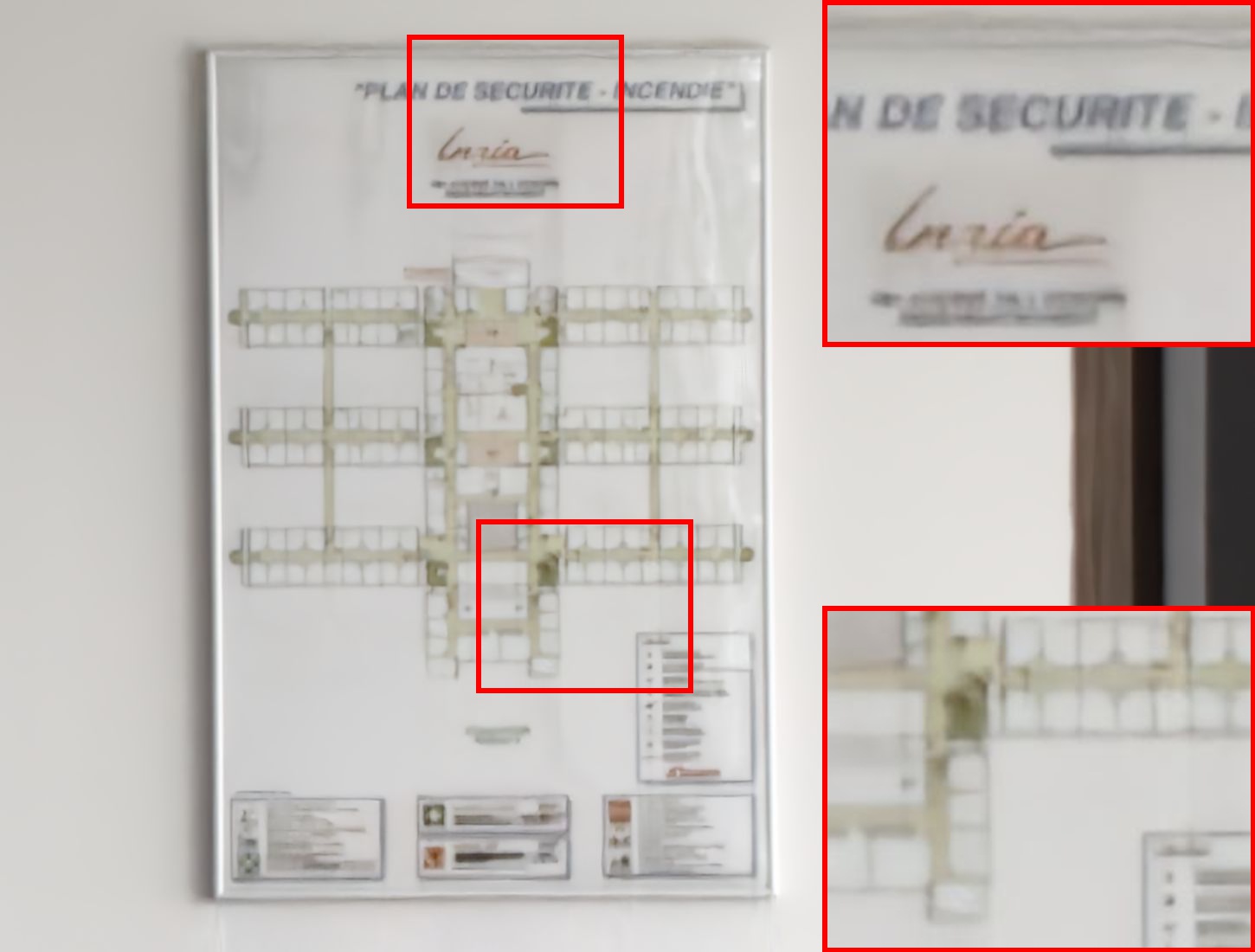}     \\   
        \parbox[t]{4mm}{\rotatebox[origin=l]{90}{Canon G7X}} & 
        \includegraphics[width=0.12\textwidth]{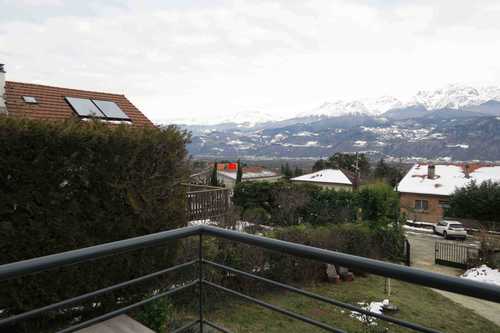}     & 
        \includegraphics[width=\w\textwidth]{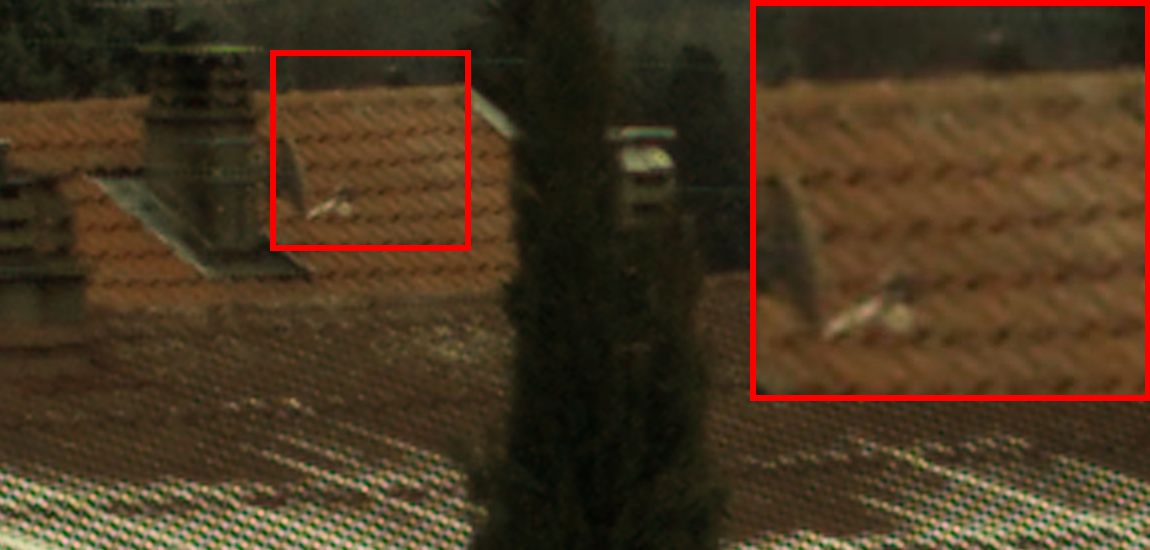}     & 
        \includegraphics[width=\w\textwidth]{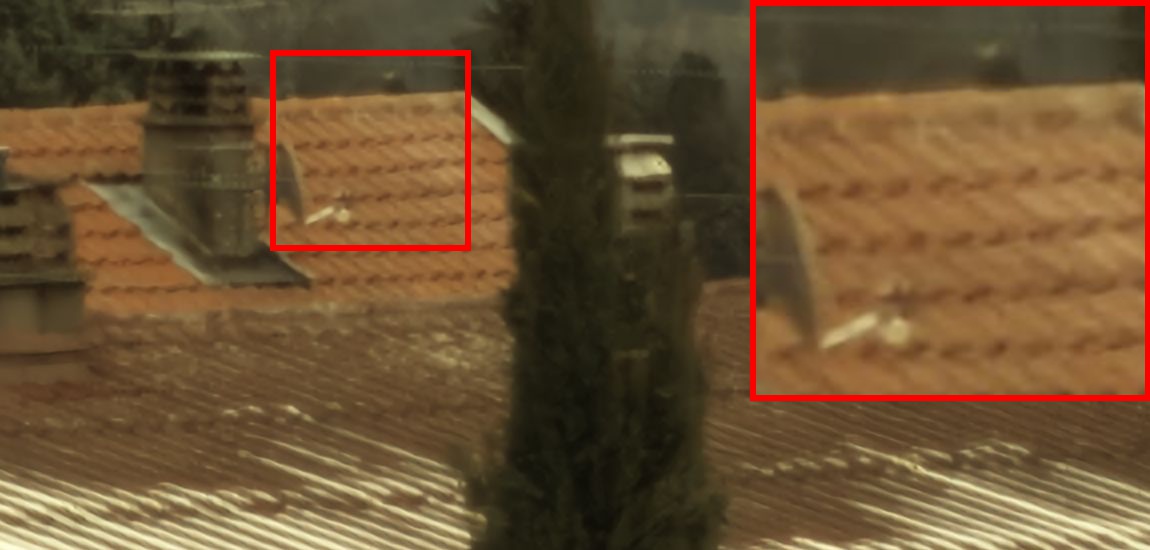}     & 
        \includegraphics[width=\w\textwidth]{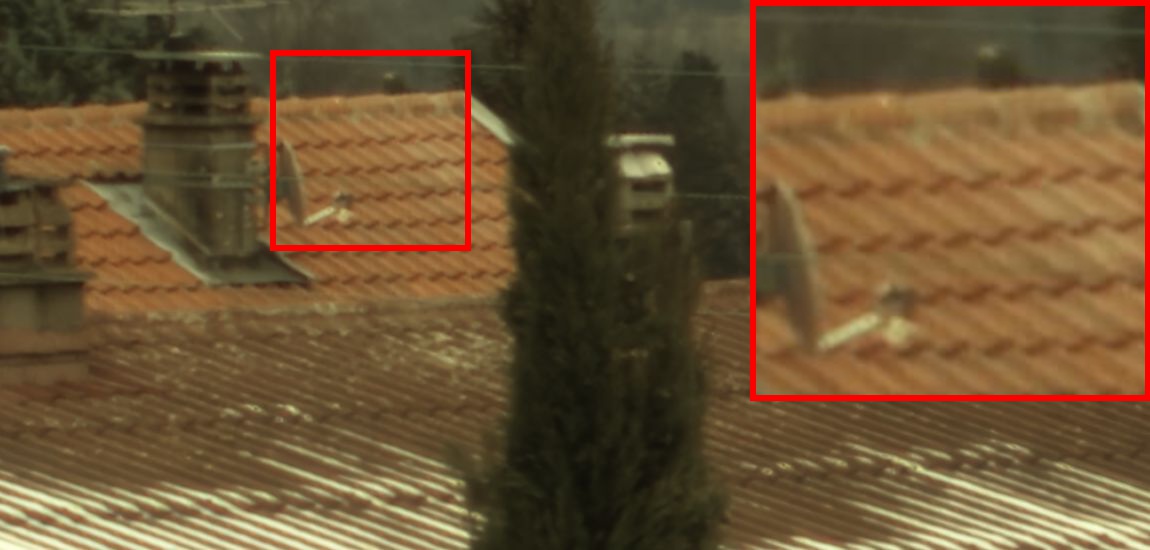}     \\

        \parbox[t]{4mm}{\rotatebox[origin=l]{90}{Samsung s7}} & 
        \includegraphics[width=0.12\textwidth]{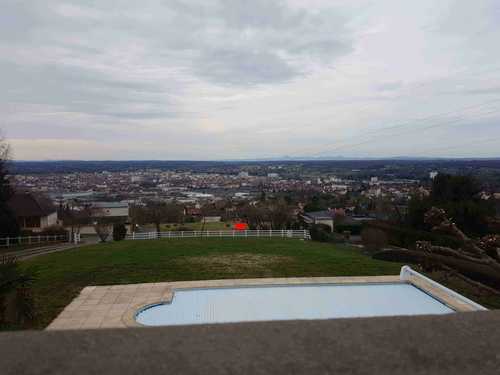}     & 
        \includegraphics[width=\w\textwidth]{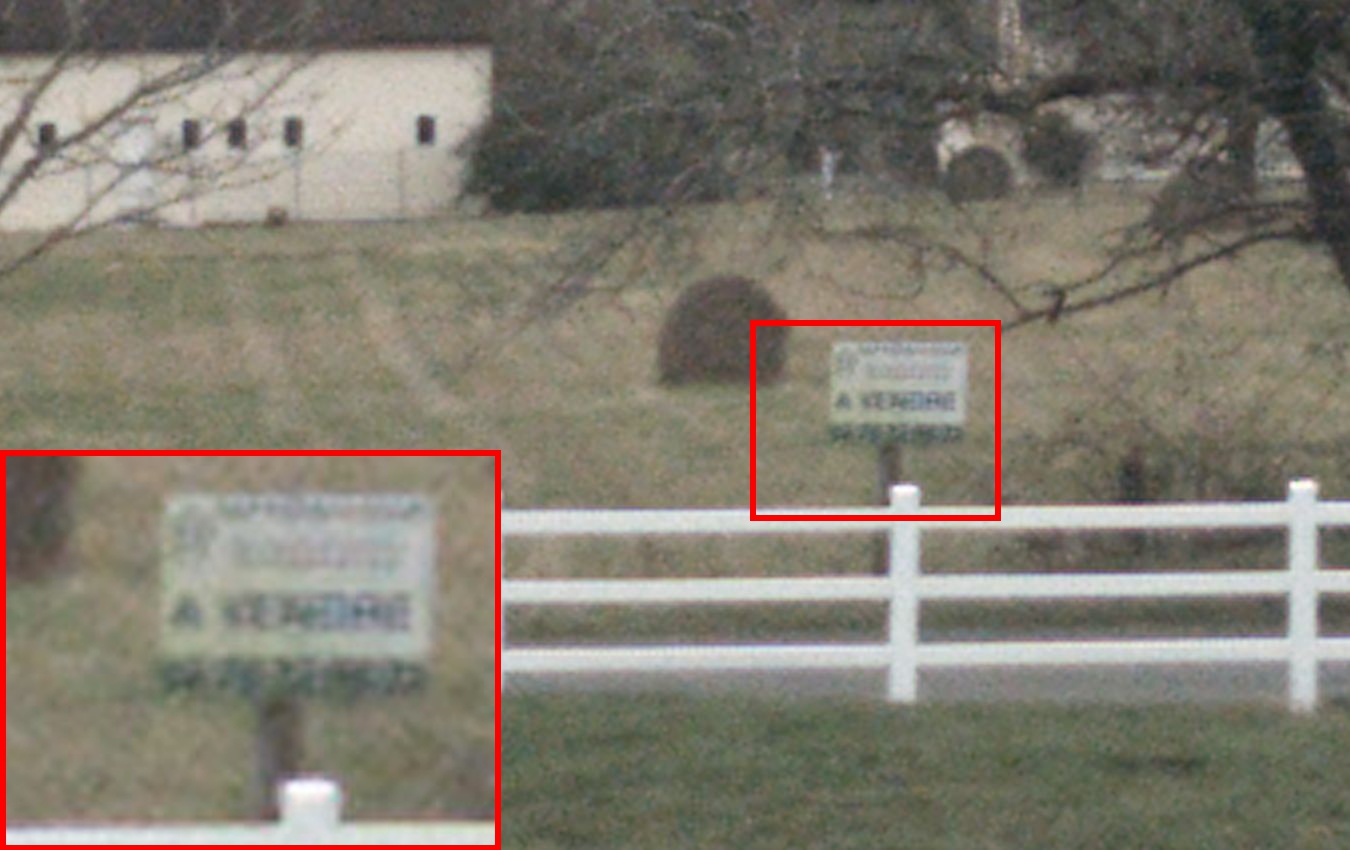}     & 
        \includegraphics[width=\w\textwidth]{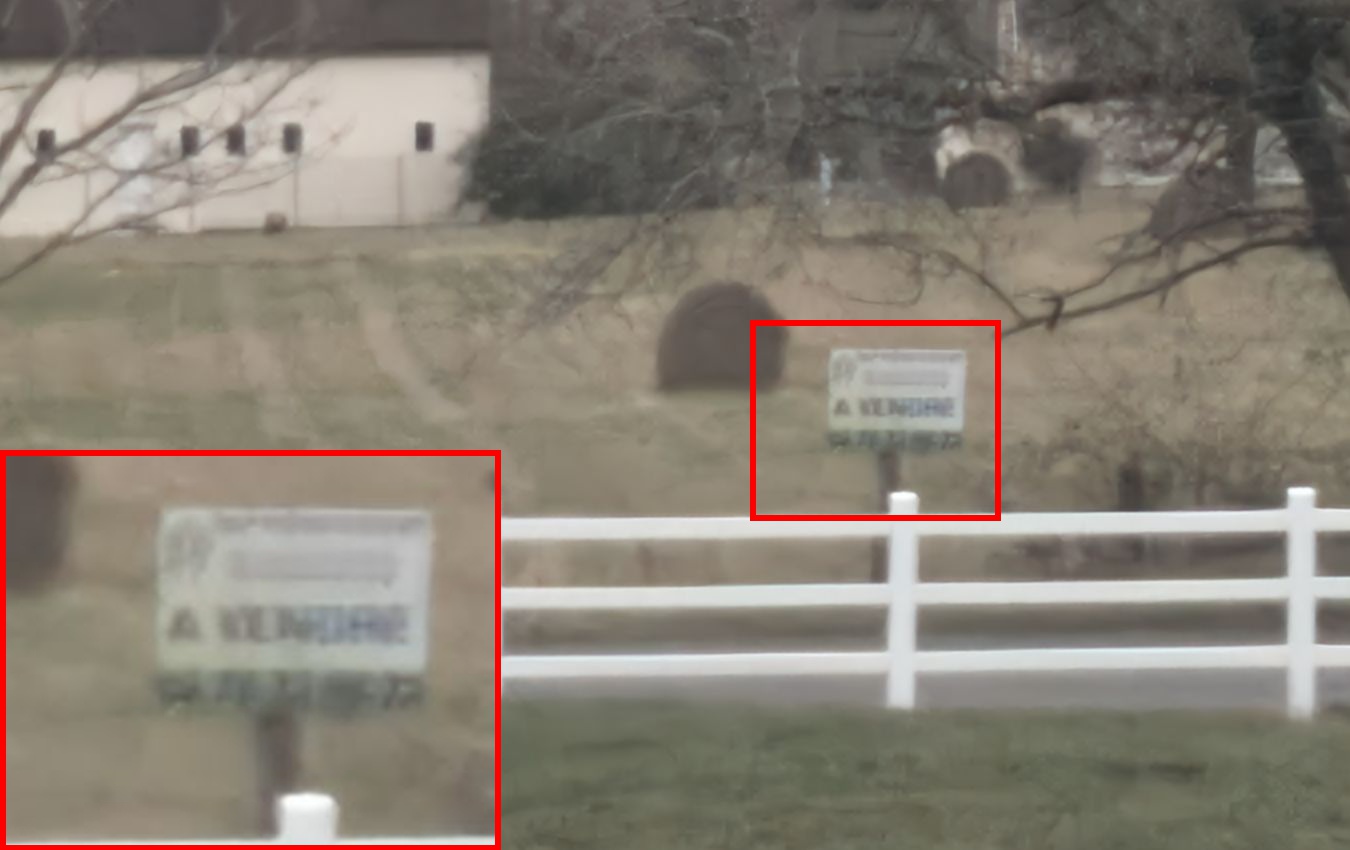}     & 
        \includegraphics[width=\w\textwidth]{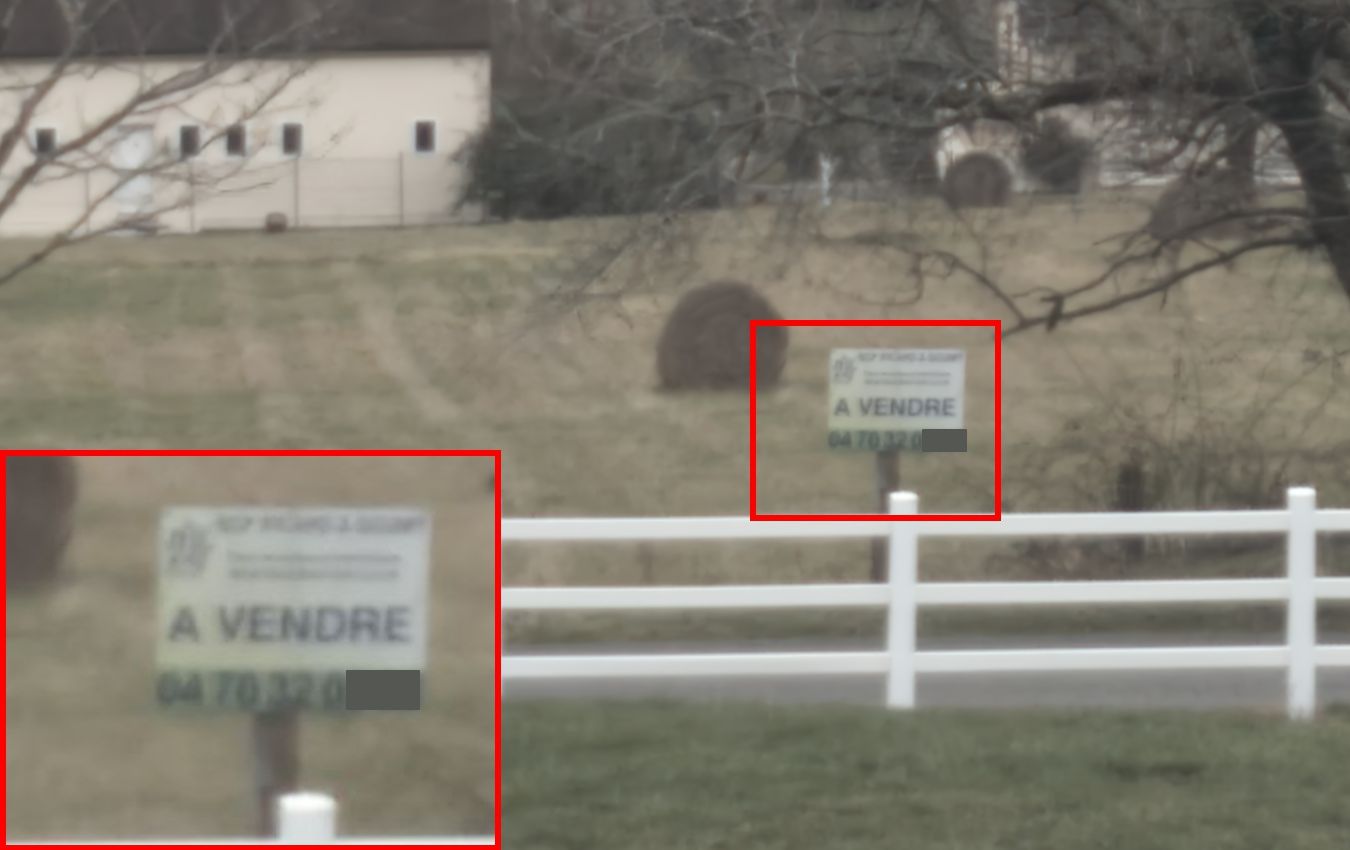}     \\
 
        \parbox[t]{4mm}{\rotatebox[origin=l]{90}{Pixel3a}} & 
        \includegraphics[width=0.12\textwidth]{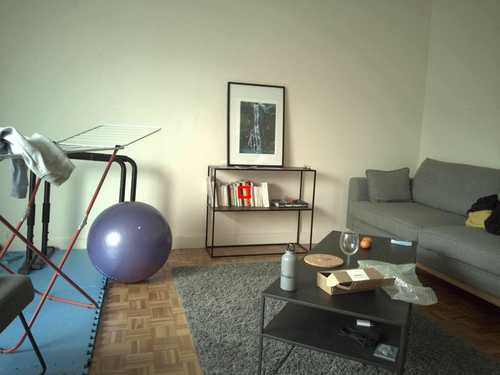}     & 
        \includegraphics[width=\w\textwidth]{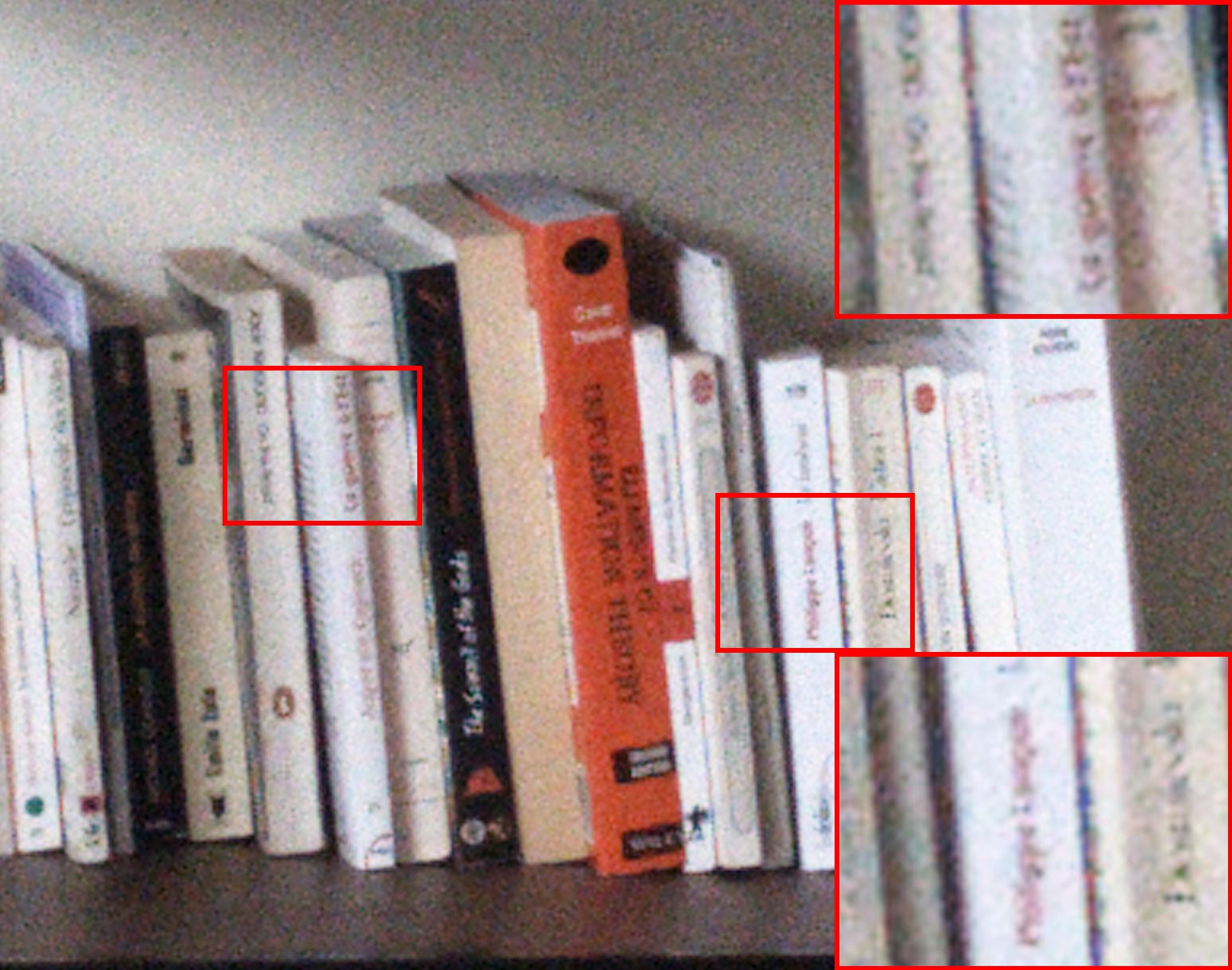}     & 
        \includegraphics[width=\w\textwidth]{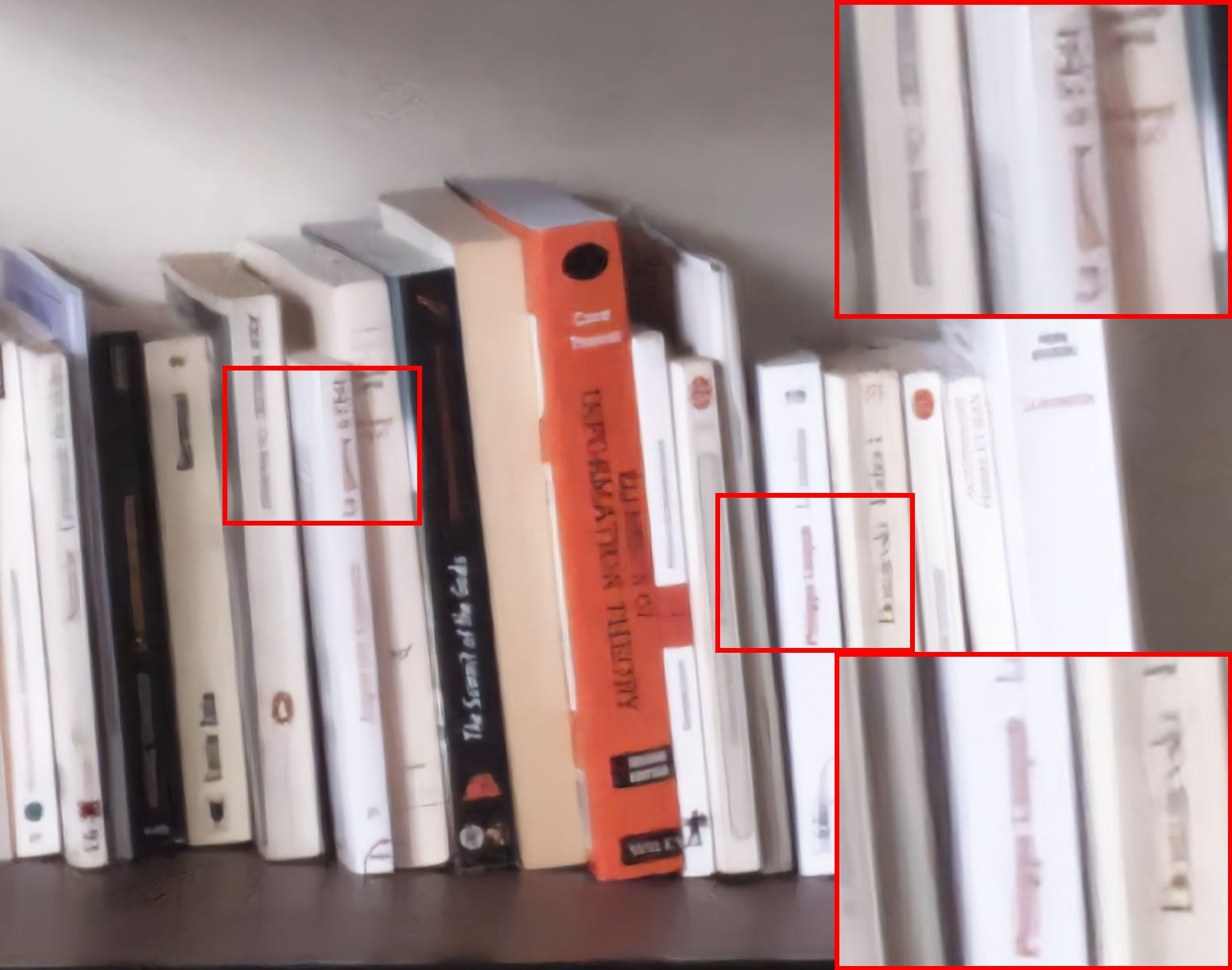}     & 
        \includegraphics[width=\w\textwidth]{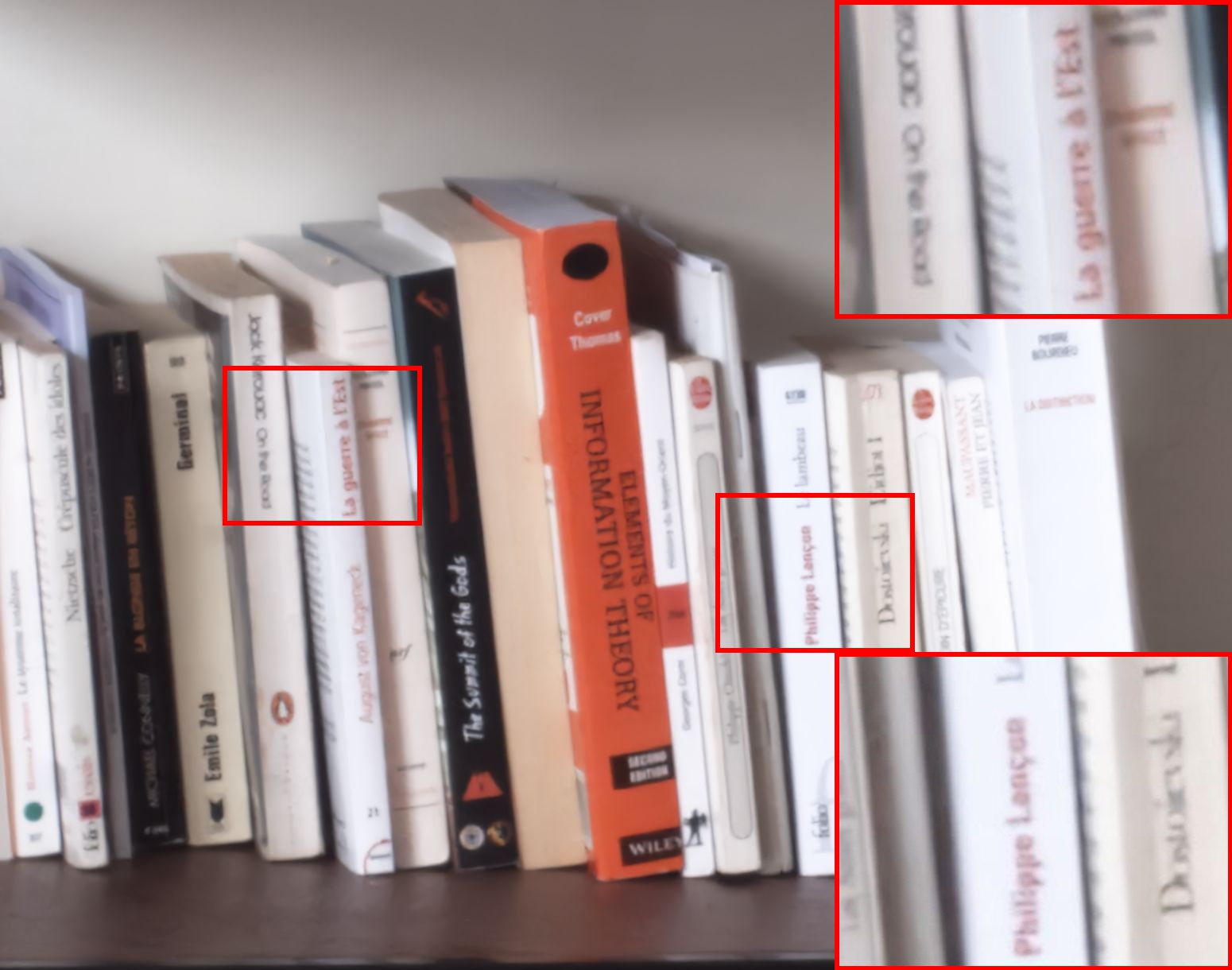}     \\

        \parbox[t]{4mm}{\rotatebox[origin=l]{90}{Pixel4a}} & 
        \includegraphics[width=0.12\textwidth]{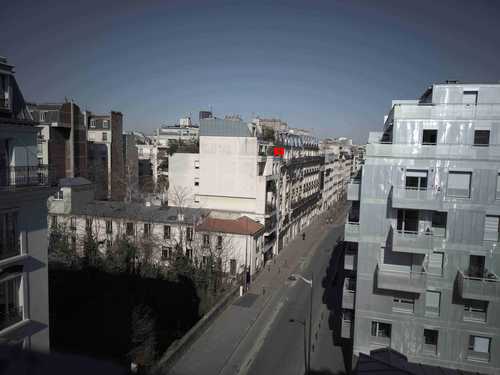}     & 
        \includegraphics[width=\w\textwidth]{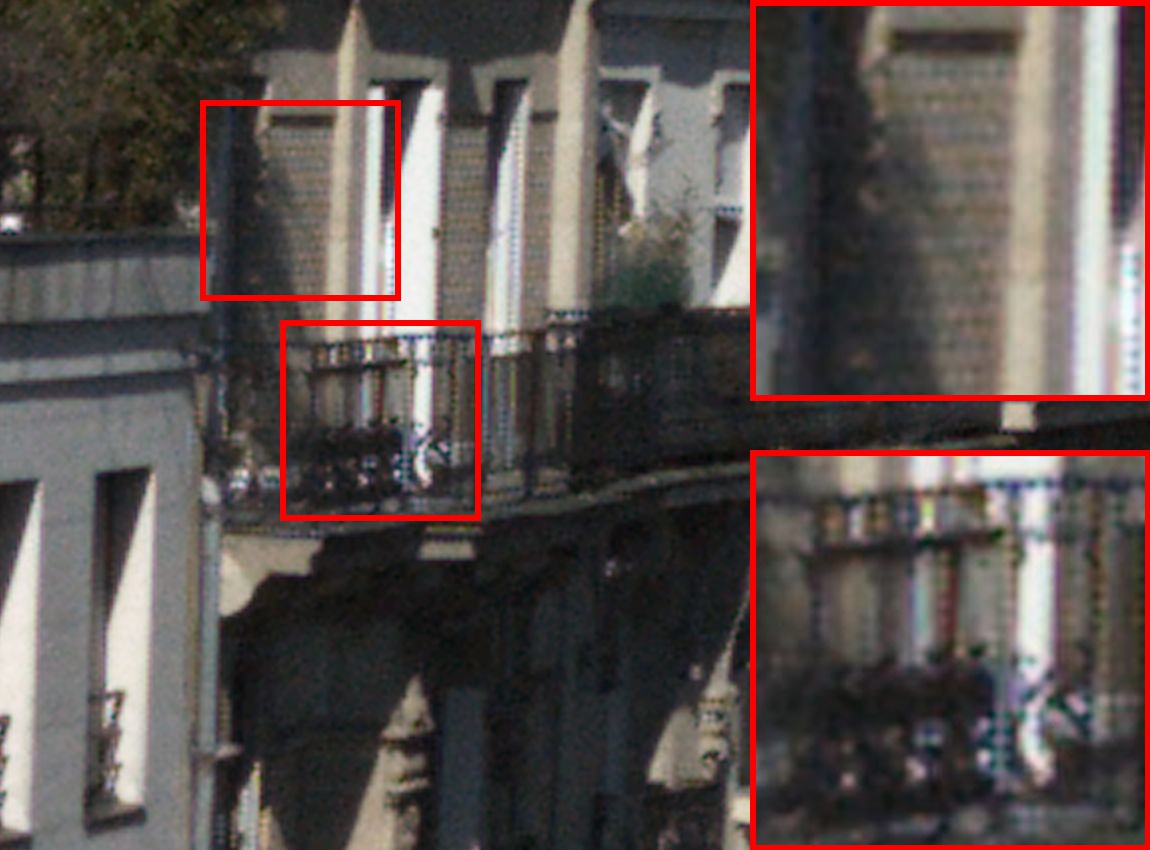}     & 
        \includegraphics[width=\w\textwidth]{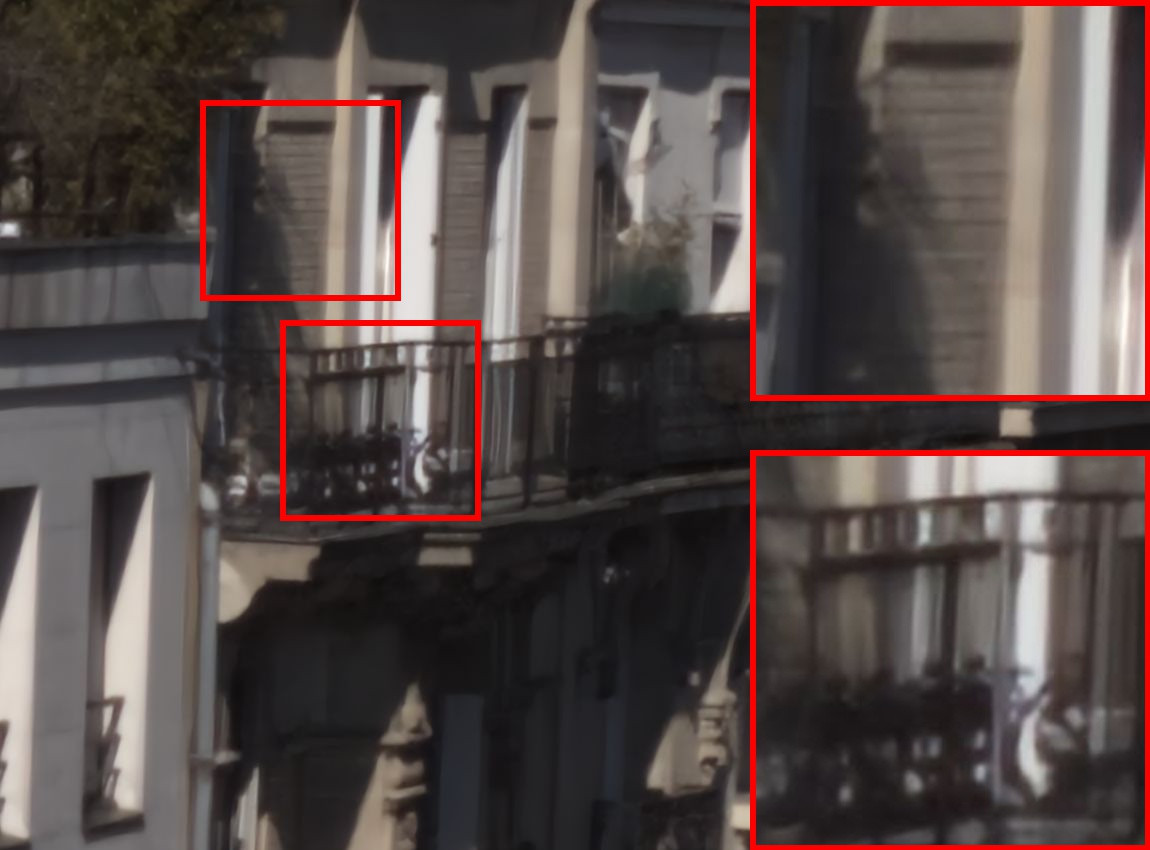}     & 
        \includegraphics[width=\w\textwidth]{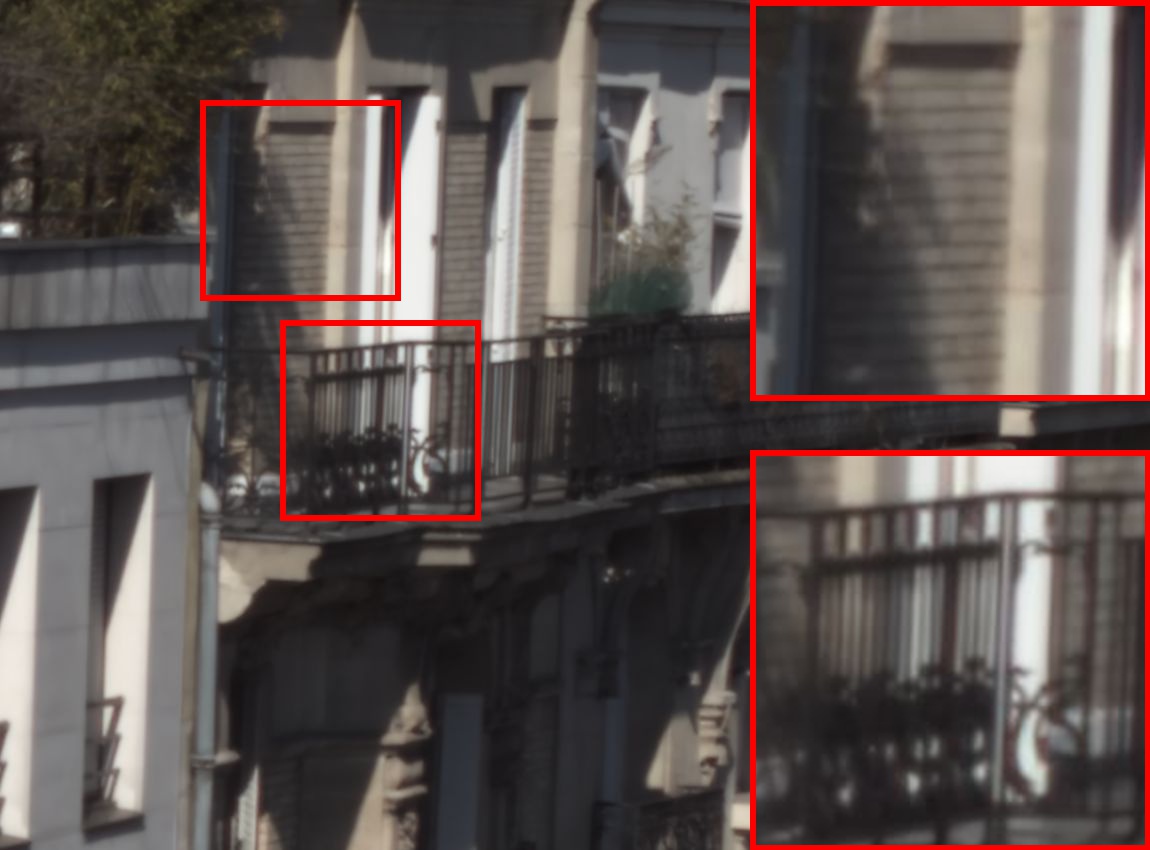}     \\

        & Source image & ISP Camera (jpeg output) & Joint demosaic+single-image SR & Our burst processing method \\

    \end{tabular}
    \vspace{-0.25cm}
    \caption{Results from real raw image bursts obtained with various cameras. We provide comparisons with single image and multiframe baselines. Finest restored details can be seen by zooming on a computer screen. The last three digits of the phone
    number, only legible in our reconstruction, are masked in the figure for
    privacy concerns.} 
    \label{fig:real}
\end{figure*}

\section{Conclusion}
 We have presented a simple but effective method for superresolution that
combines the interpretability of model-based approaches to inverse problems
with the flexibility of data-driven architectures and can be learned 
from pairs of synthetic LR and real HR images. We plan several extensions,
including using multiple cameras to add real LR-burst/HR-image
pairs to the training mix, and at test time to take advantage of the
multiplicity of imaging devices now available on high-end smartphones. This
will open the door to wide-baseline super-resolution applications, such as
the construction of high quality panoramas and finely detailed texture
maps in multi-view stereo reconstructions.  Finally, we plan to explore several
other extensions of our approach, including tackling blurry bursts, extending super-resolution to reconstruct HDR images, and pursuing applications in the astronomy
and microscopy domains

\subsection*{Acknowledgments}

We thank Fr\'ed\'eric Guichard for useful discussions and comments. 
This work was funded in part by the French government under management of Agence Nationale
de la Recherche as part of the ``Investissements d'avenir'' program, reference ANR-19-P3IA-0001
(PRAIRIE 3IA Institute). JM and BL were supported by the ERC grant number 714381 (SOLARIS
project) and by ANR 3IA MIAI@Grenoble Alpes (ANR-19-P3IA-0003). JP was supported in part by
the Louis Vuitton/ENS chair in artificial intelligence and the Inria/NYU collaboration. This
work was granted access to the HPC resources of IDRIS under the allocation 2020-AD011011252
made by GENCI.

{\small
    \bibliographystyle{ieee_fullname}
    \bibliography{reference}}

\clearpage
\normalsize	
\small
\setcounter{table}{0}
\setcounter{figure}{0}
\renewcommand{\thetable}{A\arabic{table}}
\renewcommand{\thefigure}{A\arabic{figure}}

\appendix
\onecolumn

\section*{Supplementary material}
\begin{figure}[h!]
    \newcommand\x{0.2}
    \definecolor{mycolor1}{RGB}{255,255,255}
    \footnotesize
    \vspace{-0.3cm}
    \centering
    \renewcommand{\arraystretch}{0.3}
    \setlength\tabcolsep{0.3pt}
    \small
    \begin{tabular}{cccccc}
        \includegraphics[width=\x\textwidth]{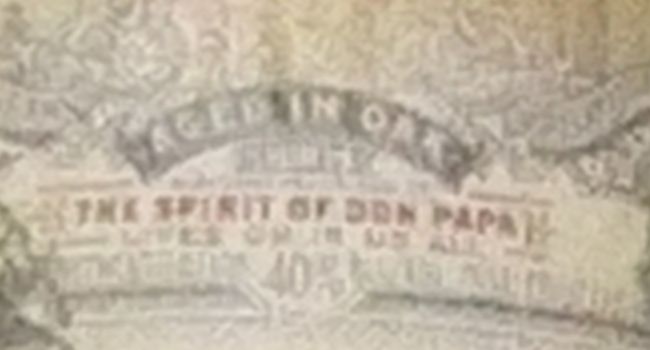}   &
        \includegraphics[width=\x\textwidth]{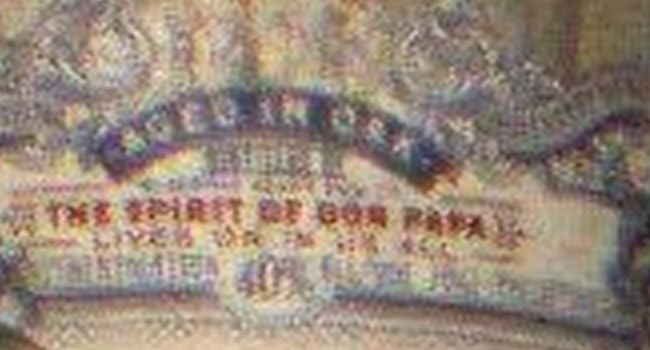}   &
        \includegraphics[width=\x\textwidth]{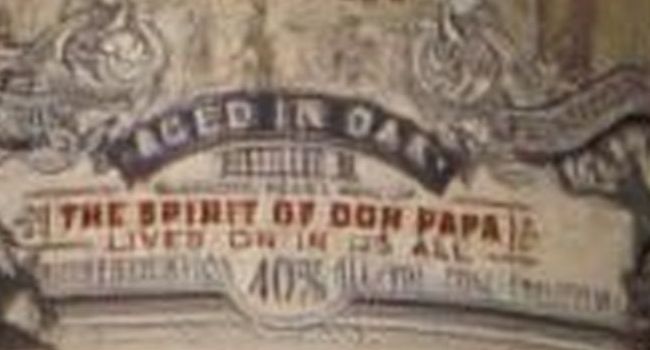}   &
        \includegraphics[width=\x\textwidth]{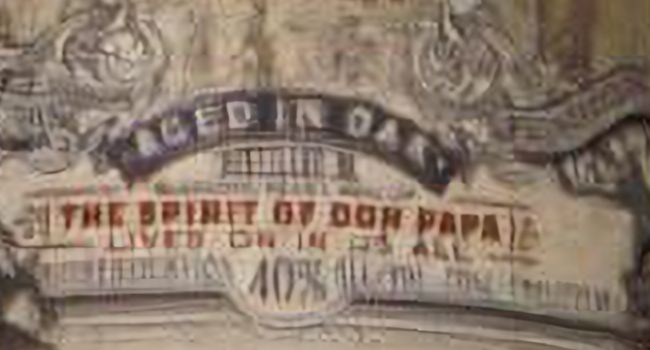}   &
        \includegraphics[width=\x\textwidth]{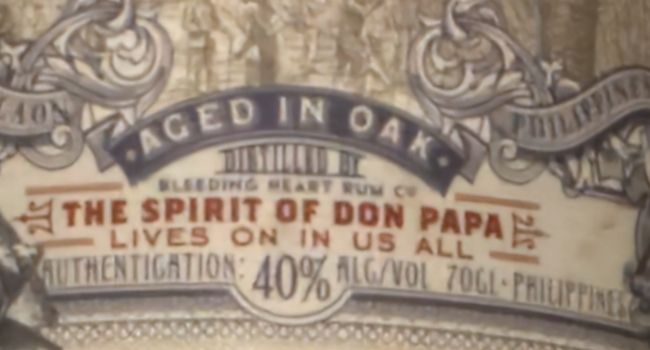}   \\
        ISP & BKPN \cite{xia2020basis}+demosaic & Mosa2mosa \cite{ehret2019joint} & Mosa2mosa \cite{ehret2019joint} + EDSR & Ours

   \end{tabular}
    \caption{Comparison with joint denoising and demosaicking methods.}\label{fig:joint}
    \vspace{-0.5cm}
    \end{figure}

\vsp
\noindent
This supplementary material presents additional qualitative and quantiative results.
In Figure \ref{fig:joint} we present additional visual comparison with two burst denoising methods on real images.
In Table \ref{tab:RGB} we present additional experiments on RGB images.
Figures~\ref{fig:real} and~\ref{fig:realb} are devoted to super-resolution experiments from real raw data from different smartphones (Google Pixel 3a and 4a, Samsung S7 and S10) and cameras (Panasonic Lumix GX9 and Canon Powershot G7X) and comparison with additional baselines.
In Figures~\ref{fig:x16b}, we present extreme upsampling results by using synthetic RGB image bursts.
In Figure~\ref{fig:lowlight}, we present restoration results obtained from real images with very low SNR  to illustrate the efficiency of our method to perform blind denoising.
In Figures~\ref{fig:ablation} and~\ref{fig:ablationb}, we study the effect of the number of frames in the burst on the reconstruction, both in the low SNR and high SNR settings.  
Finally, we present failure cases in Figure~\ref{fig:failure}, where fast moving objects are present in the scene.

\vspace{-0.2cm}
\paragraph{Comparison with burst denoising methods.}
We perform additional qualitative comparison on a real image with two burst denoising methods.
We compare our method with \cite{ehret2019joint} which performs joint denoising and demosaicking on a burst of raw images. We use the code and the pretrained model made available online.
We also use the code and pretrained model of \cite{xia2020basis}. However the model is only designed to perform grayscale burst denoising, so we perform denoising independently on each RGB channel and then perform demosaicking to get an RGB image.
Despite our best efforts for tuning the parameters of these methods to maximize visual quality, the
results obtained are not as good as our method (see Figure \ref{fig:joint} below). We believe this is not surprising since each one of these methods only addresses a subset of our problem. Adapting them successfully to our general setting is
not trivial.

\vspace*{-0.2cm}
\paragraph{Evaluation on RGB images.}
We compare our approach on the BSD68 dataset against state-of-the-art single-image and video super-resolution algorithms (considering a burst as a video sequence) and report the HR image reconstruction accuracy in terms of average PSNR in Table \ref{tab:RGB}. For the training with RGB data, we perform $80\,000$ iterations of the ADAM optimizer with a batch size of $10$, a burst size of $14$ and
with a learning rate of $3\times 10^{-5}$ decaying by a factor $2$ after
$40\,000$ iterations. 
For evaluating the model VSR-DUF~\cite{jo2018deep}, we use the code and the pretrained models made available online by the authors. Other single-image reconstruction results are from \cite{zhang2020deep}. 
In the present setting, we consistently outperform other baselines, notably demonstrating that burst SR cannot simply be addressed effectively by current video SR approaches.   
We also note that our models perfom better with less blurring (and more aliasing). 
Finally, we evaluate variations of our model in the same table, notably
comparing the registration accuracy achieved by these variants by using the geometrical error presented in~\cite{sanchez2016inverse}.
More precisely, we perform a small ablation study by introducing a simpler baseline that does not perform joint alignement and only exploits the coarse registration module (no refine baseline). 
Performing joint alignment and image estimation systematically improves motion estimation. Last, we also report the oracle performance of our model with known motions. 

\newcommand\myeq{\mkern1.5mu{=}\mkern1.5mu}
\begin{table}[h!]
    \centering
    \small
    \begin{tabular}{@{}lccc@{}}
        \toprule
        \multicolumn{1}{l}{\multirow{2}{*}{Method}}     & \multicolumn{3}{c}{ Scaling factor / blurring kernel std }                                                                                                        \\
        \multicolumn{1}{c}{}                            & \multicolumn{1}{c}{$\times 2 / \sigma \myeq0.7 $}          & \multicolumn{1}{c}{$ \times 3 / \sigma \myeq1.2$} & \multicolumn{1}{c}{$\times 4 / \sigma \myeq1.6$} \\ 
        \midrule 
        \multicolumn{4}{l}{\textit{Single Image SR}}                                                                                                                                                                        \\
        \multirow{1}{*}{RCAN \cite{zhang2018image}}     & 29.48                                                      & 27.30                                             & 25.59                                            \\ 
        \multirow{1}{*}{IRCNN \cite{zhang2017learning}} & 29.60                                                      & 26.89                                             & 25.32                                            \\  
        \multirow{1}{*}{USRNet \cite{zhang2020deep}}    & 30.55                                                      & 27.76                                             & 26.18                                            \\        
        \multicolumn{4}{l}{\textit{Video SR}}                                                                                                                                                                               \\
        \multirow{1}{*}{VSR-DUF}\cite{jo2018deep}       & -                                                      & 31.03                                             & 29.24                                            \\ 
        \midrule 
        
        \multirow{1}{*}{Ours (no refine)}               & 42.36/0.10                                                 & 32.63/0.14                                        & \textbf{30.00}/0.19                              \\
        \multirow{1}{*}{Ours}                           & \textbf{43.73}/0.07                                        & \textbf{33.10}/0.10                               & {29.87}/0.14                                     \\
        \midrule
        \multirow{1}{*}{Ours (known $\pb$)}             & 45.72/0.00                                                 & 34.47/0.00                                        & 31.32/0.00                                       \\
        \bottomrule
    \end{tabular}
    \caption{\textbf{Results for RGB with synthetic affine motions},
        of different methods for different combinations of scale factors and blur
        kernels. Results are given in term of average PSNR in dBs and geometrical
        registration error in pixels for our models. ``known $\pb$'' is the oracle performance our model could achieve, if motion estimation was perfect. }\label{tab:RGB}
\end{table}

\vspace{-0.0cm}

\begin{figure*}[h!]
    \vspace{-0.2cm}
    \renewcommand\w{0.16}
    \renewcommand{\arraystretch}{1}
    \centering
    \setlength\tabcolsep{0.2pt}
    \begin{tabular}{ccccccc}
        \parbox[t]{4mm}{\rotatebox[origin=l]{90}{Pixel 4a}} & 
        \includegraphics[trim=0 0 0 0, clip,width=0.14\textwidth]{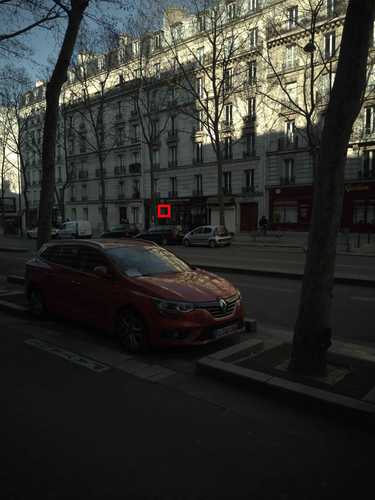}     & 
        \includegraphics[trim=0 20 0 20, clip,width=\w\textwidth]{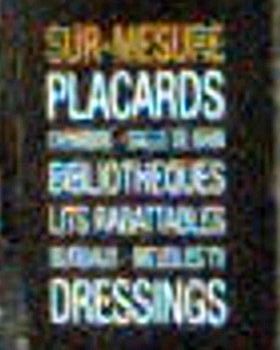}     & 
        \includegraphics[trim=0 10 0 30, clip,width=\w\textwidth]{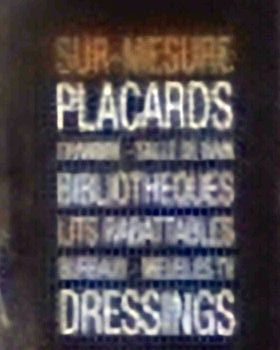}     & 
        \includegraphics[trim=0 10 0 30, clip,width=\w\textwidth]{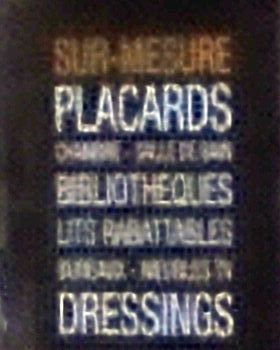}     & 
        \includegraphics[trim=0 25 0 15, clip,width=\w\textwidth]{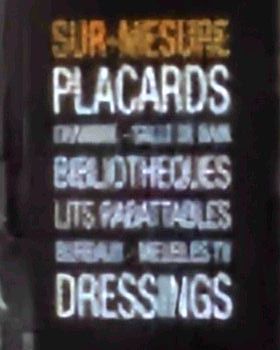}     & 
        \includegraphics[trim=0 25 0 15, clip,width=\w\textwidth]{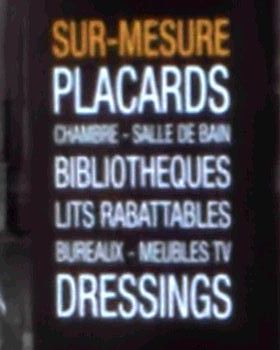}     \\
        \parbox[t]{4mm}{\rotatebox[origin=l]{90}{Pixel 4a}} & 
        \includegraphics[width=\w\textwidth]{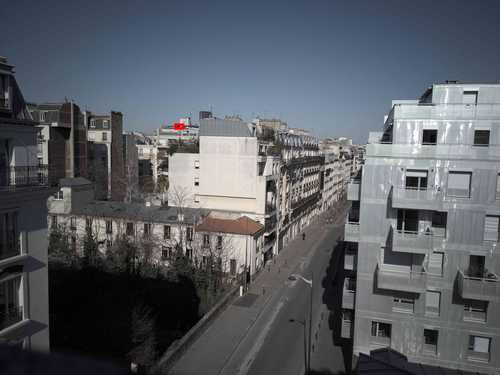}     & 
        \includegraphics[width=\w\textwidth]{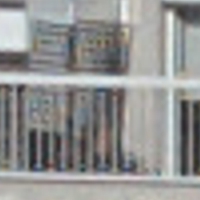}     & 
        \includegraphics[width=\w\textwidth]{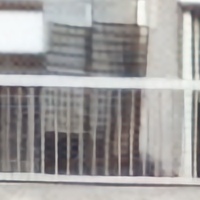}     & 
        \includegraphics[width=\w\textwidth]{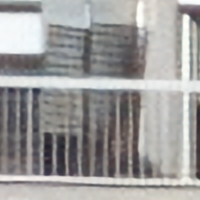}     & 
        \includegraphics[width=\w\textwidth]{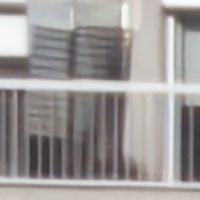}     & 
        \includegraphics[width=\w\textwidth]{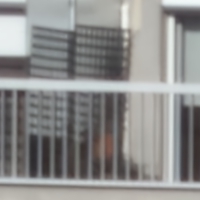}     \\
        \parbox[t]{4mm}{\rotatebox[origin=l]{90}{Pixel 4a}} & 
        \includegraphics[width=\w\textwidth]{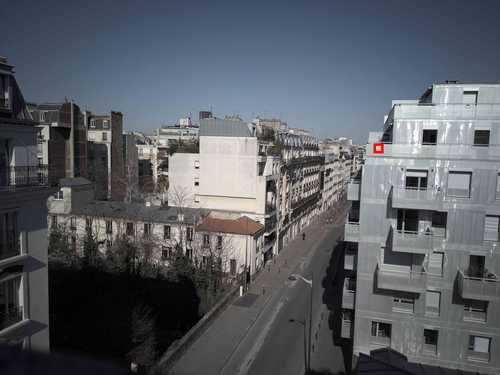}     & 
        \includegraphics[width=\w\textwidth]{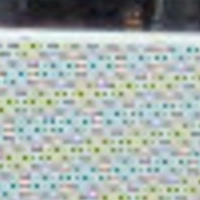}     & 
        \includegraphics[width=\w\textwidth]{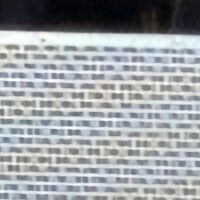}     & 
        \includegraphics[width=\w\textwidth]{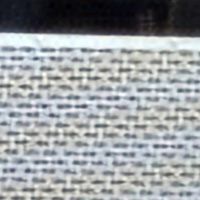}     & 
        \includegraphics[width=\w\textwidth]{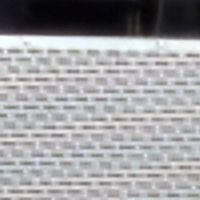}     & 
        \includegraphics[width=\w\textwidth]{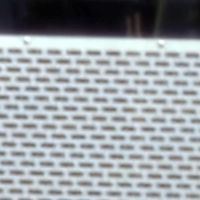}     \\
        \parbox[t]{4mm}{\rotatebox[origin=l]{90}{Pixel 3a}} & 
        \includegraphics[width=\w\textwidth]{appendix/biblio/crops/full_crop.jpg}     & 
        \includegraphics[width=\w\textwidth]{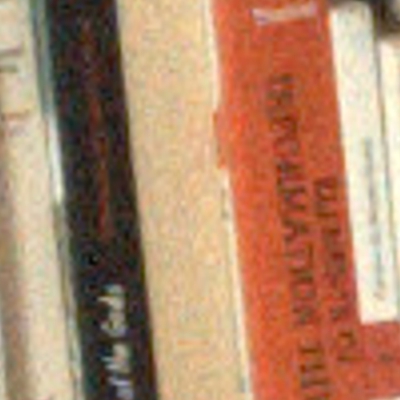}     & 
        \includegraphics[width=\w\textwidth]{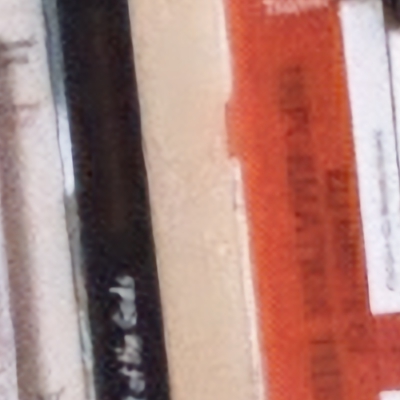}     & 
        \includegraphics[width=\w\textwidth]{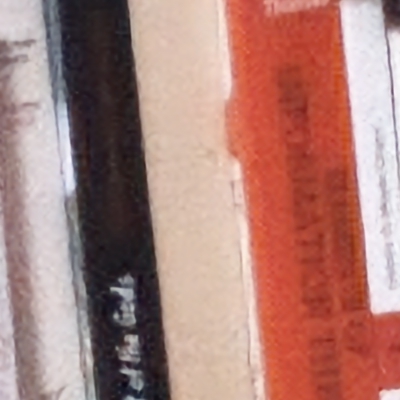}     & 
        \includegraphics[width=\w\textwidth]{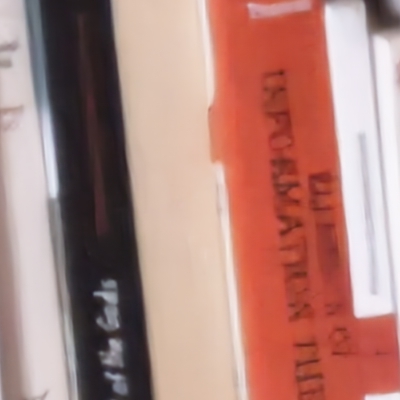}     & 
        \includegraphics[width=\w\textwidth]{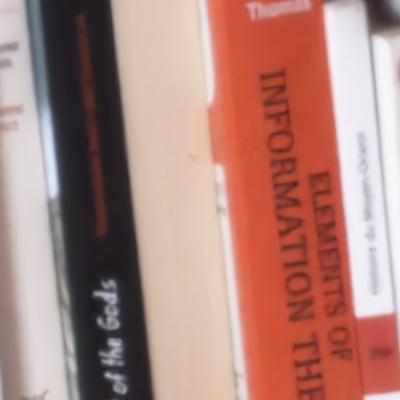}     \\
        
        \parbox[t]{4mm}{\rotatebox[origin=l]{90}{Panasonic}} & 
        \includegraphics[width=\w\textwidth]{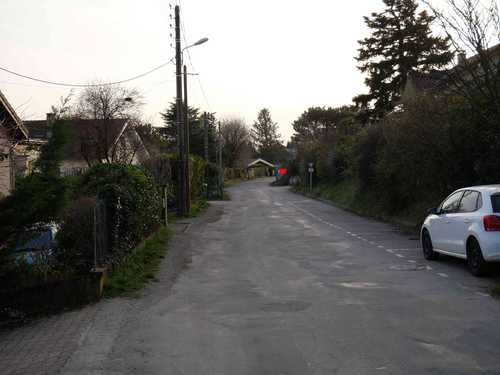}     & 
        \includegraphics[width=\w\textwidth]{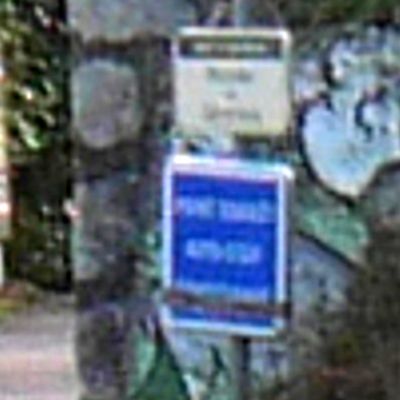}     & 
        \includegraphics[width=\w\textwidth]{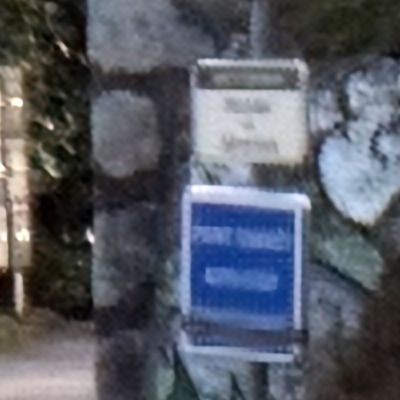}     & 
        \includegraphics[width=\w\textwidth]{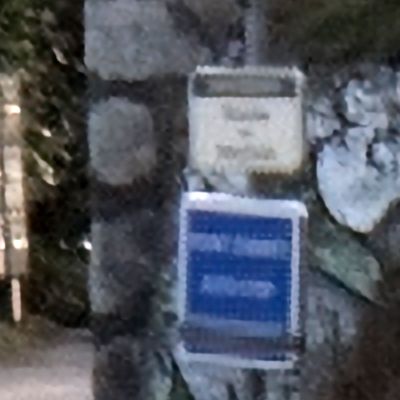}     & 
        \includegraphics[width=\w\textwidth]{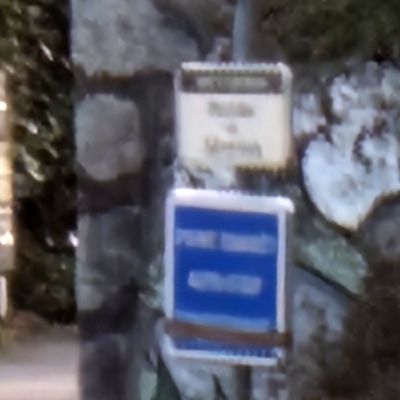}     & 
        \includegraphics[width=\w\textwidth]{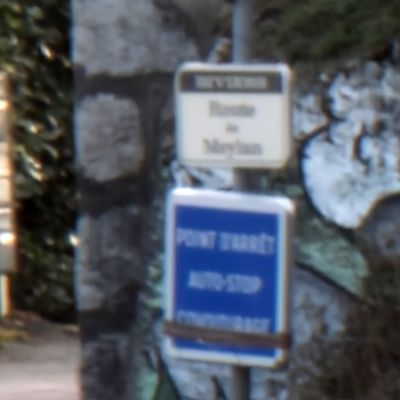}     \\
        \parbox[t]{4mm}{\rotatebox[origin=l]{90}{Panasonic}} & 
        \includegraphics[trim=0 0 0 20, clip,width=\w\textwidth]{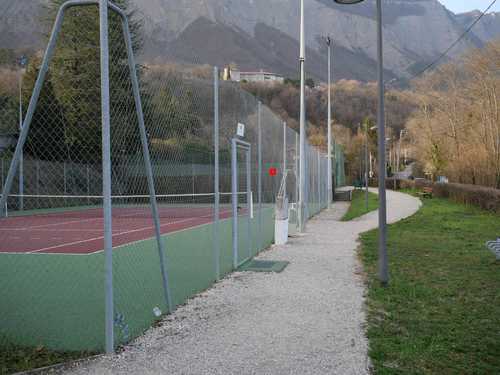}     & 
        \includegraphics[trim=0 0 0 20, clip,width=\w\textwidth]{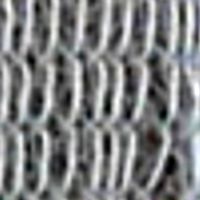}     & 
        \includegraphics[trim=0 0 0 20, clip,width=\w\textwidth]{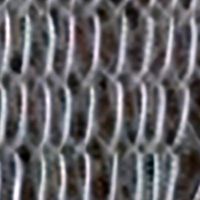}     & 
        \includegraphics[trim=0 0 0 20, clip,width=\w\textwidth]{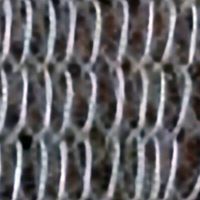}     & 
        \includegraphics[trim=0 0 0 20, clip,width=\w\textwidth]{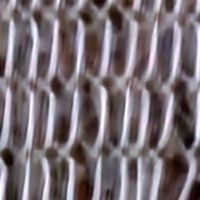}     & 
        \includegraphics[trim=0 0 0 20, clip,width=\w\textwidth]{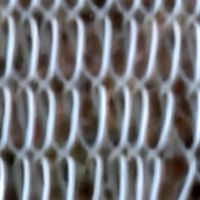}     \\
        
        & Full image & Camera ISP & \begin{tabular}{c} groupsc  \cite{lecouat2020fully} \\ + EDSR \cite{lim2017enhanced} \end{tabular} & \begin{tabular}{c} groupsc \cite{lecouat2020fully} \\ + VSR-DUF \cite{jo2018deep} \end{tabular} & Joint mosa \& SR & Ours \\
    \end{tabular}
    \vspace{-0.25cm}
    \caption{Results from real raw image bursts obtained with various cameras. We provide comparisons with single image and multiframe baselines. Finest restored details can be seen by zooming on computer screen.} \label{fig:real}
\end{figure*}

\begin{figure*}[h!]
    \vspace{-0.5cm}
    \renewcommand\w{0.16}
    \renewcommand{\arraystretch}{1}
    \centering
    \setlength\tabcolsep{0.2pt}
    \begin{tabular}{ccccccc}

        \parbox[t]{4mm}{\rotatebox[origin=l]{90}{Pixel 3a}} & 
        \includegraphics[width=\w\textwidth]{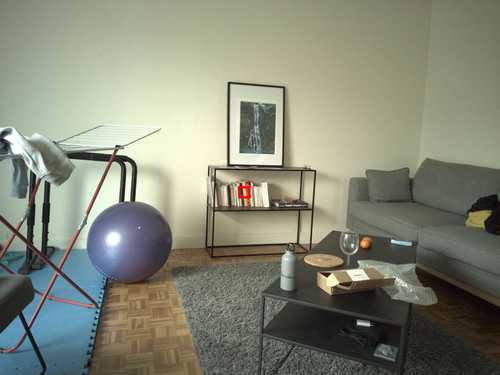}     & 
        \includegraphics[width=\w\textwidth]{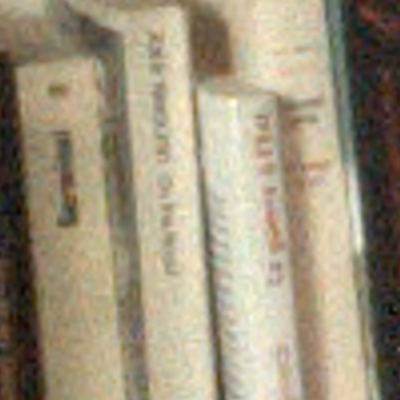}     & 
        \includegraphics[width=\w\textwidth]{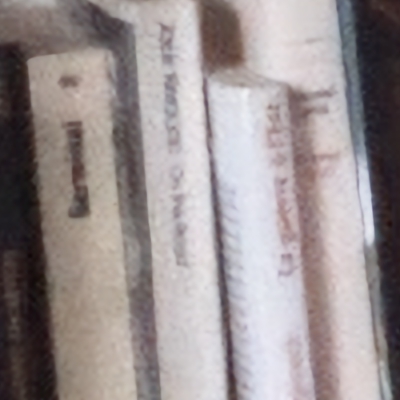}     & 
        \includegraphics[width=\w\textwidth]{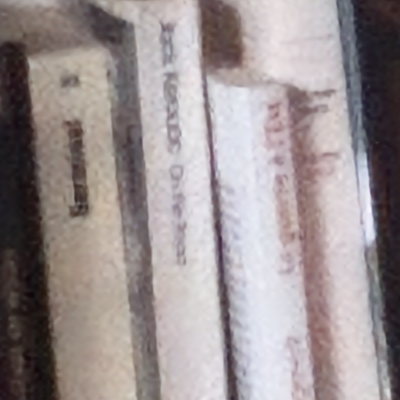}     & 
        \includegraphics[width=\w\textwidth]{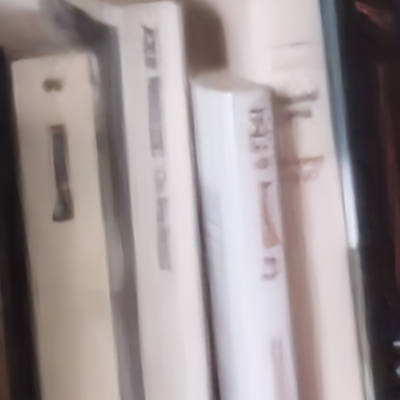}     & 
        \includegraphics[width=\w\textwidth]{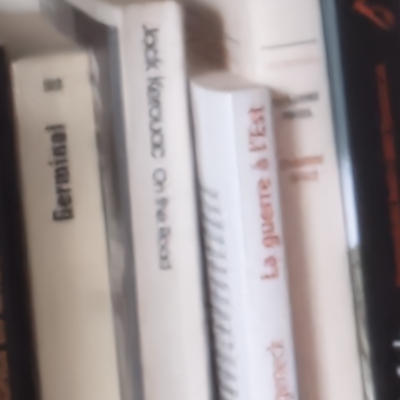}     \\
        
        \parbox[t]{4mm}{\rotatebox[origin=l]{90}{Pixel 4a}} & 
        \includegraphics[width=\w\textwidth]{appendix/hotel/crops/full_crop.jpg}     & 
        \includegraphics[width=\w\textwidth]{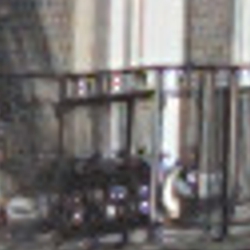}     & 
        \includegraphics[width=\w\textwidth]{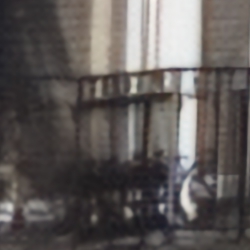}     & 
        \includegraphics[width=\w\textwidth]{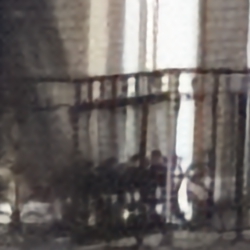}     & 
        \includegraphics[width=\w\textwidth]{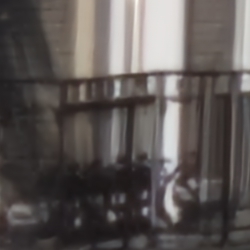}     & 
        \includegraphics[width=\w\textwidth]{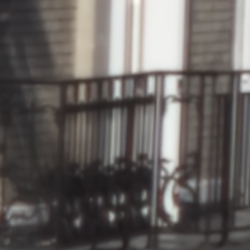}     \\

        \parbox[t]{4mm}{\rotatebox[origin=l]{90}{Samsung S7}} & 
        \includegraphics[width=\w\textwidth]{appendix/immobilier/crops/full_crop.jpg}     & 
        \includegraphics[width=\w\textwidth]{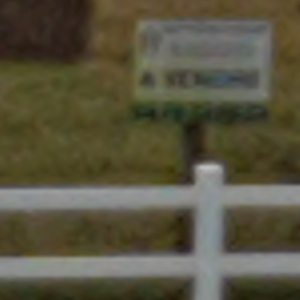}     & 
        \includegraphics[width=\w\textwidth]{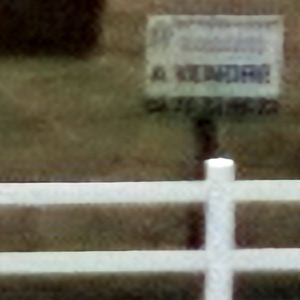}     & 
        \includegraphics[width=\w\textwidth]{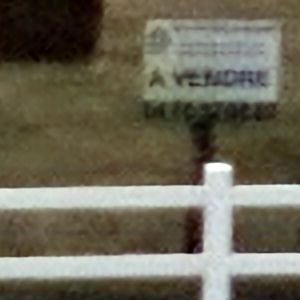}     & 
        \includegraphics[width=\w\textwidth]{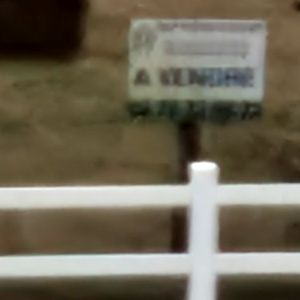}     & 
        \includegraphics[width=\w\textwidth]{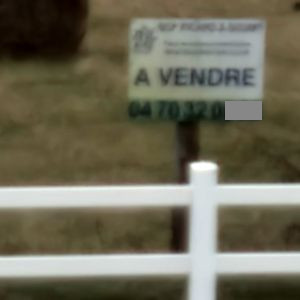}     \\

        \parbox[t]{4mm}{\rotatebox[origin=l]{90}{Canon G7X}} & 
        \includegraphics[width=\w\textwidth]{appendix/b1/crops/full_crop.jpg}     & 
        \includegraphics[width=\w\textwidth]{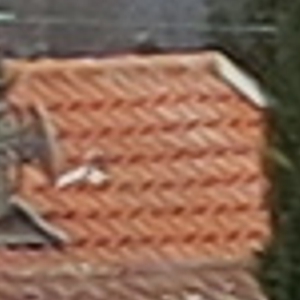}     & 
        \includegraphics[width=\w\textwidth]{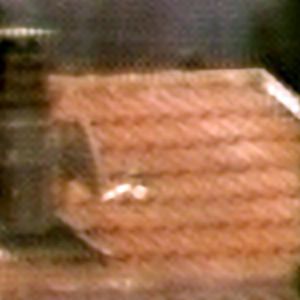}     & 
        \includegraphics[width=\w\textwidth]{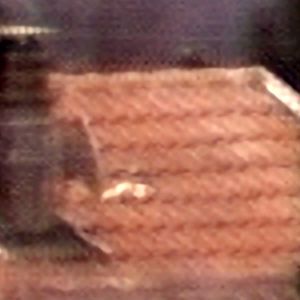}     & 
        \includegraphics[width=\w\textwidth]{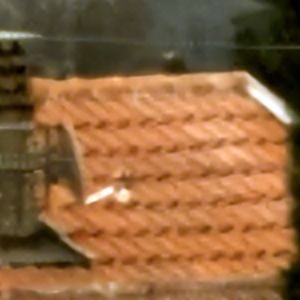}     & 
        \includegraphics[width=\w\textwidth]{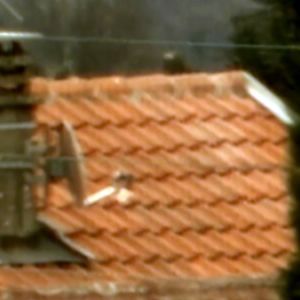}     \\

        \parbox[t]{4mm}{\rotatebox[origin=l]{90}{Panasonic}} & 
        \includegraphics[width=\w\textwidth]{appendix/plan/crops/full_crop.jpg}     & 
        \includegraphics[width=\w\textwidth]{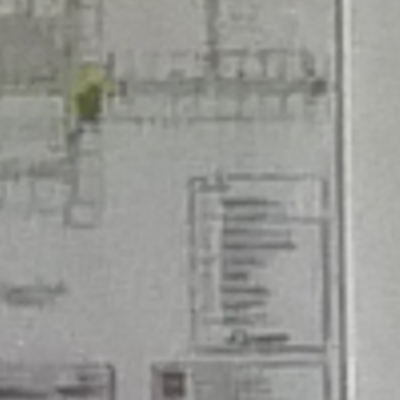}     & 
        \includegraphics[width=\w\textwidth]{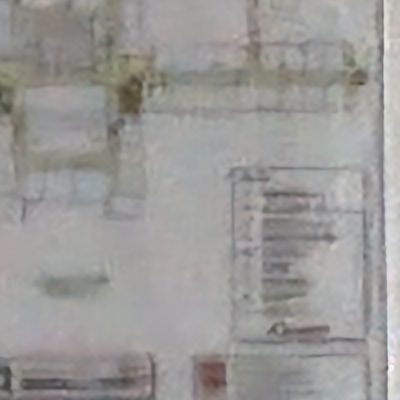}     & 
        \includegraphics[width=\w\textwidth]{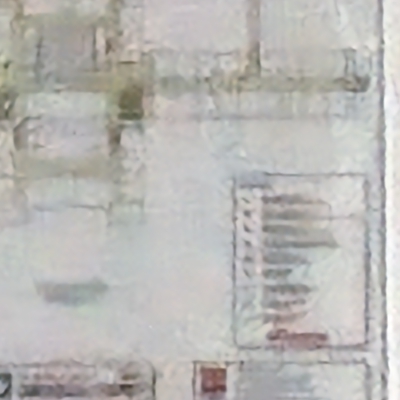}     & 
        \includegraphics[width=\w\textwidth]{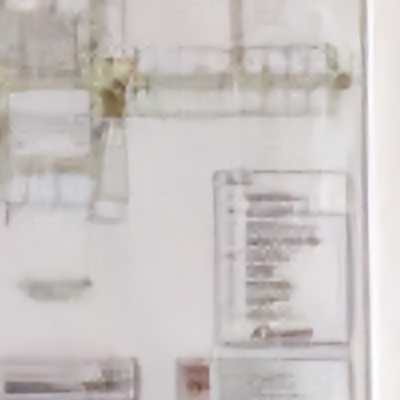}     & 
        \includegraphics[width=\w\textwidth]{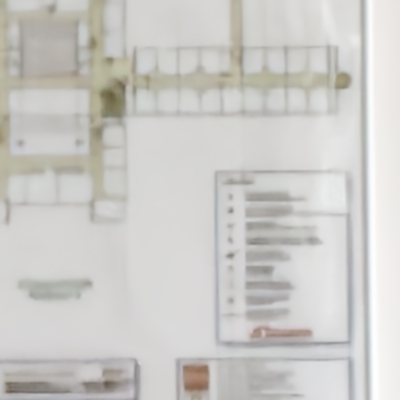}     \\

        \parbox[t]{4mm}{\rotatebox[origin=l]{90}{Pixel 4a}} & 
        \includegraphics[width=\w\textwidth]{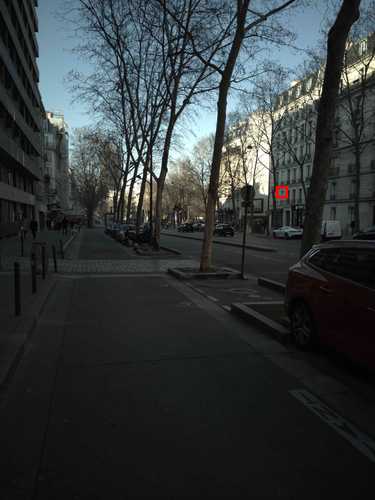}     & 
        \includegraphics[width=\w\textwidth]{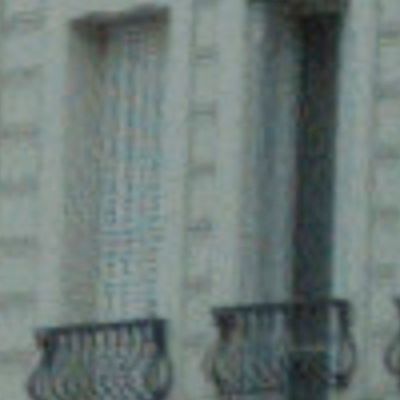}     & 
        \includegraphics[width=\w\textwidth]{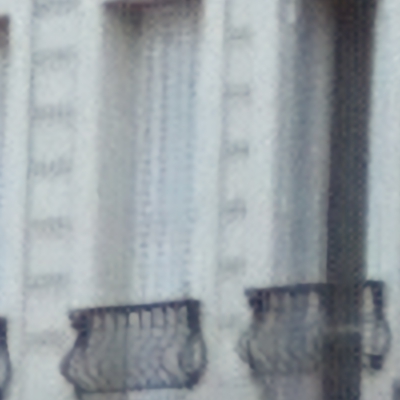}     & 
        \includegraphics[width=\w\textwidth]{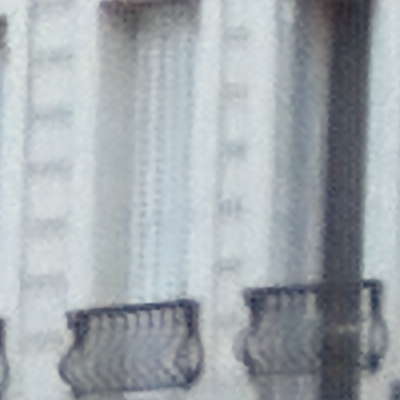}     & 
        \includegraphics[width=\w\textwidth]{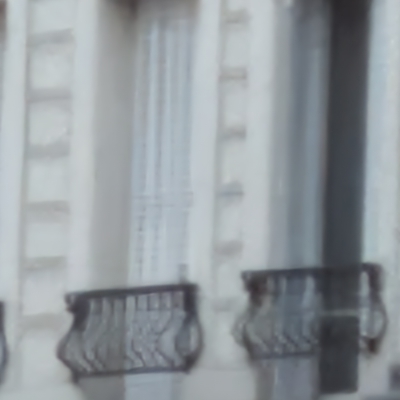}     & 
        \includegraphics[width=\w\textwidth]{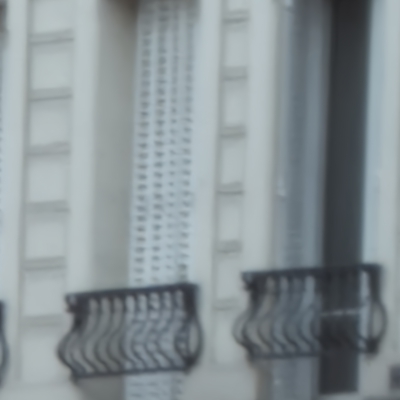}     \\

        \parbox[t]{4mm}{\rotatebox[origin=l]{90}{Canon G7X}} & 
        \includegraphics[width=\w\textwidth]{appendix/b1/crops/full_crop.jpg}     & 
        \includegraphics[width=\w\textwidth]{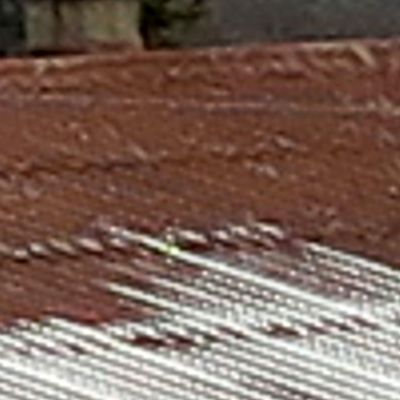}     & 
        \includegraphics[width=\w\textwidth]{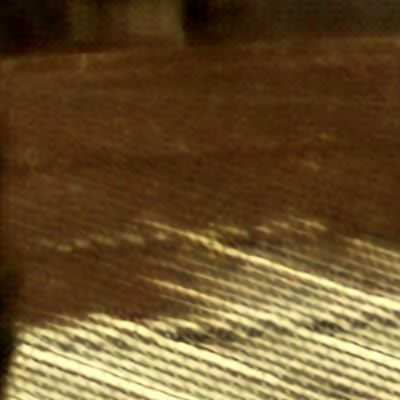}     & 
        \includegraphics[width=\w\textwidth]{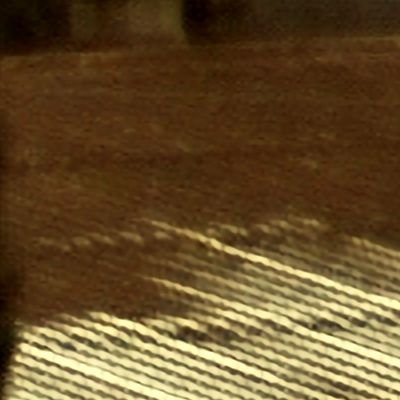}     & 
        \includegraphics[width=\w\textwidth]{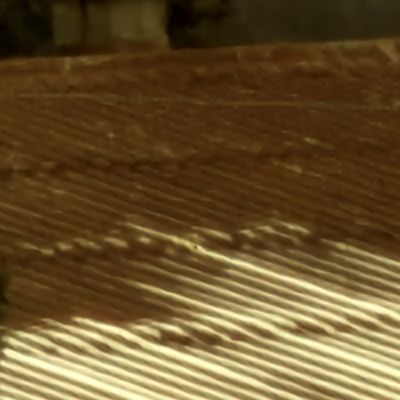}     & 
        \includegraphics[width=\w\textwidth]{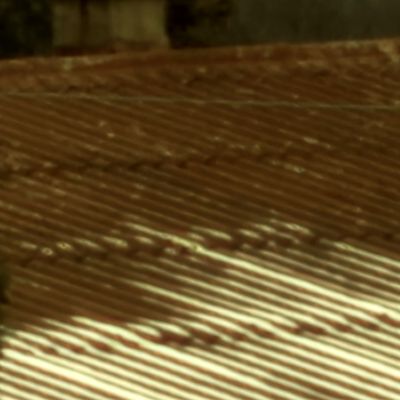}     \\


        & Full image & Camera ISP & \begin{tabular}{c} groupsc  \cite{lecouat2020fully} \\ + EDSR \cite{lim2017enhanced} \end{tabular} & \begin{tabular}{c} groupsc \cite{lecouat2020fully} \\ + VSR-DUF \cite{jo2018deep} \end{tabular} & Joint mosa \& SR & Ours \\
        \end{tabular}
        \vsp
    \caption{Results from real raw image bursts obtained with various cameras. We provide comparisons with single image and multiframe baselines. Finest restored details can be seen by zooming on computer screen.}\label{fig:realb}
\end{figure*}

\begin{figure*}[h!]
    \renewcommand\w{0.45}
    \renewcommand{\arraystretch}{1}
    \centering
    \setlength\tabcolsep{0.5pt}
    \begin{tabular}{cc}
        \includegraphics[width=\w\textwidth]{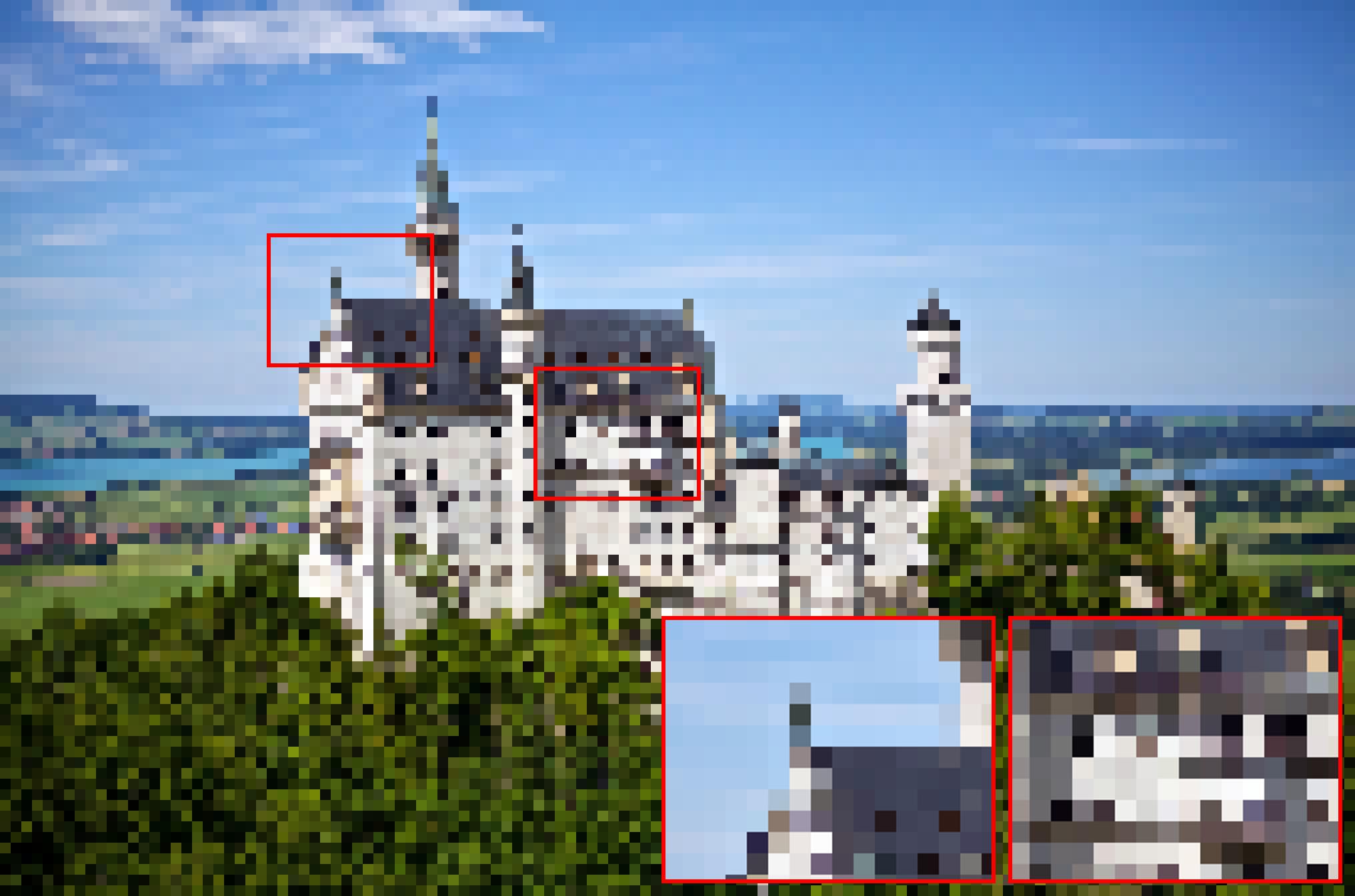}     & 
        \includegraphics[width=\w\textwidth]{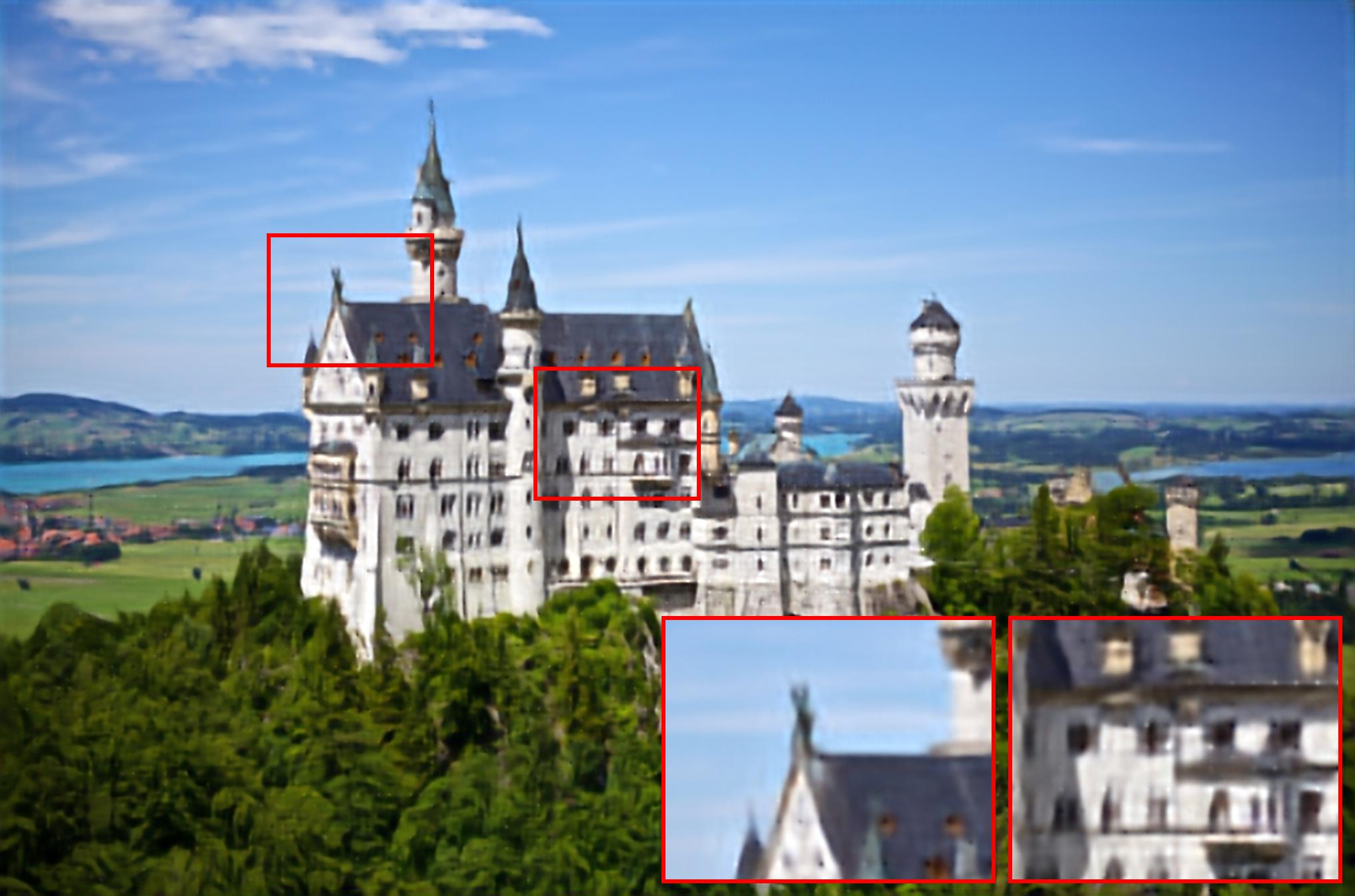}    \\ 
        \includegraphics[width=\w\textwidth]{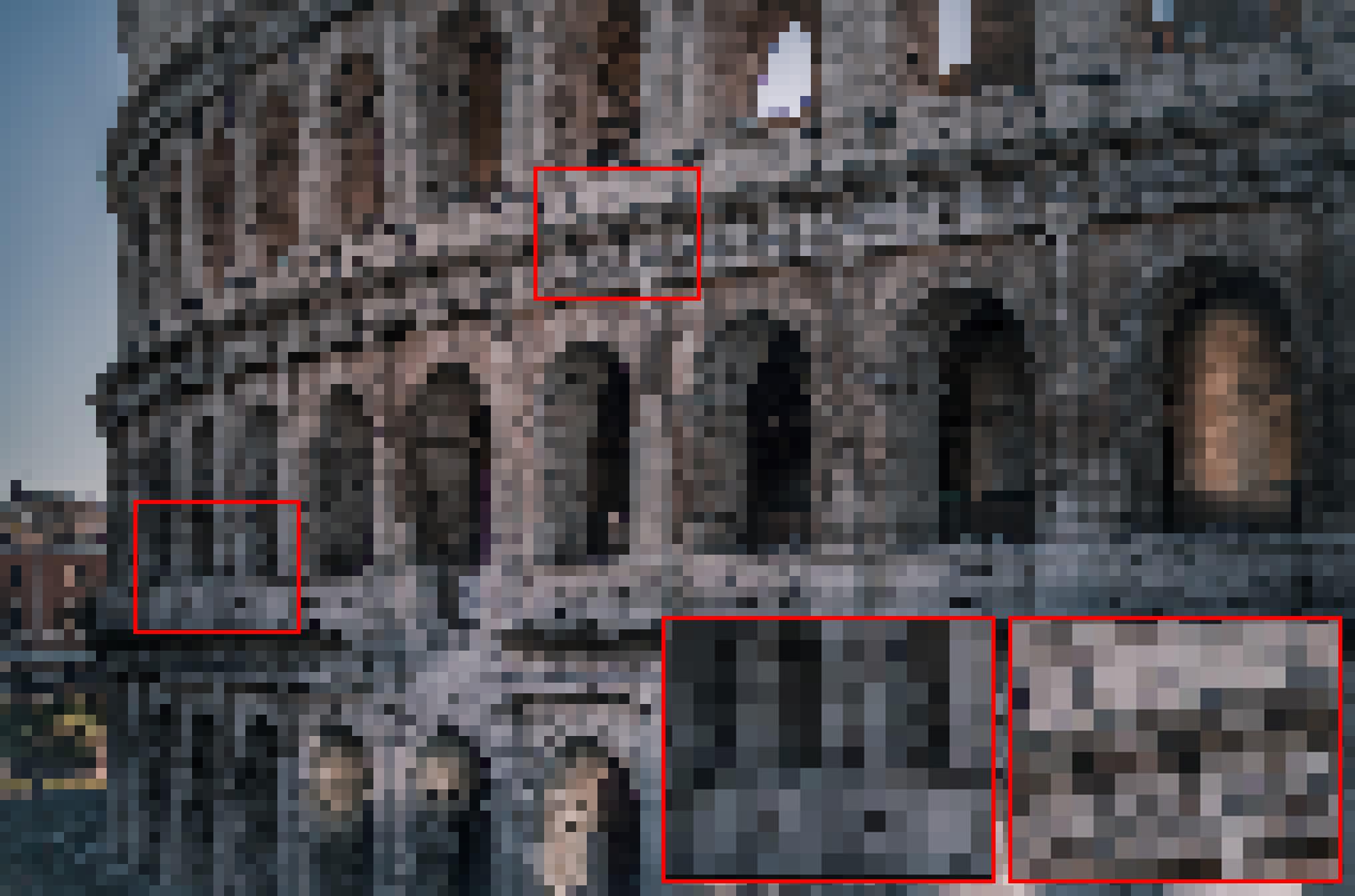}     & 
        \includegraphics[width=\w\textwidth]{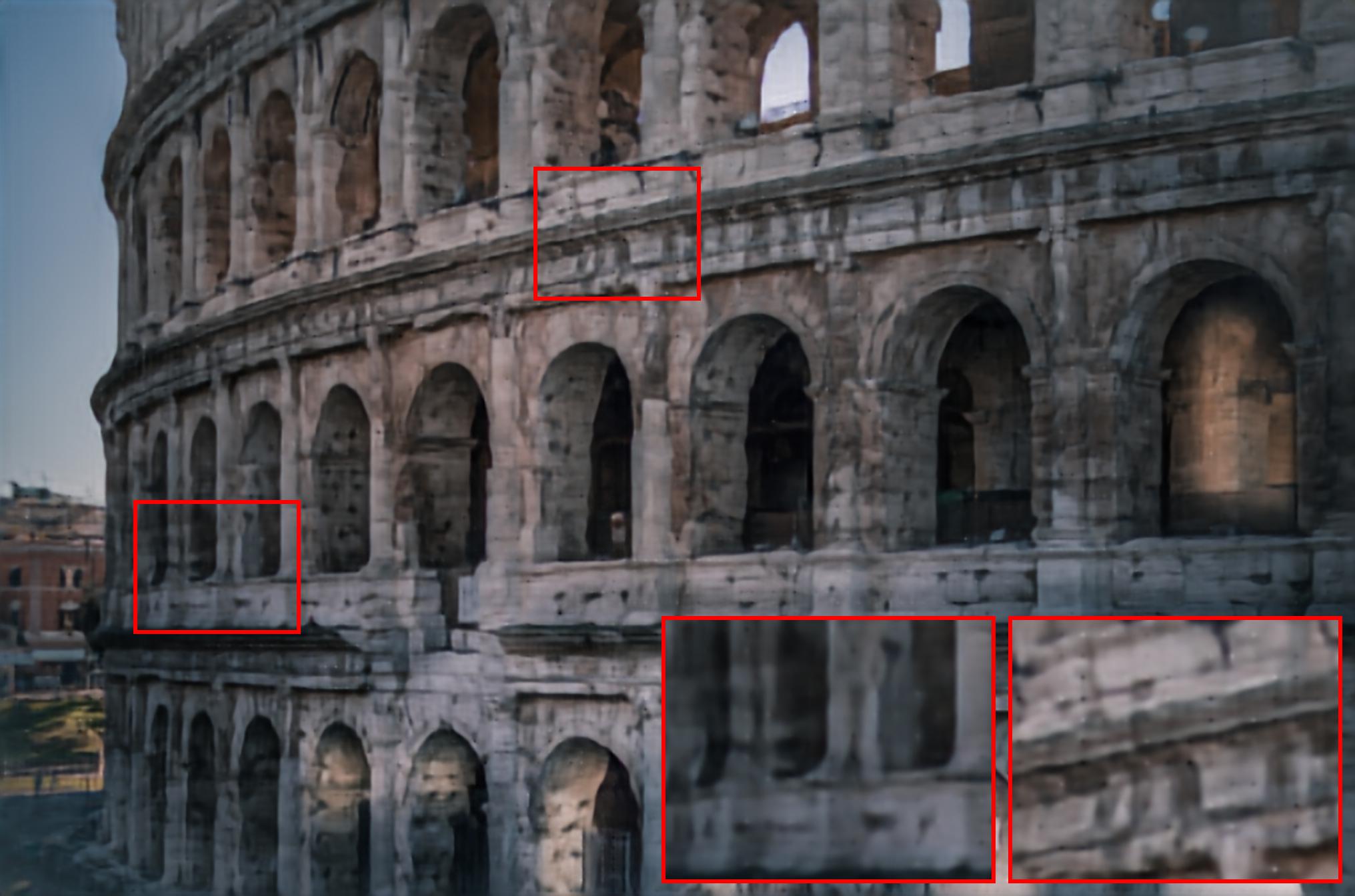}    \\

        \includegraphics[width=\w\textwidth]{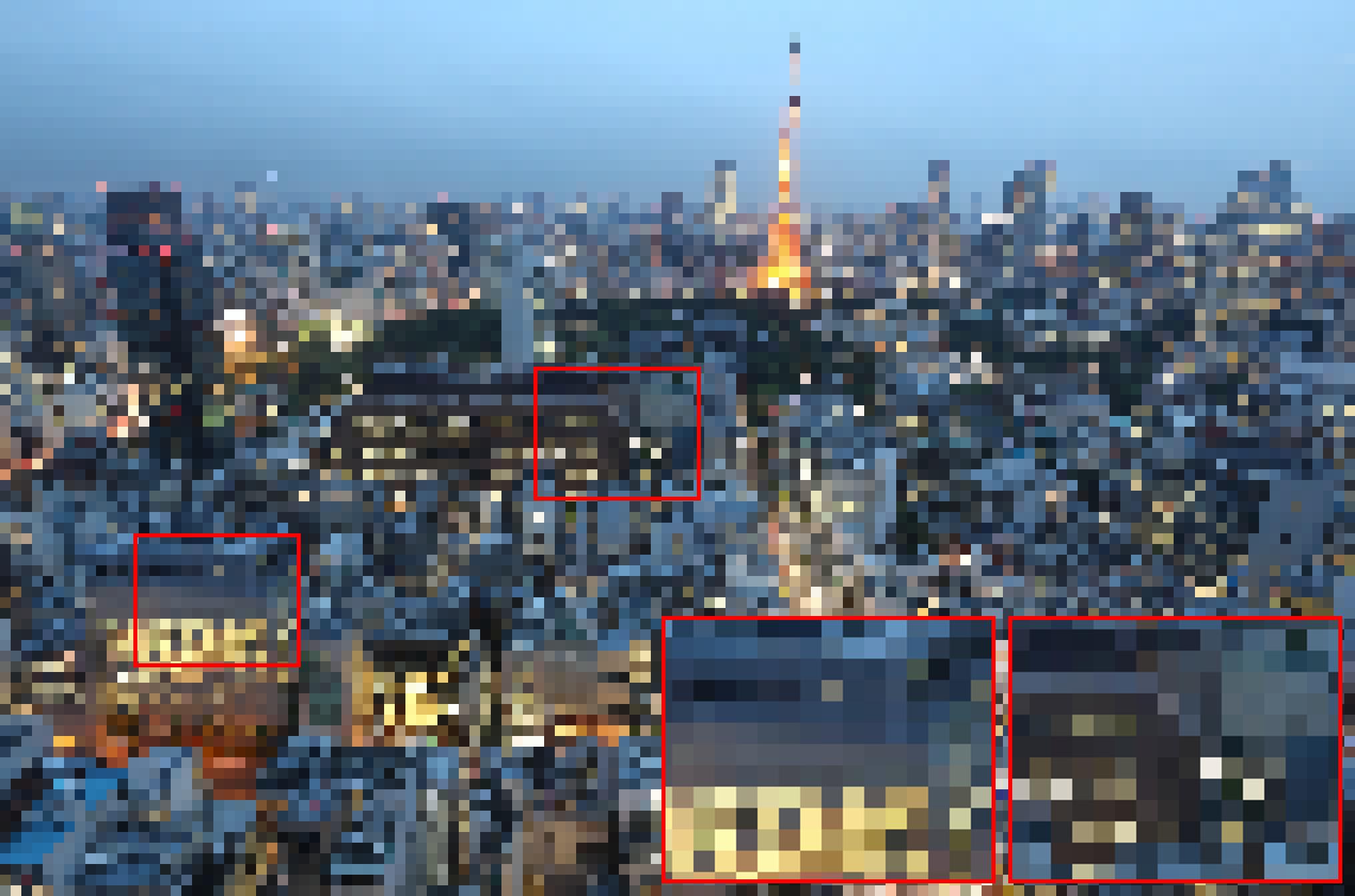}     & 
        \includegraphics[width=\w\textwidth]{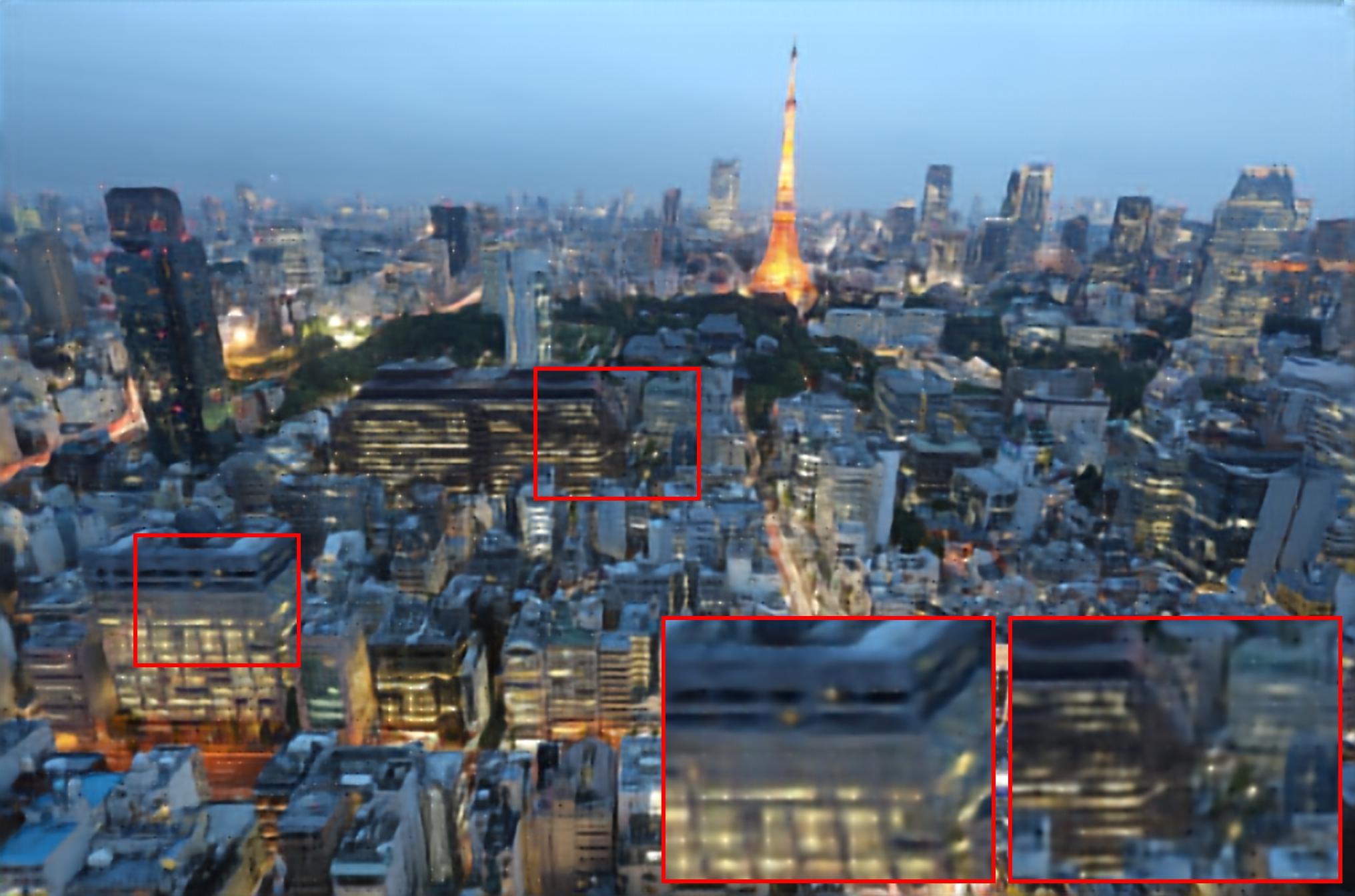}    \\ 
        \includegraphics[width=\w\textwidth]{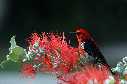}     & 
        \includegraphics[width=\w\textwidth]{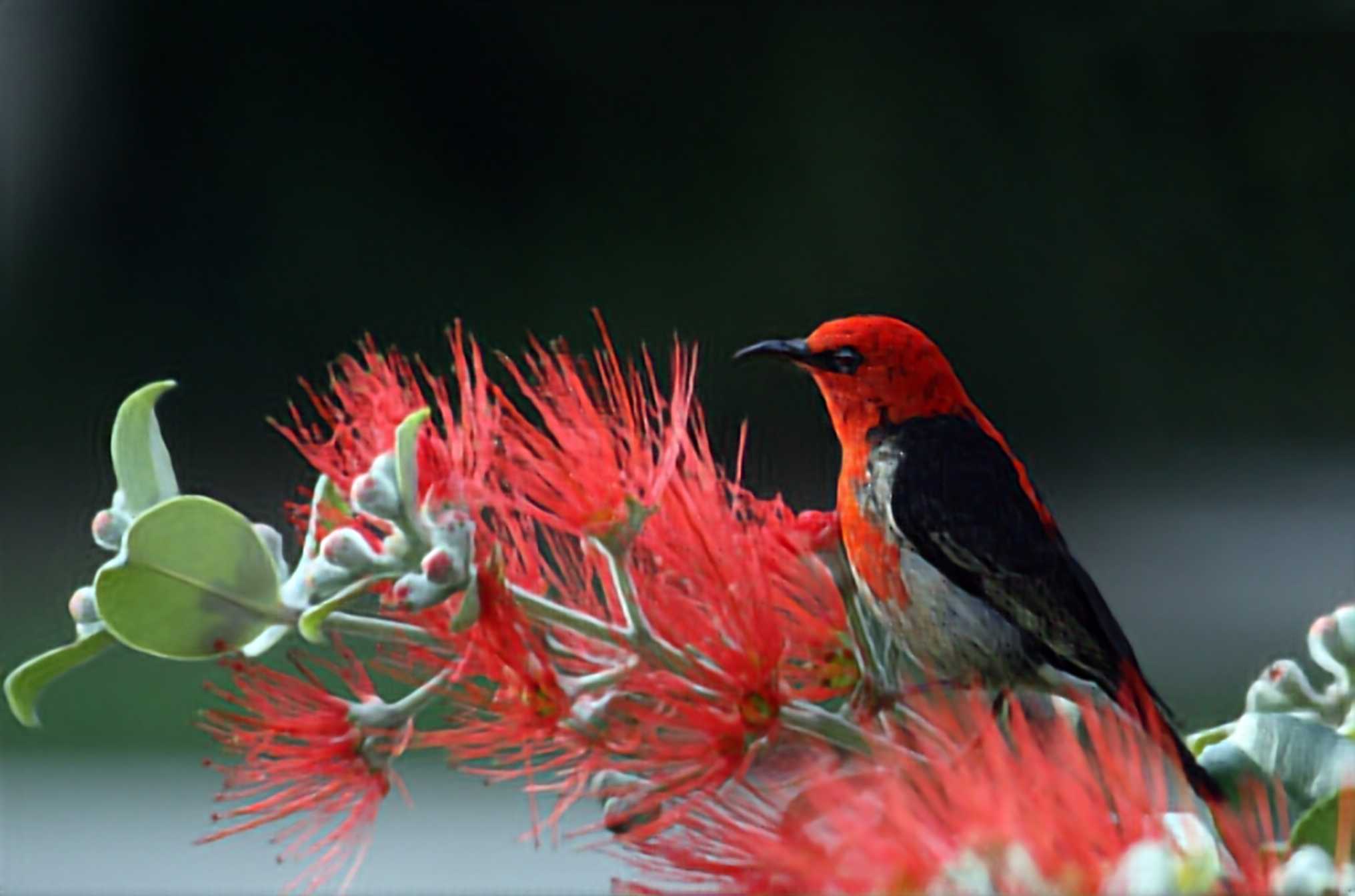}    \\ 
    \end{tabular}
    \caption{Extreme $\times 16$ upsampling experiment.  The right image is obtained by processing a burst of 20 LR  images presented on the left obtained with synthetic random  affine movements and average pooling downsampling} \label{fig:x16b}
\end{figure*}
 
\begin{figure*}[h!]

    \renewcommand\w{0.3}
    \renewcommand{\arraystretch}{1}
    \centering
    \setlength\tabcolsep{0.2pt}
    \begin{tabular}{cccc}
        \parbox[t]{4mm}{\rotatebox[origin=l]{90}{Panasonic}} & 
        \includegraphics[width=\w\textwidth]{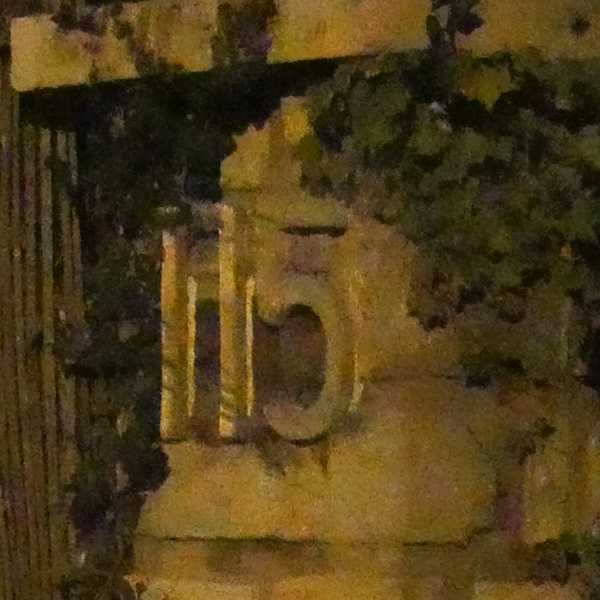}     & 
        \includegraphics[width=\w\textwidth]{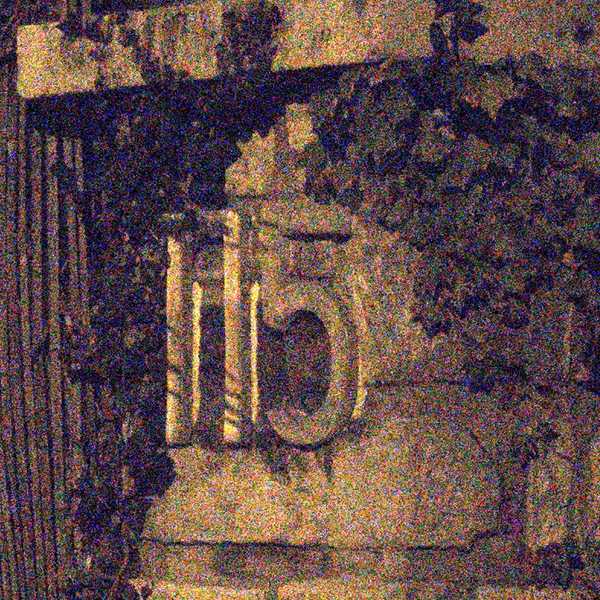}     & 
        \includegraphics[width=\w\textwidth]{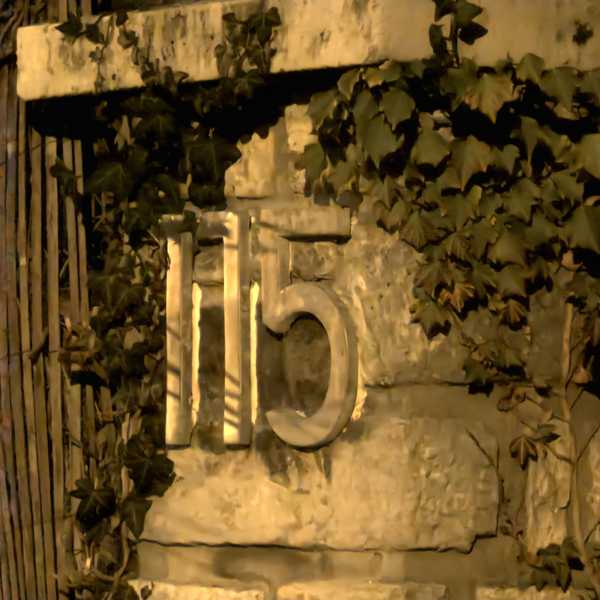}     \\
        \parbox[t]{4mm}{\rotatebox[origin=l]{90}{Panasonic}} & 
        \includegraphics[width=\w\textwidth]{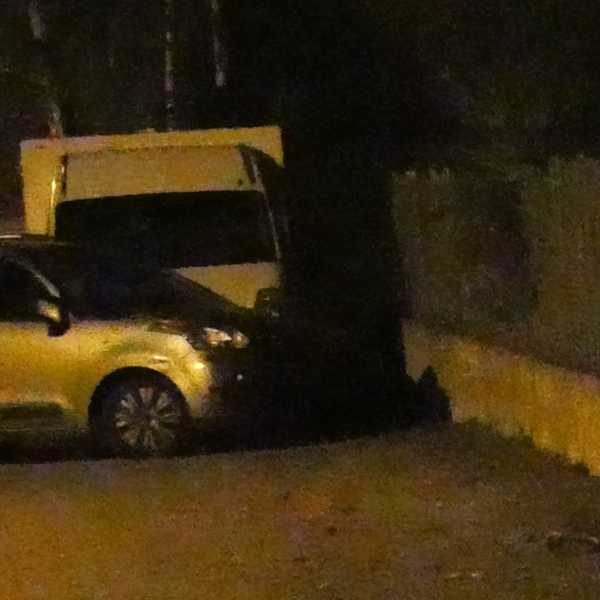}     & 
        \includegraphics[width=\w\textwidth]{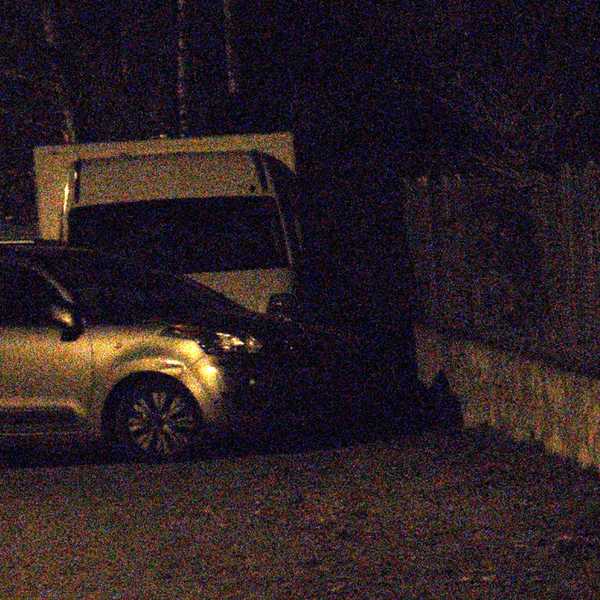}     & 
        \includegraphics[width=\w\textwidth]{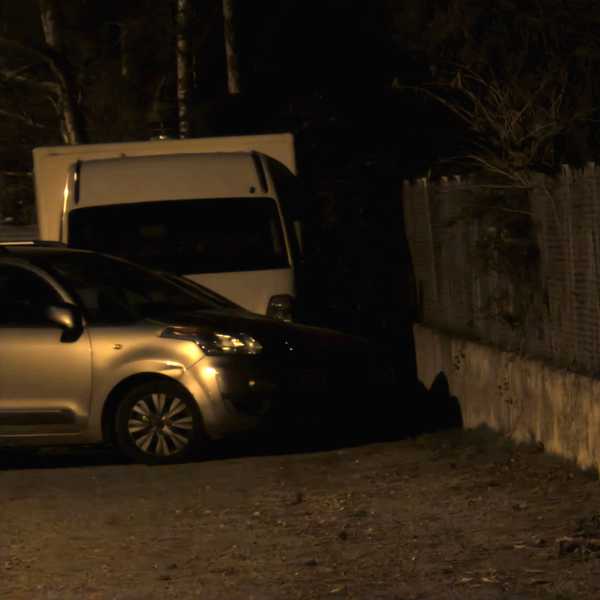}     \\
        \parbox[t]{4mm}{\rotatebox[origin=l]{90}{Panasonic}} & 
        \includegraphics[width=\w\textwidth]{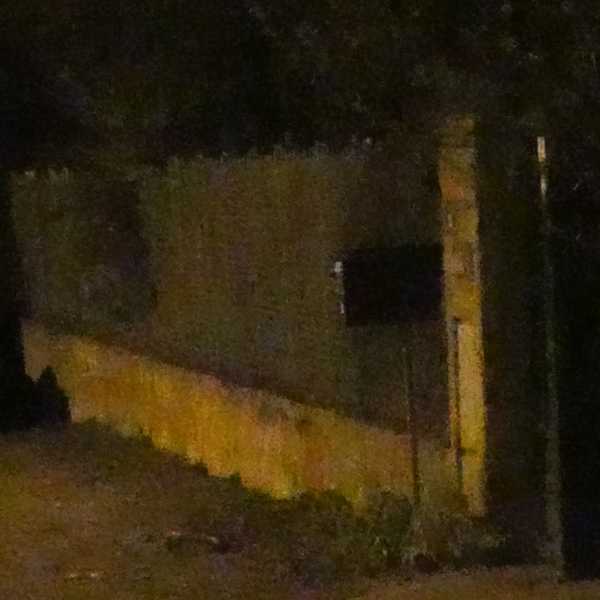}     & 
        \includegraphics[width=\w\textwidth]{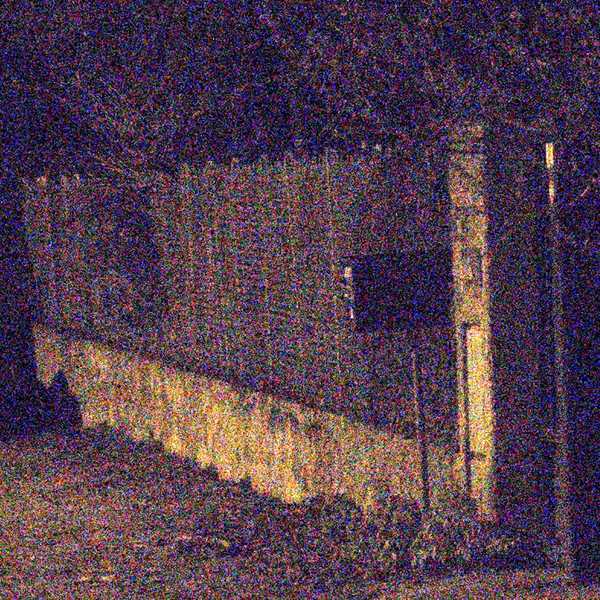}     & 
        \includegraphics[width=\w\textwidth]{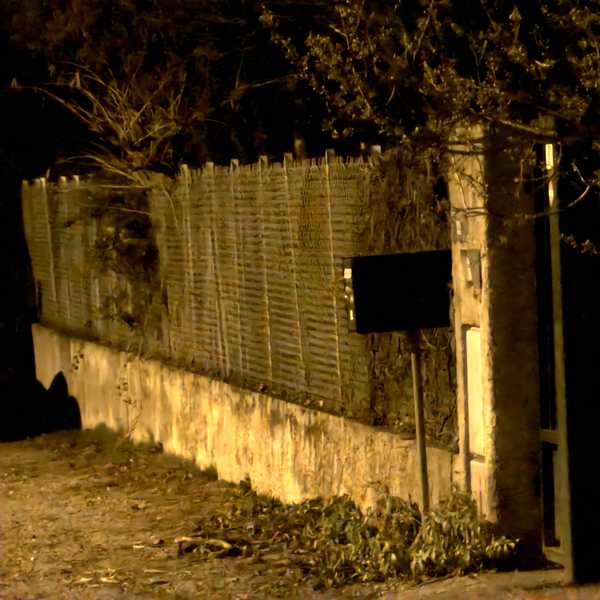}     \\
        & ISP camera & Demosaicked images & Ours \\ 
    \end{tabular}
    \caption{Image restoration of images taken at night with very low signal to noise ratio by using a Panasonic GX9 camera.}\label{fig:lowlight}
\end{figure*}


\begin{figure*}[h!]
    \renewcommand\w{0.16}
    \renewcommand{\arraystretch}{1}
    \centering
    \setlength\tabcolsep{0.2pt}
    \begin{tabular}{ccccccc}
        \parbox[t]{4mm}{\rotatebox[origin=l]{90}{Panasonic}} & 
        \includegraphics[width=\w\textwidth]{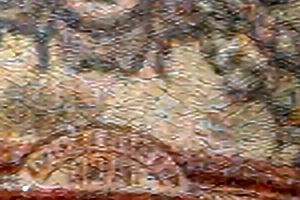}     & 
        \includegraphics[width=\w\textwidth]{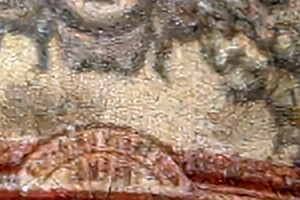}     & 
        \includegraphics[width=\w\textwidth]{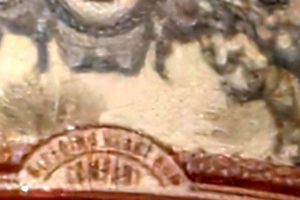}     & 
        \includegraphics[width=\w\textwidth]{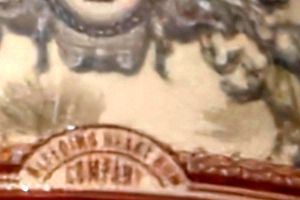}     & 
        \includegraphics[width=\w\textwidth]{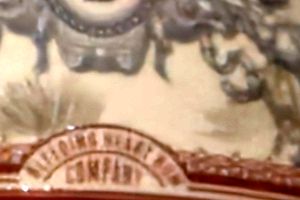}     &
        \includegraphics[width=\w\textwidth]{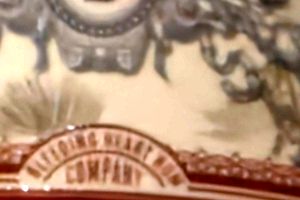}    \\
        \parbox[t]{4mm}{\rotatebox[origin=l]{90}{Panasonic}} & 
        \includegraphics[width=\w\textwidth]{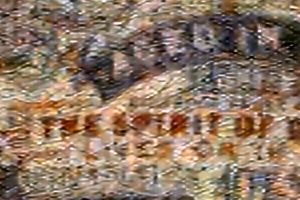}     & 
        \includegraphics[width=\w\textwidth]{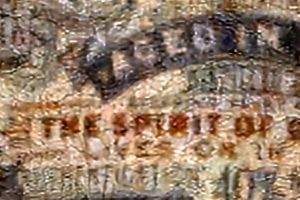}     & 
        \includegraphics[width=\w\textwidth]{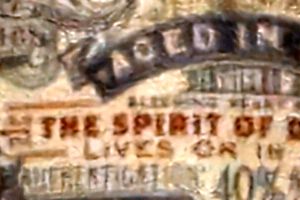}     & 
        \includegraphics[width=\w\textwidth]{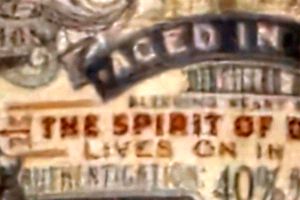}     & 
        \includegraphics[width=\w\textwidth]{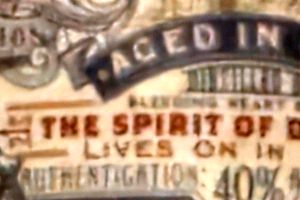}     &
        \includegraphics[width=\w\textwidth]{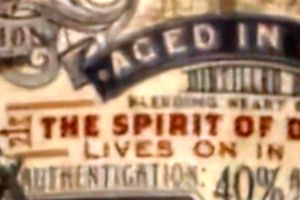}    \\
        \parbox[t]{4mm}{\rotatebox[origin=l]{90}{Panasonic}} & 
        \includegraphics[width=\w\textwidth]{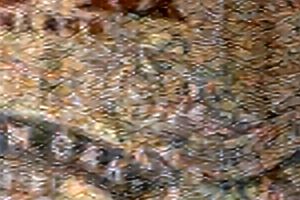}     & 
        \includegraphics[width=\w\textwidth]{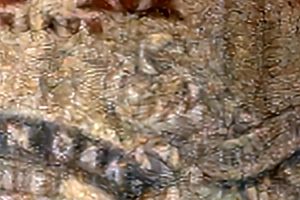}     & 
        \includegraphics[width=\w\textwidth]{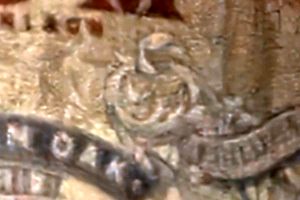}     & 
        \includegraphics[width=\w\textwidth]{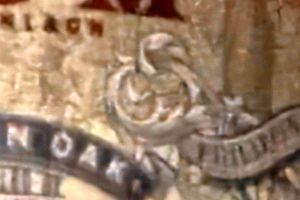}     & 
        \includegraphics[width=\w\textwidth]{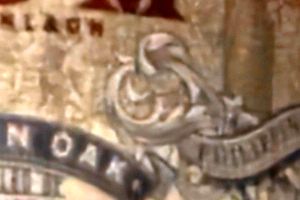}     &
        \includegraphics[width=\w\textwidth]{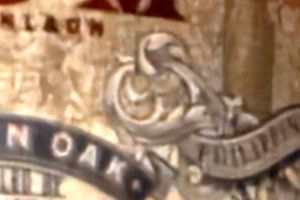}    \\
        
       & 2 & 4 &8 & 14 & 20 & 30 \\
    \end{tabular}
    \caption{Visual differences caused by merging a different number of frames in the case of low SNR scenes. With a larger number of frames we can observe a quality increase and better denoising.}\label{fig:ablation}
\end{figure*}

\begin{figure*}[h!]
    \renewcommand\w{0.16}
    \renewcommand{\arraystretch}{1}
    \centering
    \setlength\tabcolsep{0.2pt}
    \begin{tabular}{ccccccc}
        \parbox[t]{4mm}{\rotatebox[origin=l]{90}{Panasonic}} & 
        \includegraphics[width=\w\textwidth]{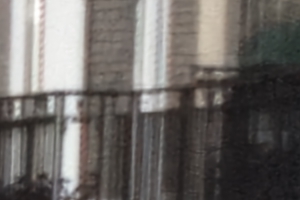}     & 
        \includegraphics[width=\w\textwidth]{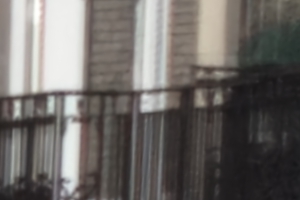}     & 
        \includegraphics[width=\w\textwidth]{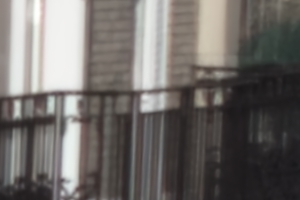}     & 
        \includegraphics[width=\w\textwidth]{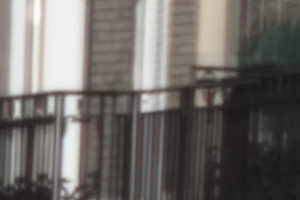}     & 
        \includegraphics[width=\w\textwidth]{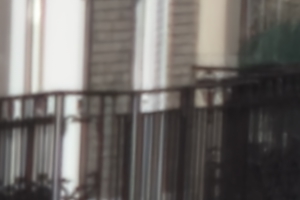}     &
        \includegraphics[width=\w\textwidth]{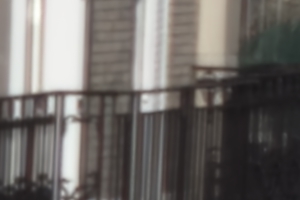}    \\
        \parbox[t]{4mm}{\rotatebox[origin=l]{90}{Panasonic}} & 
        \includegraphics[width=\w\textwidth]{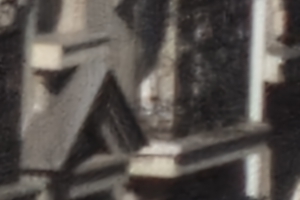}     & 
        \includegraphics[width=\w\textwidth]{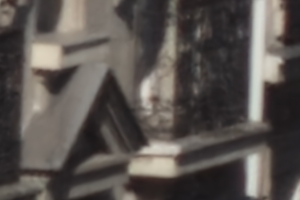}     & 
        \includegraphics[width=\w\textwidth]{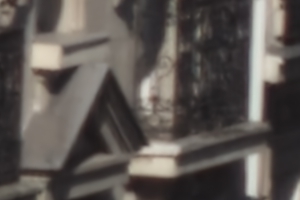}     & 
        \includegraphics[width=\w\textwidth]{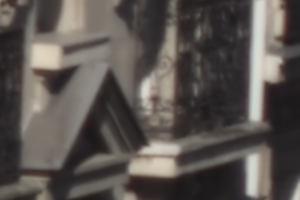}     & 
        \includegraphics[width=\w\textwidth]{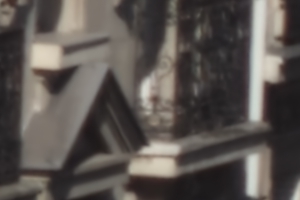}     & 
        \includegraphics[width=\w\textwidth]{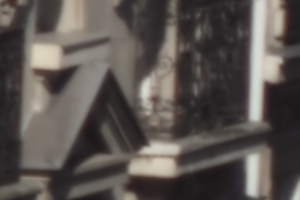}     \\        
       & 2 & 4 &8 & 12 & 14 & 20 \\
    \end{tabular}
    \caption{Visual differences caused by merging a different number of frames in the case of high SNR scenes. With a larger number of frames we can observe a quality increase.}\label{fig:ablationb}
\end{figure*}

\begin{figure}[h!]
    \renewcommand\w{0.16}
    \renewcommand{\arraystretch}{1}
    \centering
    \setlength\tabcolsep{0.2pt}
    \begin{tabular}{cccccc}
        \includegraphics[width=0.12\textwidth]{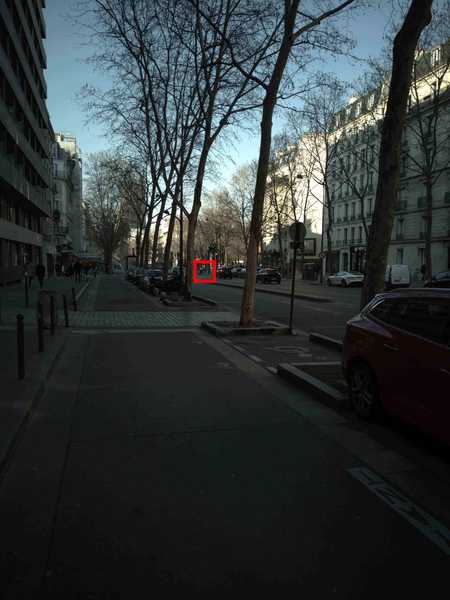}   & 
        \includegraphics[width=\w\textwidth]{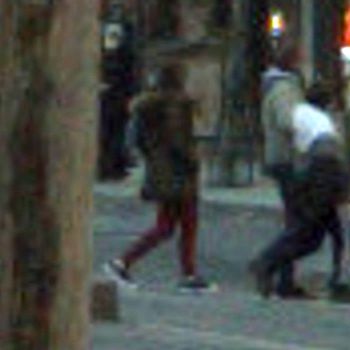}     & 
        \includegraphics[width=\w\textwidth]{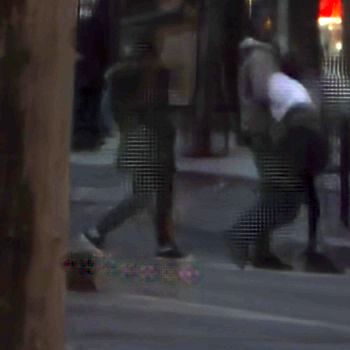}     &

        \includegraphics[width=0.12\textwidth]{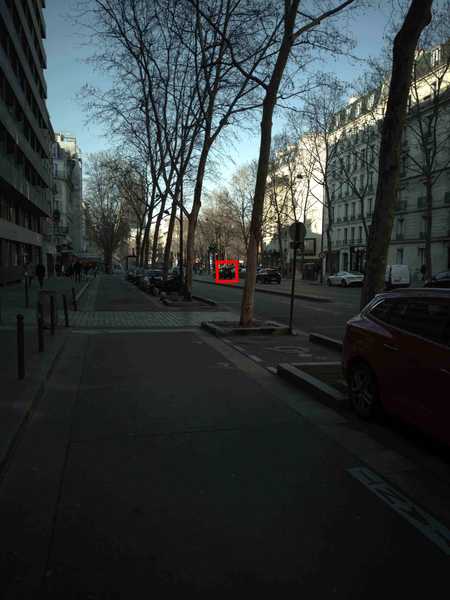}   & 
        \includegraphics[width=\w\textwidth]{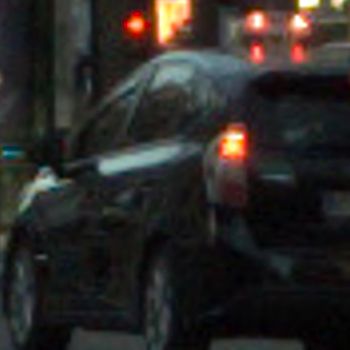}     & 
        \includegraphics[width=\w\textwidth]{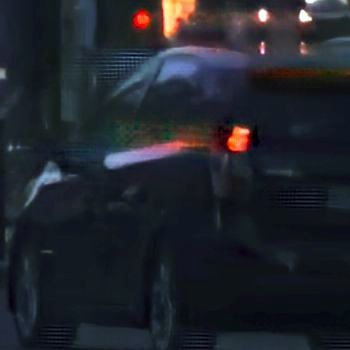}     \\

        \includegraphics[width=0.12\textwidth]{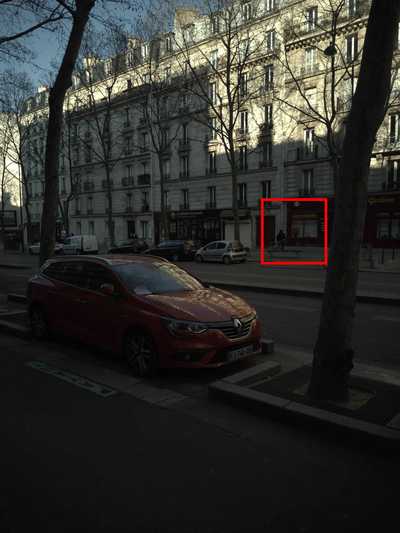}   & 
        \includegraphics[width=\w\textwidth]{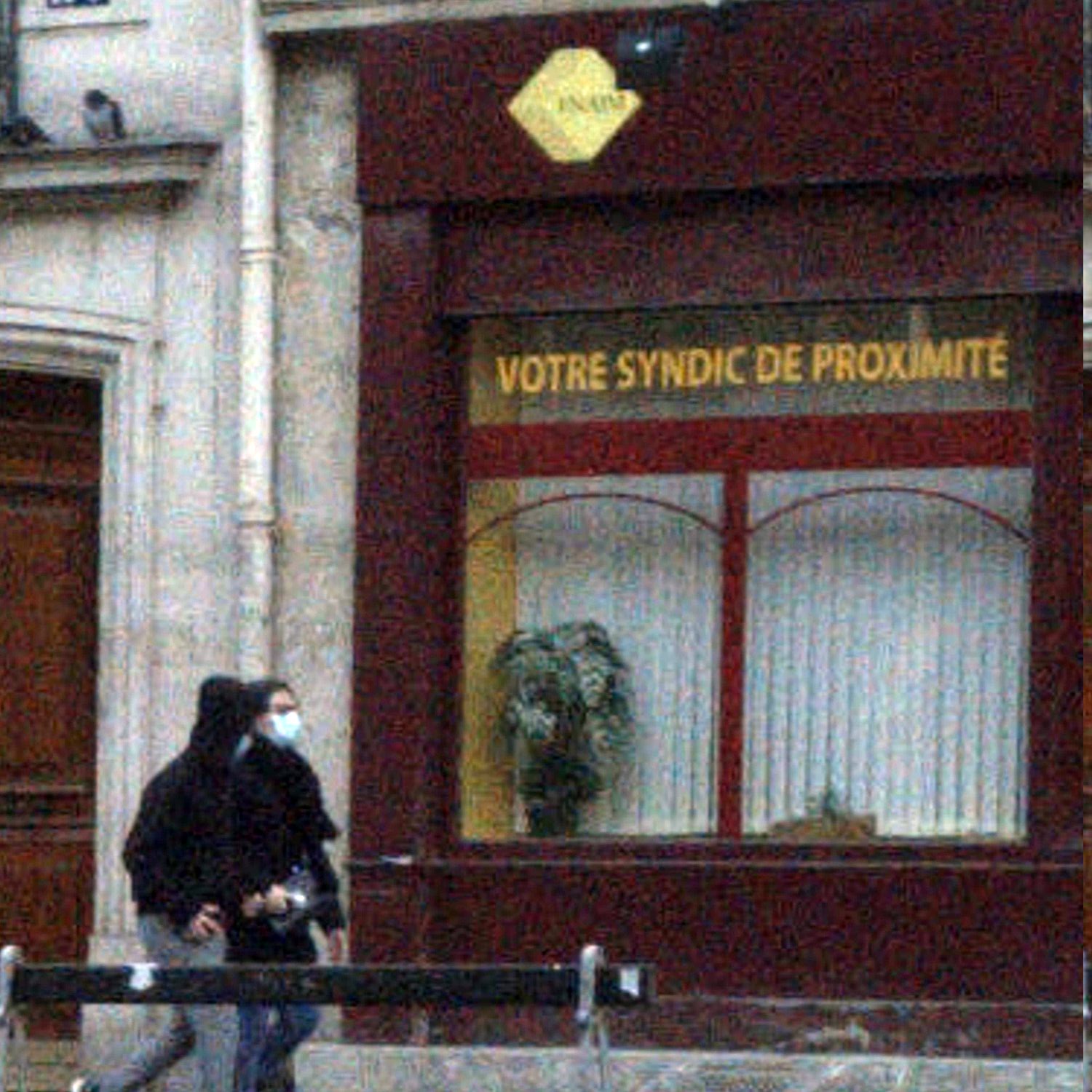}     & 
        \includegraphics[width=\w\textwidth]{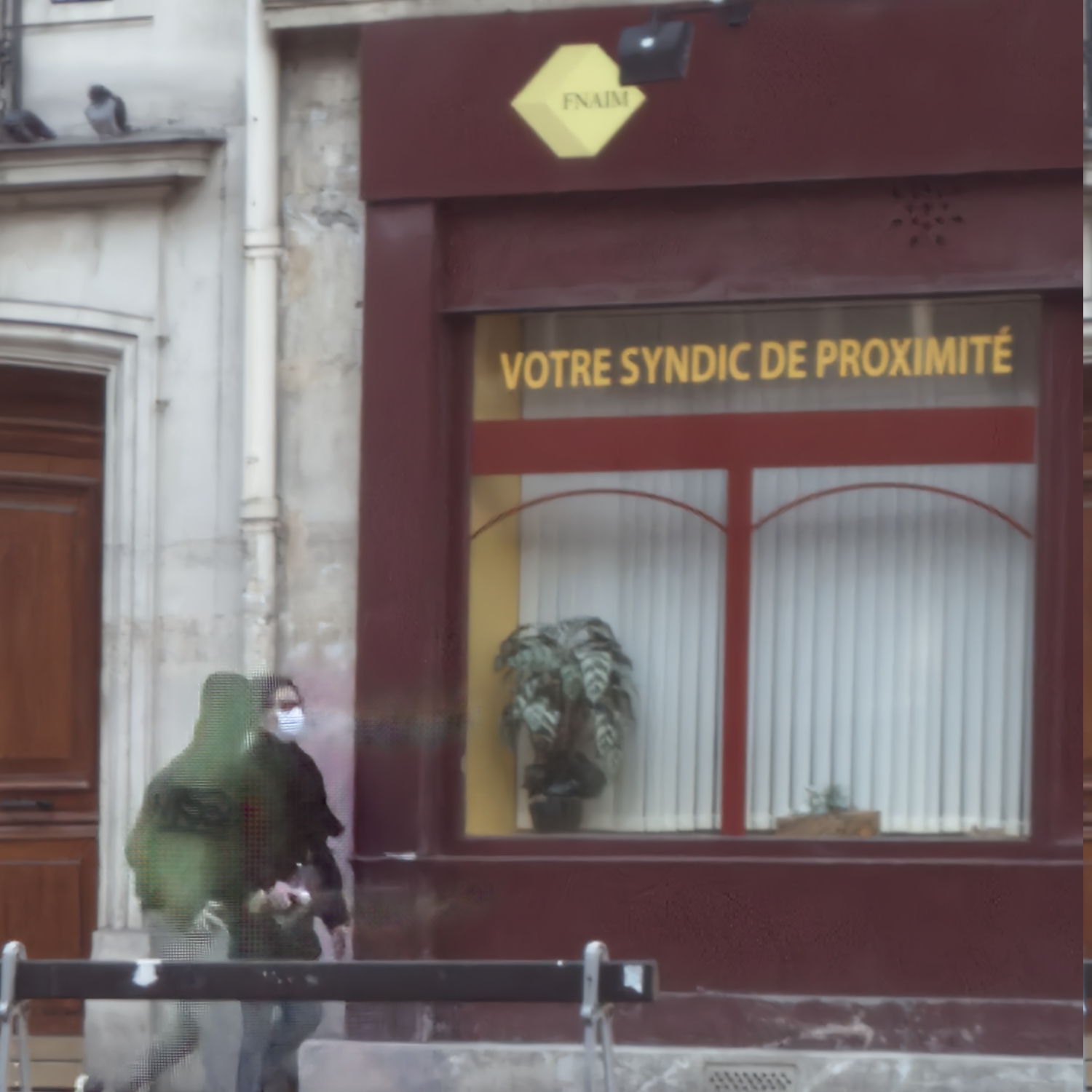}     &

        \includegraphics[width=0.12\textwidth]{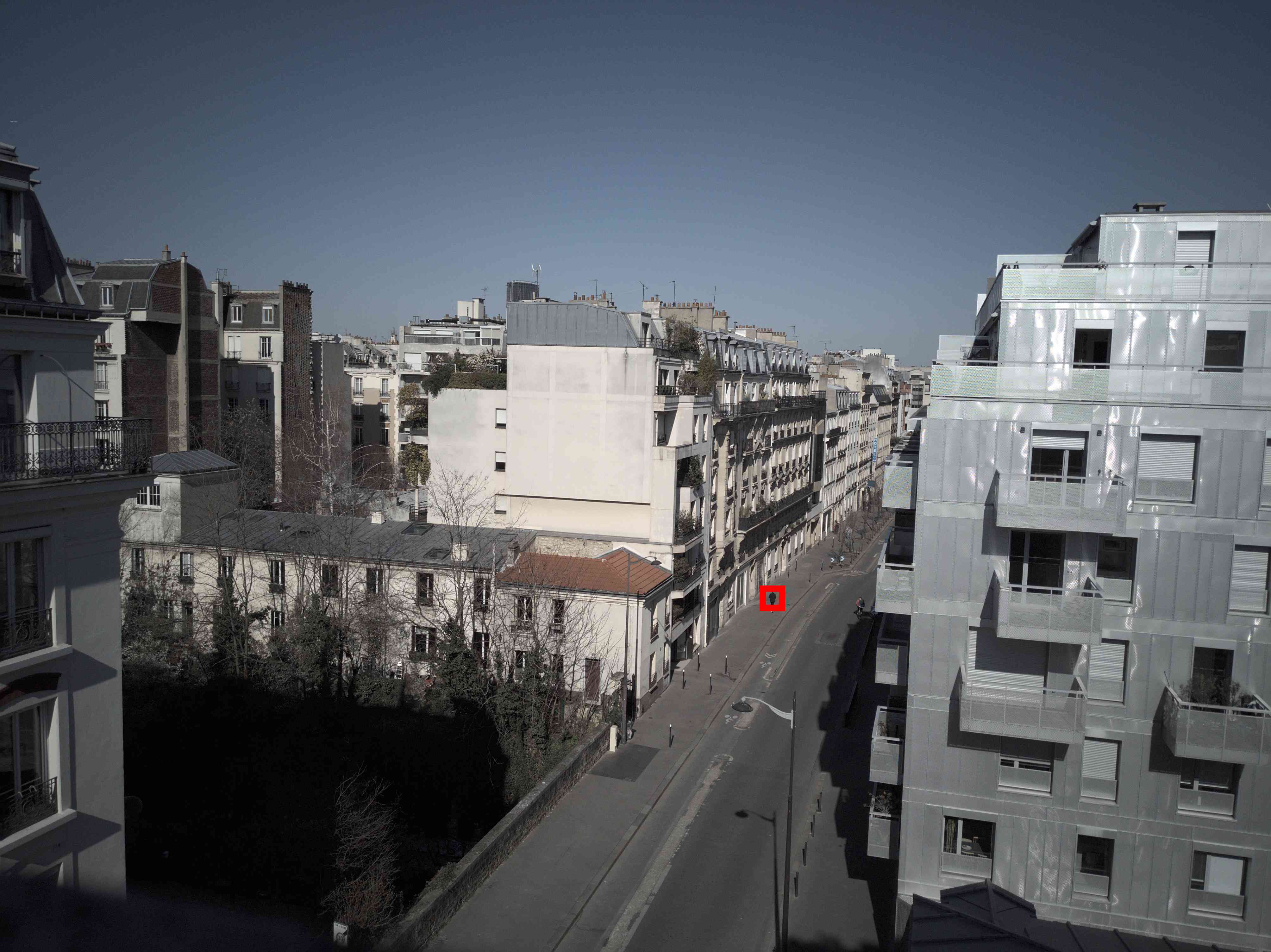}   & 
        \includegraphics[width=\w\textwidth]{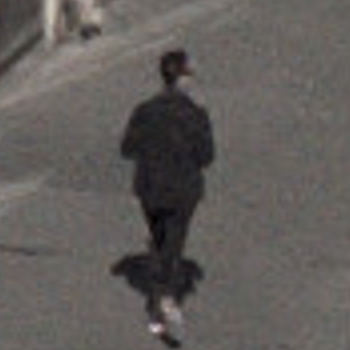}     & 
        \includegraphics[width=\w\textwidth]{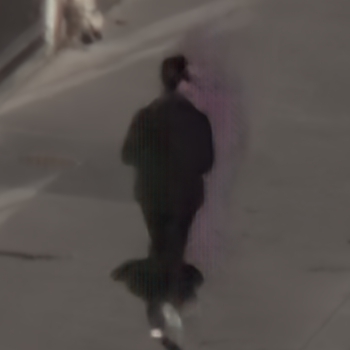}      \\

    full frame & ISP camera & Ours &  full frame & ISP camera & Ours \\
    \end{tabular}
    \caption{Misalignements artefacts due to moving objects in the scene. Our current implementation does not handle fast moving objects and then generates visual artefacts. Dealing with fast dynamic scenes will be the focus of future work.}\label{fig:failure}
\end{figure}

\end{document}